\documentclass[10pt,twocolumn,letterpaper]{article}

\usepackage[pagenumbers]{cvpr} %

\usepackage{booktabs}
\usepackage{multirow}
\usepackage{xspace}
\usepackage{tikz}
\usetikzlibrary{arrows.meta,positioning,calc,fit,backgrounds,shapes.geometric}
\usepackage{pgfplots}
\pgfplotsset{compat=1.17}
\newcommand{\ours}{\textsc{Dose}\xspace}

\newcommand{\cltta}{CL\,+\,TTA\xspace}  %
\newcommand{\setupname}{continual learning with domain preview\xspace}  %
\newcommand{\clp}{CL-P\xspace}  %
\newcommand{\appref}[1]{Supp.~\cref*{#1}}

\newcommand{\sm}[1]{{\color{black!50}#1}}  %
\usepackage{algorithm}
\usepackage{algpseudocode}
\usepackage{array}
\usepackage{graphicx}
\newsavebox{\fitbox}
\newcommand{\fitwidth}[1]{\sbox{\fitbox}{#1}\ifdim\wd\fitbox>\textwidth\resizebox{\textwidth}{!}{\usebox{\fitbox}}\else\usebox{\fitbox}\fi}
\usepackage{float}  %

\definecolor{labc}{HTML}{7B4F80}
\definecolor{unlc}{HTML}{C46A12}
\newcommand{\Dtr}[1]{{\color{labc}\mathcal{D}^{\mathrm{tr}}_{#1}}}
\newcommand{\Du}[1]{{\color{unlc}\mathcal{D}^{\mathrm{u}}_{#1}}}
\definecolor{evalc}{HTML}{3E5C7C}
\newcommand{\Dte}[1]{{\color{evalc}\mathcal{D}^{\mathrm{te}}_{#1}}}
\newcommand{\trn}[1]{{\color{labc}#1}}
\newcommand{\prv}[1]{{\color{unlc}#1}}
\newcommand{\evl}[1]{{\color{evalc}#1}}
\definecolor{dosec}{HTML}{2E7D32}

\definecolor{cvprblue}{rgb}{0.21,0.49,0.74}
\usepackage[breaklinks,colorlinks,allcolors=cvprblue]{hyperref}

\def\confName{CVPR}
\def\confYear{2027}

\title{Bridging the Gap between Continual Learning and Test-Time Adaptation with Domain Preview}

\author{%
Fr\'ed\'eric Li Combeau$^{1\dagger}$ \qquad Nicolas Michel$^{2\dagger}$ \qquad Toshihiko Yamasaki$^{3}$\\
$^{1}$Sorbonne University \qquad $^{2}$Mercari \qquad $^{3}$The University of Tokyo\\
{\small $^{\dagger}$Work done at The University of Tokyo}}

\begin{document}
\newcommand{\nCells}{24}\newcommand{\nWinAlone}{24}\newcommand{\nWinPlain}{23}\newcommand{\nBelow}{5}\newcommand{\nBelowFair}{11}\newcommand{\nPairCells}{304}\newcommand{\nPairWins}{280}\newcommand{\gainAloneResNet}{7.7}\newcommand{\gainAloneViT}{0.9}\newcommand{\gainPlainResNet}{15.7}\newcommand{\gainPlainViT}{0.6}\newcommand{\medPair}{0.9}
\newcommand{\gainAloneMin}{+0.3}\newcommand{\gainAloneMax}{+35.6}\newcommand{\gainPlainResMin}{+0.2}\newcommand{\gainPlainResMax}{+7.3}\newcommand{\gainPlainVitMin}{+0.1}\newcommand{\gainPlainVitMax}{+1.5}\newcommand{\meanFTAlone}{+15.6}\newcommand{\meanFTPlain}{+3.5}\newcommand{\meanERAlone}{+3.3}\newcommand{\meanERPlain}{+2.5}\newcommand{\meanVFTAlone}{+1.1}\newcommand{\meanVERAlone}{+1.1}\newcommand{\nMethodCells}{122}\newcommand{\nMethodWins}{121}\newcommand{\medResNet}{$+$3.2}\newcommand{\medViT}{$+$0.6}\newcommand{\nFtCells}{138}\newcommand{\nFtWins}{103}\newcommand{\nFtAlone}{111}\newcommand{\medFt}{$+$0.6}\newcommand{\nCorrCells}{56}\newcommand{\nCorrWins}{45}\newcommand{\nCorrAlone}{38}
\newcommand{\nXTen}{25}\newcommand{\nXTenWin}{23}\newcommand{\nXTenAlone}{16}
\newcommand{\maxStd}{1.7}\newcommand{\medStd}{0.3}\newcommand{\maxStdWhere}{ViT FT, C100-C}
\newcommand{\nHeldCells}{24}\newcommand{\nHeldWins}{23}\newcommand{\nSeeds}{5}\newcommand{\nSigAlone}{21}\newcommand{\nSigPlain}{23}\newcommand{\nSigCells}{24}
\newcommand{\nTbErWins}{165}\newcommand{\nTbErCells}{168}\newcommand{\nTbFtWins}{104}\newcommand{\nTbFtCells}{148}
\newcommand{\nStrictAlone}{24}\newcommand{\nStrictPlain}{22}\newcommand{\nStrictMin}{6}\newcommand{\nStrictMax}{14}\newcommand{\nAllAlone}{24}\newcommand{\nAllPlain}{22}\newcommand{\nAllMin}{9}\newcommand{\nAllMax}{20}\newcommand{\nCoreAlone}{24}\newcommand{\nCorePlain}{23}\newcommand{\nCoreMin}{6}\newcommand{\nCoreMax}{18}
\providecommand{\nXOne}{25}\providecommand{\nXOneWin}{25}\providecommand{\nXOneAlone}{25}

\maketitle
\begin{abstract}

Domain shift is a major problem for training machine learning models. While Continual Learning (CL) assumes distribution shift during training and Continual Test-Time Adaptation (CTTA) considers new domains at test time, setups that require explicit online adaptation to domain shift during both phases remain under-explored.
We propose \setupname (CL-P), a new combination of CL and TTA where both train and test data suffer from distribution shift. The training phase of CL-P is analogous to domain-incremental learning while the testing phase consists of a TTA phase on a future domain (which we call preview), and an evaluation phase on all seen domains.
The model must then adapt to new domains at train and test time.
In this paper, we evaluate mainstream CL and TTA methods that adapt only at train or test time, as well as their naive combination. Experimentally, TTA heavily damages the domain on which the CL method has just trained and increases forgetting.
Notably, this damage mostly comes from the BatchNorm statistics shared by all domains. In that sense, we propose \ours (\textbf{D}osing by \textbf{O}nline \textbf{S}tatistics of \textbf{E}ach domain), which allows a seamless combination of any CL and TTA methods by adaptive domain inference at test time, where the adaptation is applied to each image in proportion to the probability that it comes from the preview domain. Through extensive experiments with ResNet and ViT backbones, various combinations of CL and TTA baseline methods and datasets considering diverse domains, we show that \ours matches or improves on CL methods alone in all \nCells{} settings and on the naive combination in all but one. The code is provided in the supplementary material.

\end{abstract}

\section{Introduction}
\label{sec:intro}

\begin{figure*}[t]
\centering
\begin{tikzpicture}
\node[inner sep=0, anchor=south west] (img) at (0,0) {\includegraphics[width=0.78\textwidth]{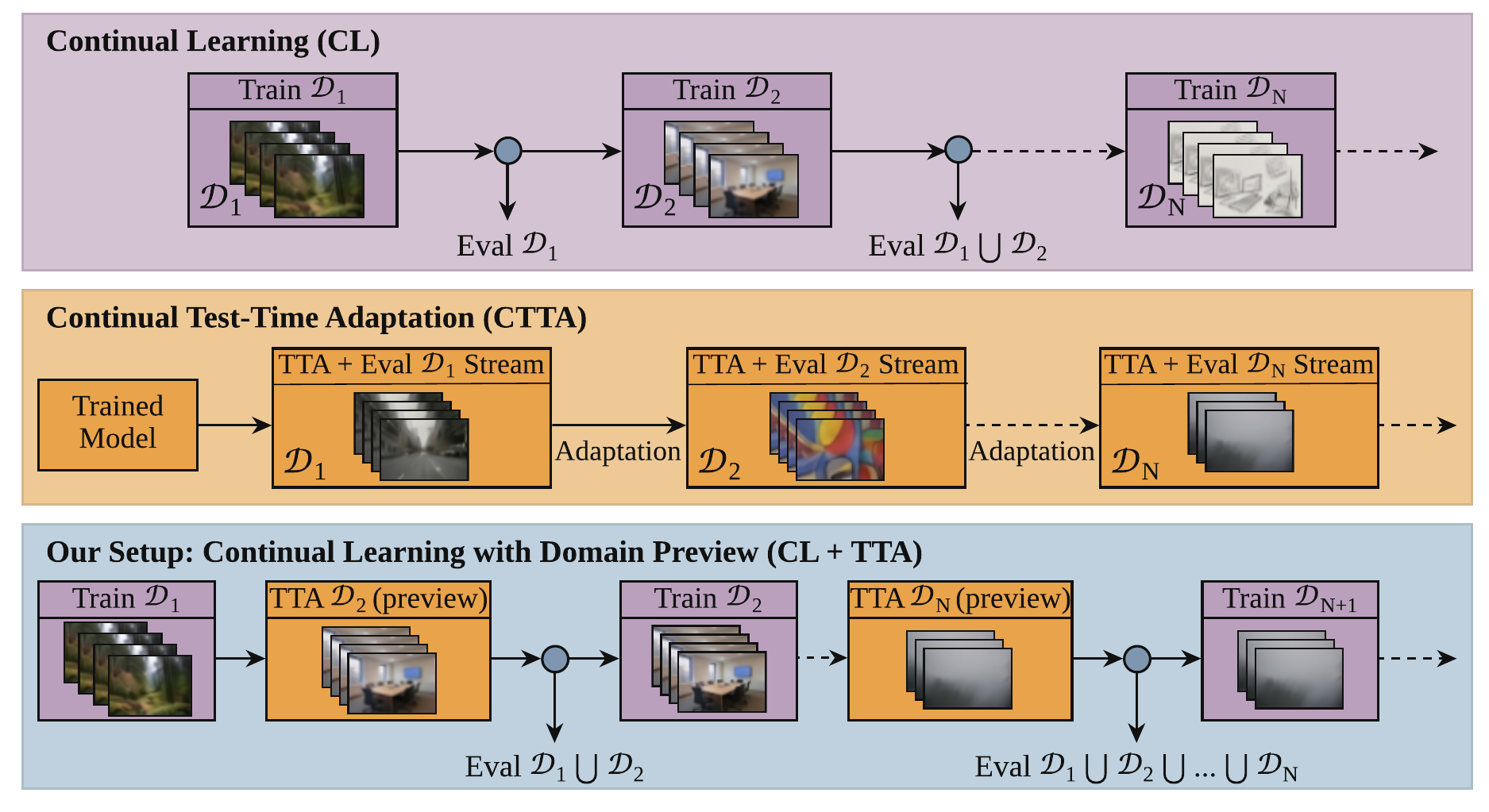}};
\begin{scope}[x={(img.south east)}, y={(img.north west)}]
\fill[fill={rgb,255:red,190;green,208;blue,222}] (0.512,0.295) rectangle (0.626,0.341);
\end{scope}
\end{tikzpicture}
\vspace{-2mm}
\caption{\textbf{Continual learning with domain preview.} \emph{Top:} CL trains on labeled domains in sequence and is evaluated on all seen domains. \emph{Middle:} CTTA adapts a trained model to a stream of unlabeled domains and is evaluated on this stream. \emph{Bottom:} our setup consists of three phases per task, namely \trn{training} on a labeled domain, \prv{preview}, where the model adapts online to the unlabeled \emph{next} domain, and \evl{evaluation} on all domains seen so far.}
\label{fig:setting}
\vspace{-3mm}
\end{figure*}

Continual learning (CL) and test-time adaptation (TTA) are well-established paradigms in machine learning. Usually studied separately, they both tackle the more generic problem of domain shift. In domain-incremental CL, a model trains on a sequence of labeled
domains and is evaluated on how well it retains what it has learned. In
continual TTA (CTTA), a pretrained model adapts online to a changing stream
of unlabeled data~\cite{wang2022cotta,marsden2024roid,maharana2025palm}.
However, in real-world applications, a model can face both kinds of shift. An image classifier, for example, may start receiving images from a new camera before anyone has labeled them, while images from
earlier cameras keep coming in.
Until these labels arrive, the same deployed model must classify the images of every camera, so that its accuracy on the new camera and on the earlier ones matters during this delay, and not only after the next training phase.

We introduce \setupname (\clp) to study how CL and TTA methods interact in this situation (\cref{fig:setting}). Specifically, each task consists of three phases. During the \trn{\emph{training}} phase, the model is trained on a labeled domain with any CL method. During the \prv{\emph{preview}} phase, it receives an unlabeled stream from the domain it will train on next, and can adapt to it online with any TTA method~\cite{wang2021tent,niu2022eata,niu2023sar}. Finally, during the \evl{\emph{evaluation}} phase, the adapted model is evaluated on all domains seen so far, including the \prv{preview} one. Similarly to label delay~\cite{csaba2024labeldelay}, data thus arrive before their labels, except that the delay here spans a whole domain.

On the one hand, CL methods alone have to rely on generalization when they encounter the \prv{preview}
domain, since they have yet to train on that domain. Naturally, their performance on the \prv{preview} domain stays far below what they reach once they have trained on this domain with labels (\cref{tab:story}). On the other hand, TTA methods can use the
unlabeled \prv{preview} immediately. While they improve accuracy on the \prv{preview} domain, applying the adapted model to every image can drastically reduce performance on the domain the model has just trained on.
The direct combination of CL and TTA methods (\cltta) therefore helps on the preview domain, at the cost of accelerated forgetting across tasks.

Experimentally, we find that most of this damage comes from the BatchNorm statistics, which are shared by all domains and re-estimated by TTA methods on the \prv{preview}. Notably, in our setup, the \trn{training} phase and the \prv{preview} each contain images of a single domain, so the statistics of every domain can be recorded separately.
Therefore, we propose \ours (\textbf{D}osing by \textbf{O}nline \textbf{S}tatistics of \textbf{E}ach domain), which decides for each test image how much of the \prv{preview} adaptation to use. Specifically, \ours compares a few statistics of the early-layer features of the image with the statistics recorded for every domain seen so far, and weights the adaptation by the probability that the image comes from the \prv{preview}. Our contributions are as follows:

\begin{itemize}
    \item We introduce \setupname, a setup that brings online supervised continual
    learning and test-time adaptation into the same sequence of tasks.

    \item We analyze the strengths and limitations of CL and TTA methods as well as their naive combination in our setup, and show that most of the damage of TTA comes from shared normalization statistics.

    \item We propose \ours, a plug-in that estimates whether each image comes
    from the \prv{preview} domain and weights the TTA update accordingly.

    \item We evaluate \ours on six benchmarks with ResNet and ViT backbones,
    four CL methods and 20 TTA methods, and show that it matches or improves on the CL method alone in all \nCells{} settings and on \cltta in all but one.
\end{itemize}

\section{Related work}
\label{sec:related}

\paragraph{Continual learning.}
In domain-incremental learning~\cite{vandeven2022three}, regularization~\cite{kirkpatrick2017ewc,li2017lwf}, replay~\cite{chaudhry2019tiny,buzzega2020derpp} and prompt-based methods~\cite{wang2022l2p,wang2022sprompts,smith2023coda} limit forgetting, but assume that the labels of a domain are available when it is first met. Label delay~\cite{csaba2024labeldelay} relaxes this assumption in online CL, but its unlabeled data are the latest batches of the same stream, so that the effect of adapting to a new domain on the domains already learned is not measured.

\paragraph{Test-time data in continual learning.}
DoSAPP~\cite{singh2026dosapp} interleaves supervised sessions with unsupervised ones on test data from classes already seen, while ARC~\cite{chen2025arc} corrects at test time the bias of the classifier towards recent classes. Both use test-time data to \emph{retain} the past, not to adapt to a domain that has not been trained on yet (\appref{tab:s_setups}).

\paragraph{(Continual) test-time adaptation.}
TTA adapts a source model to unlabeled target batches. It re-estimates normalization statistics~\cite{schneider2020improving,li2016adabn}, minimizes entropy~\cite{wang2021tent}, filters samples~\cite{niu2022eata,niu2023sar}, adjusts the classifier~\cite{iwasawa2021t3a,boudiaf2022lame} or restricts the updated layers~\cite{lee2023surgical}.
Continual TTA~\cite{wang2022cotta,dobler2023rmt,marsden2024roid,press2023rdumb,maharana2025palm,yuan2023rotta,brahma2023petal,lee2024cmf,park2024law,gong2022note} adapts to a sequence of target domains and fights error accumulation, while prompt- and adapter-based variants~\cite{liu2024vida,zhang2025dpcore,gao2022dept} add extra parameters. These methods are evaluated on the domain currently streamed, and sometimes on the source domain~\cite{niu2022eata}, but never on a sequence of trained domains that a CL method keeps updating. Most of them also normalize with the statistics of the test batch, so their predictions depend on how a batch is composed, as NOTE~\cite{gong2022note} shows.
ReservoirTTA~\cite{vray2025reservoirtta} routes each test sample to one of several adapted models by online clustering of style features, but all its domains are adapted from a fixed source model and never trained on with labels.

\paragraph{Per-domain normalization and experts.}
Keeping normalization statistics per domain is a known remedy for domain shift. DSBN~\cite{chang2019dsbn} needs the domain of each test image, AdaBN~\cite{li2016adabn} re-estimates the statistics on one target domain and UnMix-TNS~\cite{tomar2024unmixtns} mixes several statistics per image within one domain. In domain-incremental learning, S-Prompts~\cite{wang2022sprompts} infers the domain of a test image from feature centroids and other methods select a domain expert by test-time training~\cite{swinnen2026remembering}, but all of them learn each domain from its labels before it is evaluated. Continual unsupervised domain adaptation~\cite{saporta2022muhdi,cho2023codag,sun2025mpm} adapts to a sequence of unlabeled domains offline, with no later supervised task.

\section{Continual learning with domain preview}
\label{sec:setting}
In this section, we recall the CL and CTTA setups, then show how our setup, \setupname (\clp), combines them and why it requires its own evaluation protocol.

\subsection{Setup}
\label{sec:setup_def}
\paragraph{Preliminaries.}
Throughout this paper, a domain refers to a distribution of images sharing the same label space but differing in appearance or content, such as images corrupted by fog, paintings, or images taken by a specific camera.
In domain-incremental learning (DIL), a learner encounters domains $0,1,\dots,T$ sequentially. At each task $t$, it is trained on the labeled set $\Dtr{t}$ and then evaluated on held-out sets $\Dte{j}$ of all domains $j\le t$.
In contrast, continual TTA (CTTA) considers a trained model that adapts online to a stream of unlabeled batches whose domain changes over time, and evaluates it on this stream.
We call a batch \emph{single-domain} when all its images come from the same domain, and \emph{mixed} when its images come from several domains.

Continual learning with domain preview lies in between these two paradigms, where each task consists of three phases (see \cref{fig:setting}).

\paragraph{\trn{Training} phase.}
During the \trn{training} phase, the learner is trained on $\Dtr{t}$ with any CL method, as in DIL. Its batches are single-domain, unless the CL method replays images of older domains from a buffer, in which case they are mixed.

\paragraph{\prv{Preview} phase.}
During the \prv{preview} phase, the learner receives an unlabeled stream $\Du{t+1}$ from domain $t{+}1$, which the next task will train on, and it can adapt to this stream with any TTA method. The \prv{preview} thus lets the model adapt to a new domain as soon as its first images arrive, instead of waiting for their labels. Notably, all \prv{preview} images come from this single new domain, so that the \prv{preview} batches are single-domain, which is exactly what TTA usually assumes. However, we consider that TTA methods are only applied during the \prv{preview} phase. The motivations behind this restriction are twofold: 1) TTA methods require additional computation at test time, which we consider limited in an online setup, and 2) TTA methods require test data to come from one unique domain which does not differ significantly from the training domain (see~\cite{niu2023sar,marsden2024roid,tomar2024unmixtns} and \cref{sec:analysis}), which cannot be guaranteed outside the \prv{preview} phase.

\paragraph{\evl{Evaluation} phase.}
Finally, during the \evl{evaluation} phase, the learner is evaluated on all trained domains $j\le t$ as well as on the \prv{preview} images, before the labels of domain $t{+}1$ arrive. Since the domain of a test image is unknown in real-world scenarios, the \evl{evaluation} batches are mixed (\cref{sec:eval_protocols}).

\paragraph{Online data.}
As in online CL with label delay~\cite{csaba2024labeldelay}, images arrive faster than they can be labeled. Each training image is presented once in the CL stream, with replay allowed for buffer-based methods.
Each \prv{preview} stream is processed in one pass, and each \prv{preview} batch is predicted upon arrival\footnote{Several passes over the \prv{preview} or storing it would turn it into offline source-free adaptation~\cite{liang2020shot}. Three training epochs per task give the same gains (\appref{sec:s_epochs}).}.
Some TTA implementations additionally revisit unlabeled source training images to initialize auxiliary statistics (\appref{sec:s_tta_impl}).
Evaluation labels and the domain identities of evaluation images are unavailable.
The boundaries of the \trn{training} and \prv{preview} phases are known, and each \prv{preview} contains only the next domain.

\subsection{Evaluation protocols}
\label{sec:eval_protocols}
\paragraph{BatchNorm statistics.}
A BatchNorm layer~\cite{ioffe2015batchnorm} normalizes each channel of its input with a mean and a variance, then applies a learned affine map. During training, it normalizes with the statistics of the current batch and keeps a running average of them, which it stores and uses at test time. Therefore, the stored statistics of a network describe the data it was trained on, and every domain is normalized with them.
On networks with BatchNorm, most TTA methods re-estimate the normalization statistics on the test data, which brings most of their gain~\cite{schneider2020improving,wang2021tent}. In our setup, TTA thus replaces the statistics shared by all trained domains with those of the preview. Therefore, we consider two protocols, which differ in the statistics used to normalize the evaluation images.

\paragraph{Test-batch protocol.}
In the \emph{test-batch} protocol, which is common in TTA, each image is normalized with the statistics of its evaluation batch, so that its prediction depends on the other images of the batch. While each batch holds a single domain in CTTA, the domain of a test image is unknown in real-world scenarios. Therefore, each evaluation batch in our setup is drawn from the shuffled union of all evaluated sets (\appref{sec:s_impl}), and its statistics mix several domains, which can make this usual TTA strategy counter-productive.

\paragraph{Batch-independent protocol.}
Moreover, normalizing with the evaluation batch is itself a form of adaptation to the test data. In contrast, our main \emph{batch-independent} protocol forbids this for all methods. In particular, the naive combination of CL and TTA methods (\cltta), which predicts every image with the network adapted during the preview, normalizes with the statistics estimated during the preview. In that sense, no prediction depends on the rest of its batch, and mixed and single-domain batches give the same score. We report both protocols, and every method is always given the same statistics.

\subsection{Metrics}
\label{sec:metrics}
The ultimate objective is to maximize performance on all seen domains, namely the training and preview domains.

\paragraph{Score.}
Therefore, as usual in CL, we average the accuracy over all domains seen so far, which here include the preview domain.
In particular, all domains, including the preview one, are evaluated after the preview with the same model, unlike the online accuracy usually reported in CTTA.
This score measures post-preview performance rather than prediction accuracy when each image first arrives.
Online preview accuracy is reported separately in \appref{fig:stream}, and held-out preview-domain evaluation is reported in \appref{tab:heldout}.
Let $a_j(t)$ be the accuracy on $\Dte{j}$ after the preview of task $t$, and $n(t)$ the accuracy on $\Du{t+1}$ after adaptation. The score of a run is then
\begin{equation}
A = \frac{1}{T}\sum_{t=0}^{T-1}\frac{1}{t+2}\Big(\sum_{j\le t} a_j(t) + n(t)\Big).
\label{eq:score}
\end{equation}

\paragraph{Old and new accuracy.}
To disentangle stability and plasticity, we decompose the accuracy into its \emph{old} part (the trained domains) and its \emph{new} part (the preview domain).

\paragraph{Forgetting and latest-domain damage.}
As usual in CL, the \emph{forgetting} of a trained domain is the drop of its accuracy between the end of its task and the last task, averaged over the trained domains.
Finally, to measure how the preview affects freshly learned knowledge, we define the \emph{latest-domain damage} as $a_t(t)-a^\star_t$, where $a^\star_t$ is the accuracy on domain $t$ right after its training phase. In other words, this is the immediate forgetting caused by the adaptation during the preview.

\section{Limitations of existing approaches in CL-P}
\label{sec:analysis}

In this section, we analyze how existing CL and TTA methods, as well as their naive combination, behave in our setup. We describe the main trends here, while \cref{sec:experiments} introduces the benchmarks and gives the detailed results. \Cref{tab:story} summarizes these trends on ResNets. Since the conclusions hold for most TTA methods, each of its numbers is the median over the TTA methods we run that keep every core component, then averaged over six benchmarks, rather than the result of a single combination of CL and TTA methods.

\begin{table}[t]
\centering
\caption{\textbf{CL and TTA methods do not combine well out of the box.} Accuracy (\%) of a ResNet trained with FT (no buffer) or ER (buffer of 200 images), averaged over the six benchmarks of \cref{sec:exp_setup}, with mixed \evl{evaluation} batches. No TTA: the CL method alone. +TTA: the naive combination, where the network adapted by a TTA method during the \prv{preview} predicts every image. Each +TTA and \ours number is the median over the TTA methods run on each benchmark that keep every core component (up to 18, \appref{sec:s_tta_impl}). Old, new: accuracy on the trained and \prv{preview} domains. Dmg.: latest-domain damage (\cref{sec:metrics}), in points. Forg.: forgetting, the drop of a trained domain between the end of its task and the last task. Oracle: +TTA when each \evl{evaluation} batch holds a single domain and is normalized with its own statistics (two TTA methods). Trained: the \prv{preview} domain once trained on with labels.}
\label{tab:story}
\vspace{-2mm}
\small
\setlength{\tabcolsep}{3pt}
\begin{tabular}{@{}l cccc cccc@{}}
\toprule
 & \multicolumn{4}{c}{FT (no buffer)} & \multicolumn{4}{c}{ER (200 images)} \\
\cmidrule(lr){2-5}\cmidrule(lr){6-9}
 & old & new & dmg. & forg. & old & new & dmg. & forg. \\
\midrule
no TTA & 50.6 & 30.4 & -- & 34.1 & 65.9 & 43.2 & -- & 11.5 \\
+TTA & 39.3 & 56.4 & $-$27.5 & 36.6 & 42.3 & 55.7 & $-$25.2 & 32.9 \\
\textbf{\ours} & 58.6 & 55.1 & $-$0.5 & 24.6 & 65.7 & 55.1 & $-$0.2 & 11.8 \\
\midrule
\emph{Oracle} & 64.7 & 56.7 & $-$0.6 & 10.9 & 64.7 & 55.9 & $-$2.6 & 11.1 \\
\emph{Trained} & & 71.5 & & & & 71.1 & & \\
\bottomrule
\end{tabular}

\vspace{-3mm}
\end{table}

\paragraph{CL methods ignore the preview domain.}
Naturally, CL methods are never exposed to the preview domain during training, and they keep their weights unchanged during the preview phase. In that sense, they do not adapt to the preview and rely solely on their generalization capabilities when evaluated on the preview domain. As a result, CL methods alone stay far below on the preview domain what they reach once they have trained on it with labels (\cref{tab:story}). While ViTs generalize better across domains, this gap remains large (\appref{sec:s_baselines}).

\paragraph{TTA damages the trained domains.}
While various TTA strategies exist, the main idea is generally to adapt the normalization statistics (such as those of BatchNorm) or some model weights, using the unlabeled test data. In our setup, TTA is applied during the preview phase, where it updates the model on the fly for the preview domain. Specifically, when the adapted network is then used for every image with the statistics estimated on the preview, the accuracy on the preview domain rises sharply (\cref{tab:story}). While this is especially beneficial for this domain, it comes at the cost of performance on the previously learned domains, and above all on the domain just learned, while forgetting also increases. The cause is that all domains share the BatchNorm statistics (\cref{sec:eval_protocols}). TTA methods re-estimate them on the preview, and an old image normalized with the statistics of another domain is misclassified. This is usually referred to as forgetting of the source domain in CTTA~\cite{niu2022eata,wang2022cotta}, except that here it affects what the CL method has just learned. As a result, naively combining CL and TTA methods performs worse than the CL methods alone on average, even though it improves on the preview domain (see \cref{sec:main_results} for a detailed comparison).

\paragraph{Test-batch statistics are not a solution.}
Since the damage comes from the BatchNorm statistics, a naive fix is to normalize each evaluation batch with its own statistics, as TTA methods do on their test stream~\cite{schneider2020improving,wang2021tent}. Old images then bring the statistics of their own domains, which reduces the damage a lot without removing it (\appref{tab:oldnew}). However, this adapts to the evaluation data (\cref{sec:eval_protocols}), and statistics estimated from a mixed batch fit no domain in particular, so that the same model scores clearly higher with single-domain batches (\appref{tab:protocols}), whose composition cannot be chosen in practice.

\paragraph{Separate statistics would remove most of the damage.}
Interestingly, most of the damage of TTA comes from normalizing the images of every domain with the statistics of a single domain, namely those that TTA estimated on the preview. Specifically, when each evaluation batch holds a single domain and is normalized with its own statistics (Oracle), each image is normalized with the statistics of its own domain. In that case, the naive combination barely damages the domain just learned, and memory-free CL forgets about as little as a ViT, which keeps no domain statistics (Oracle in \cref{tab:story}). While test images cannot be sorted by domain in practice, the training phase and the preview of our setup each contain a single domain.

\paragraph{Limitations.}
Overall, two limitations remain. On the one hand, the domain of a test image is unknown, so that the naive combination adapts every image. On the other hand, all domains share a single network, whose statistics and parameters only fit one domain at a time. Therefore, a method for our setup should infer the domain of each image without labels, and give each image the statistics and parameters of its domain, without storing images.

\section{\ours: Dosing by Online Statistics of Each domain}
\label{sec:method}

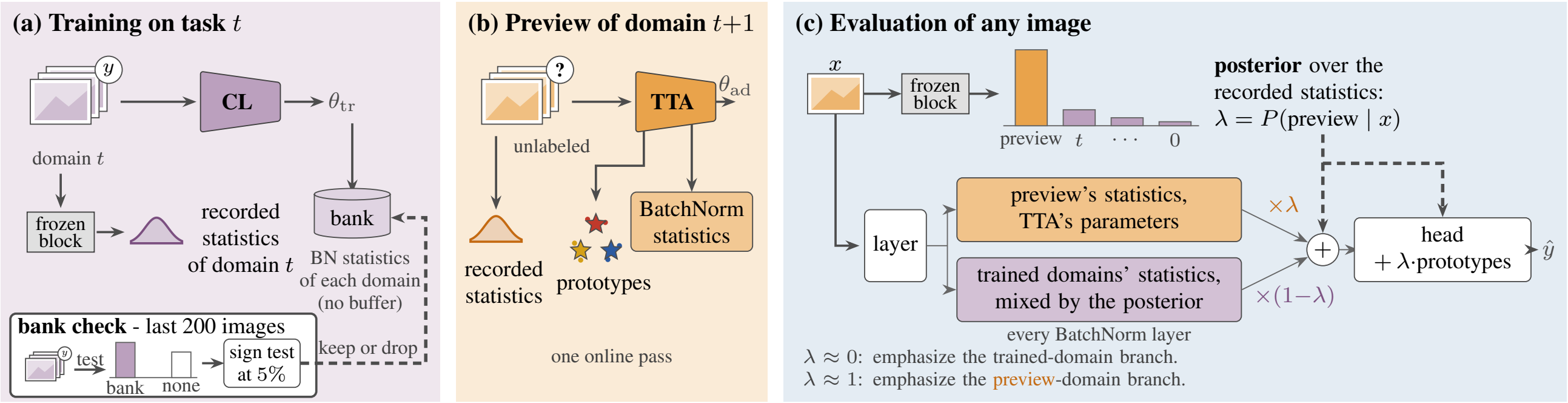
\begin{figure*}[t]
\centering
\resizebox{\textwidth}{!}{\definecolor{trainbg}{HTML}{EFE7EE}\definecolor{trainbox}{HTML}{BBA0BE}\definecolor{traindk}{HTML}{7B4F80}
\definecolor{ttabg}{HTML}{FAEBD7}\definecolor{ttabox}{HTML}{E9A34B}\definecolor{ttadk}{HTML}{C46A12}
\definecolor{dosebg}{HTML}{E4ECF3}\definecolor{dosebox}{HTML}{7E96B0}\definecolor{dosedk}{HTML}{3E5C7C}
\definecolor{protoA}{HTML}{C0392B}\definecolor{protoB}{HTML}{D4A017}\definecolor{protoC}{HTML}{2C5AA0}
\begin{tikzpicture}[font=\footnotesize, >={Stealth[length=5pt]},
  lab/.style={font=\small\bfseries, anchor=north west, inner sep=4pt},
  arr/.style={->, thick, black!70}, darr/.style={->, very thick, black!70, densely dashed},
  lbl/.style={font=\footnotesize, align=center, inner sep=1pt},
  blue/.style={font=\footnotesize, align=center, inner sep=1pt},
  note/.style={font=\scriptsize, align=center, inner sep=1pt, text=black!75},
  box/.style={draw=black!70, rounded corners=2pt, align=center, inner sep=3pt, font=\footnotesize}]
\newcommand{\dimgstack}[4]{%
  \foreach \k in {2,1,0} {\filldraw[draw=black!70, fill=white, shift={(#1+0.1*\k,#2+0.08*\k)}]
     (-0.38,-0.28) rectangle (0.38,0.28);
   \fill[#3!35, shift={(#1+0.1*\k,#2+0.08*\k)}] (-0.33,-0.23) rectangle (0.33,0.23);
   \fill[#3!70, shift={(#1+0.1*\k,#2+0.08*\k)}] (-0.33,-0.23) -- (-0.06,0.07) -- (0.1,-0.06) -- (0.33,0.14) -- (0.33,-0.23) -- cycle;}
  \node[circle, draw=black!70, fill=white, inner sep=0.3pt, font=\scriptsize\bfseries, minimum size=3.4mm] at (#1+0.6,#2+0.4) {#4};}
\newcommand{\dnet}[4]{%
  \filldraw[draw=black!75, fill=#3, rounded corners=1pt] (#1-0.5,#2-0.38) -- (#1-0.5,#2+0.38) -- (#1+0.5,#2+0.24) -- (#1+0.5,#2-0.24) -- cycle;
  \node[font=\footnotesize\bfseries] at (#1-0.04,#2) {#4};}
\newcommand{\dbell}[3]{%
  \filldraw[draw=#3, fill=#3!30, thick, smooth, samples=25, domain=-0.32:0.32, shift={(#1,#2)}]
     plot (\x,{0.32*exp(-\x*\x*24)}) -- (0.32,0) -- (-0.32,0) -- cycle;}
\newcommand{\dfrozen}[2]{%
  \filldraw[draw=black!75, fill=black!12] (#1-0.42,#2-0.25) rectangle (#1+0.42,#2+0.25);
  \node[font=\scriptsize, align=center] at (#1,#2) {frozen\\[-2pt]block};}
\begin{scope}[on background layer]
\fill[trainbg] (0,0) rectangle (5.5,5.0);
\fill[ttabg] (5.7,0) rectangle (9.6,5.0);
\fill[dosebg] (9.8,0) rectangle (19.6,5.0);
\end{scope}
\node[lab] at (0,5.0) {(a) Training on task $t$};
\node[lab] at (5.7,5.0) {(b) Preview of domain $t{+}1$};
\node[lab] at (9.8,5.0) {(c) Evaluation of any image};

\dimgstack{0.75}{3.75}{trainbox}{$y$}
\node[note] at (0.85,3.05) {domain $t$};
\dnet{3.0}{3.75}{trainbox}{CL}
\draw[arr] (1.5,3.75) -- (2.45,3.75);
\draw[arr] (3.55,3.75) -- (4.05,3.75) node[right, lbl] {$\theta_\mathrm{tr}$};
\draw[arr] (0.75,2.85) -- (0.75,2.42);
\dfrozen{0.75}{2.12}
\draw[arr] (1.2,2.12) -- (1.55,2.12);
\dbell{1.95}{1.98}{traindk}
\node[blue, anchor=west] at (2.3,2.05) {recorded\\statistics\\of domain $t$};
\draw[arr] (4.4,3.48) -- (4.4,2.6);
\node[cylinder, shape border rotate=90, aspect=0.22, draw=black!70, fill=trainbox!45, minimum width=0.95cm,
      minimum height=0.62cm, inner sep=1.5pt, font=\footnotesize] (bank) at (4.4,2.3) {bank};
\node[note] at (4.50,1.5) {BN statistics\\of each domain\\(no buffer)};
\filldraw[draw=black!70, thick, fill=white, rounded corners=2pt] (0.12,0.06) rectangle (3.95,1.1);
\node[anchor=west, inner sep=1pt, font=\footnotesize] at (0.17,0.92) {\textbf{bank check} - last 200 images};
\begin{scope}[shift={(0.5,0.4)}, scale=0.5, transform shape]
\dimgstack{0}{0}{trainbox}{$y$}
\end{scope}
\draw[arr] (0.92,0.4) -- (1.32,0.4) node[midway, above, note] {test};
\draw[black!60] (1.36,0.3) -- (2.46,0.3);
\filldraw[draw=black!60, fill=trainbox] (1.44,0.3) rectangle (1.68,0.74);
\filldraw[draw=black!60, fill=white] (2.14,0.3) rectangle (2.38,0.62);
\node[note] at (1.56,0.18) {bank};
\node[note] at (2.26,0.18) {none};
\draw[arr] (2.52,0.48) -- (2.8,0.48);
\node[box, fill=white, font=\scriptsize, inner sep=2pt] (check) at (3.27,0.48) {sign test\\at $5\%$};
\draw[darr] (check.east) -- (5.32,0.48) |- (bank.east);
\node[note, anchor=south] at (4.6,0.5) {keep or drop};

\dimgstack{6.4}{3.75}{ttabox}{?}
\node[note] at (6.9,3.2) {unlabeled};
\dnet{8.45}{3.75}{ttabox}{TTA}
\draw[arr] (7.15,3.75) -- (7.9,3.75);
\draw[arr] (8.95,3.75) -- (9.2,3.75) node[above, lbl, yshift=1pt] {$\theta_\mathrm{ad}$};
\draw[arr] (6.2,3.32) -- (6.2,2.4);
\dbell{6.2}{2.0}{ttadk}
\node[blue] at (6.3,1.5) {recorded\\statistics};
\draw[arr] (8.65,3.47) -- (8.65,2.62);
\node[box, fill=ttabox!60] at (8.65,2.25) {BatchNorm\\statistics};
\draw[arr] (8.05,3.384) |- (7.45,2.95) -- (7.45,2.5);
\foreach \a/\c in {90/protoA, 210/protoB, 330/protoC} {
  \fill[\c] ($(7.45,2.0)+(\a:0.22)+(\a+60:0.12)$) circle (0.032);
  \fill[\c] ($(7.45,2.0)+(\a:0.22)+(\a-80:0.12)$) circle (0.032);
  \node[star, star points=5, star point ratio=2.2, fill=\c, draw=black!70, inner sep=0pt, minimum size=2.6mm] at ($(7.45,2.0)+(\a:0.22)$) {};}
\node[blue] at (7.55,1.45) {prototypes};
\node[note] at (7.65,0.55) {one online pass};

\filldraw[draw=black!70, fill=white] (10.1,3.6) rectangle (10.8,4.1);
\fill[ttabox!45] (10.15,3.65) rectangle (10.75,4.05);
\fill[ttabox!80] (10.15,3.65) -- (10.4,3.9) -- (10.55,3.8) -- (10.75,3.98) -- (10.75,3.65) -- cycle;
\node[lbl, anchor=south] at (10.45,4.1) {$x$};
\dfrozen{11.7}{3.85}
\draw[arr] (10.8,3.85) -- (11.28,3.85);
\draw[arr] (12.12,3.85) -- (12.5,3.85);
\draw[black!60] (12.55,3.45) -- (15.0,3.45);
\foreach \i/\h/\c/\l in {0/0.95/ttabox/{preview}, 1/0.2/trainbox/{$t$}, 2/0.1/trainbox/{$\cdots$}, 3/0.05/trainbox/{$0$}} {
  \fill[\c, draw=black!60] (12.7+0.6*\i,3.45) rectangle ++(0.4,\h);
  \node[note] at (12.9+0.6*\i,3.27) {\l};}
\node[blue, anchor=west, text width=4.2cm, align=left] (post) at (15.15,3.85) {\textbf{posterior} over the\\recorded statistics:\\$\lambda = P(\text{preview}\mid x)$};
\draw[arr] (10.45,3.6) -- (10.45,1.95) -- (10.75,1.95);
\node[box, fill=white, minimum height=0.9cm] (conv) at (11.2,1.95) {layer};
\node[box, fill=ttabox!65, text width=3.35cm, minimum height=0.62cm] (bp) at (13.75,2.4) {preview's statistics,\\TTA's parameters};
\node[box, fill=trainbox!65, text width=3.35cm, minimum height=0.62cm] (bt) at (13.75,1.4) {trained domains' statistics,\\mixed by the posterior};
\draw[black!60] (conv.east) -- (11.85,1.95); \draw[black!60] (11.85,1.4) -- (11.85,2.4);
\draw[black!60, ->] (11.85,2.4) -- (bp.west); \draw[black!60, ->] (11.85,1.4) -- (bt.west);
\node[circle, draw=black!70, fill=white, inner sep=0pt, minimum size=4mm, font=\footnotesize] (sum) at (16.55,1.9) {$+$};
\draw[black!60, ->] (bp.east) -- (sum); \node[font=\footnotesize, text=ttadk, anchor=south] at (16.05,2.22) {$\times\lambda$};
\draw[black!60, ->] (bt.east) -- (sum); \node[font=\footnotesize, text=traindk, anchor=north] at (16.2,1.58) {$\times(1{-}\lambda)$};
\node[note] at (13.75,0.82) {every BatchNorm layer};
\draw[black!60, ->] (sum) -- (17.0,1.9);
\node[box, fill=white, text width=2.0cm] (head) at (18.05,1.9) {head\\$+\,\lambda\cdot$prototypes};
\draw[arr] (head.east) -- (19.25,1.9) node[right, lbl] {$\hat y$};
\draw[darr] (post.south -| sum.north) -- (sum.north);
\draw[darr] (16.55,2.98) -| (head.north);
\node[note, anchor=west, text width=8.6cm, align=left] at (10.0,0.4) {$\lambda\approx0$: emphasize the trained-domain branch.\\$\lambda\approx1$: emphasize the \prv{preview}-domain branch.};
\end{tikzpicture}}
\vspace{-5mm}
\caption{\textbf{\ours.} \textbf{(a)} During \trn{training}, the frozen first block of the initial network gives the recorded statistics of domain $t$, while a bank stores the BatchNorm statistics of each domain for memory-free CL. \textbf{(b)} TTA adapts online, while \ours records the statistics of the \prv{preview} and builds prototypes from the predictions of TTA. \textbf{(c)} For any image, the posterior over the recorded statistics gives $\lambda$, which weighs the branch of the \prv{preview} and the prototypes.}
\label{fig:method}
\vspace{-3mm}
\end{figure*}

In this section, we present \textbf{D}osing by \textbf{O}nline \textbf{S}tatistics of \textbf{E}ach domain (DOSE) which builds onto the analysis of Section~\ref{sec:analysis}. The main motivation is to find an efficient inference policy that allows leveraging TTA methods without interfering with the CL training scheme that can easily be plugged into the pipeline. To do so, DOSE combines a domain detector, described in \cref{sec:detector}, which infers the domain of each test image, and an adaptive model, described in \cref{sec:dosing}, which sets the statistics and parameters used for each image from those learned by CL and TTA and from the detected domain. An overview of the methodology is given in Figure~\ref{fig:method}.

\subsection{Domain detection}
\label{sec:detector}
By definition of our setup (\cref{sec:setup_def}), the training phase and the preview each hold a single domain, so that the statistics measured during each of them describe one domain. We consider two kinds of domain shift. On the one hand, low-level shifts, such as a new camera or a change of weather, change the statistics of the pixels. On the other hand, semantic shifts, such as paintings instead of photos, change the content of the images, which appears in the statistics of their features. Therefore, for each image $x$, \ours computes a vector $s(x)$ that captures both, namely the channel means and log-variances of the network input, for the pixels, and of the output of the first block (ResNet first stage, ViT patch embedding) of a \emph{frozen copy of the initial network}, for the features. The recorded statistics of a domain are the mean and variance of $s(x)$ over its training images, or over the preview stream, which form a diagonal Gaussian of a few kilobytes. The copy is frozen because the trained network changes at every task, so that statistics recorded at one task would no longer describe its features at a later task.

For a test image $x$, \ours then infers the domain from the recorded statistics. Under a uniform prior over the recorded domains (the preview, the latest domain $t$ and every older one), the posterior is
\begin{equation}
P(d\mid x)=\frac{\exp\ell_d(x)}{\sum_{d'}\exp\ell_{d'}(x)},\qquad \lambda(x)=P(\mathrm{preview}\mid x),
\end{equation}
where $\ell_d$ is the log-likelihood of $s(x)$ under the recorded statistics of domain $d$. Nothing is calibrated and no label is read. Every domain keeps its recorded statistics, since an old image that resembles neither the latest domain nor the preview must still be recognized as old.

\subsection{Dosing the adaptation}
\label{sec:dosing}
Since all domains share a single network (\cref{sec:analysis}), \ours assembles for each test image the network of its domain from what CL and TTA have learned. While the TTA method adapts online on the preview, \ours keeps the BatchNorm statistics $(\mu_\mathrm{P},\sigma^2_\mathrm{P})$ that it estimates on the stream and the affine parameters $(\gamma_\mathrm{ad},\beta_\mathrm{ad})$ that it fits under them. For an activation $h$ of $x$, each BatchNorm layer then returns
\begin{equation}
\begin{split}
&\lambda\,\mathrm{BN}\big(h;\,\mu_\mathrm{P},\sigma^2_\mathrm{P},\gamma_\mathrm{ad},\beta_\mathrm{ad}\big)\\
+\,&(1-\lambda)\,\mathrm{BN}\big(h;\,\textstyle\sum_k q_k(\mu_k,\sigma^2_k),\gamma_\mathrm{tr},\beta_\mathrm{tr}\big),
\end{split}
\label{eq:dose}
\end{equation}
where $(\gamma_\mathrm{tr},\beta_\mathrm{tr})$ are the affine parameters of $\theta_\mathrm{tr}$, $(\mu_k,\sigma^2_k)$ are the statistics of the trained domain $k$, and the weights $q_k=P(k\mid x)/(1-\lambda)$ mix them by moment matching, because old domains also differ from each other.
In \cref{eq:dose}, $\lambda=0$ selects the trained-domain BatchNorm branch, whereas $\lambda=1$ selects the preview-domain branch.
These endpoints do not generally reproduce CL alone or \cltta for the complete system because of the statistics bank, shared parameter changes, and prototype correction. Statistics and affine parameters are dosed together, because TTA fitted $\gamma_\mathrm{ad},\beta_\mathrm{ad}$ under the statistics of the preview. The dose is soft, so that no threshold is needed.
Since \ours keeps $\theta_\mathrm{tr}$ for this branch, the next task starts from $\theta_\mathrm{tr}$, so that TTA changes the predictions of \ours but never the network that the CL method keeps training (\appref{sec:s_c100c}).

\paragraph{Statistics of the trained domains.}
The second branch of \cref{eq:dose} needs the statistics of each trained domain. The latest domain uses the statistics stored in $\theta_\mathrm{tr}$ during its training phase. With a replay buffer, the training batches are mixed, so that these statistics remain a mixture of all trained domains and also serve the older ones. Without a buffer, they follow the latest domain only, while normalizing the older domains with the statistics of the preview instead, as \cltta does, makes their accuracy collapse (\appref{tab:s_bn}).
Therefore, without a buffer, \ours keeps a \emph{bank}, namely a copy of the mean and variance of every BatchNorm channel of $\theta_\mathrm{tr}$ at the end of each training phase. It stores no image, only these two numbers per channel and per trained domain.
However, the bank can become outdated. Unlike the recorded statistics, which come from a frozen network, it describes the activations of the trained network, which the CL method keeps updating after task $k$. The images of domain $k$ then no longer match the statistics stored for them, and where the recorded statistics cannot separate the domains, always keeping the bank can score below the CL method alone.
Therefore, a \emph{bank check} decides at each task whether to keep the bank. Since storing labeled images would break memory-free CL, it follows the test-then-train scheme of online CL, where each of the last 200 training images of the task is first predicted with and without the bank, then trained on. The bank is dropped only if it is significantly less accurate, under an exact one-sided sign test at $5\%$ on the images that exactly one of the two classifies correctly. No image is kept, and 200 is the size of the replay buffer of our experiments (\cref{sec:exp_setup}).

\paragraph{Parameters outside BatchNorm.}
TTA methods also change parameters outside BatchNorm, such as the LayerNorm layers~\cite{ba2016layernorm} of a ViT or the convolution weights~\cite{wang2022cotta,brahma2023petal}, which can damage the old domains. With a replay buffer, a \emph{shared-parameter check} predicts the buffer images after each preview with these changes either shared by every image or dosed like the statistics, and doses them only if this is significantly more accurate under the same test. Without a buffer, the changes are shared as in \cltta.

\paragraph{Classifier of the preview domain.}
The preview domain has no classifier of its own yet, so \ours builds class prototypes $\mu_c$ online from the stream, as the class sums of the features of the preview images weighted by their prior-corrected soft labels~\cite{saerens2002prior}. They get the weight of the branch of the preview in the head, $\log p+\lambda\,\cos(z,\mu_c)/\tau$ with $\tau=0.05$, and matter most on ViTs (\cref{tab:ablation}).

\section{Experiments}
\label{sec:experiments}

\begin{table*}[t]
\centering
\caption{\textbf{Score $A$} (\cref{eq:score}, \%) of the CL method alone (no TTA), \cltta and \ours. +TTA and Ours: medians over the TTA implementations that keep every core component of the original method (18 on ResNets and 13 on ViTs, 8 and 6 on C100-C and IN-C, \appref{sec:s_tta_impl}, every method and the strict single-pass ones in \appref{tab:s_subsets}). +TTA uses the statistics estimated during the \prv{preview}, and no method reads the \evl{evaluation} batch. Bold: best method.}
\label{tab:main}
\vspace{-2mm}
\small
\setlength{\tabcolsep}{4pt}
\begin{tabular}{@{}l ccc ccc ccc ccc@{}}
\toprule
 & \multicolumn{3}{c}{ResNet, FT} & \multicolumn{3}{c}{ResNet, ER} & \multicolumn{3}{c}{ViT-B/16, FT} & \multicolumn{3}{c}{ViT-B/16, ER} \\
\cmidrule(lr){2-4}\cmidrule(lr){5-7}\cmidrule(lr){8-10}\cmidrule(lr){11-13}
 & no TTA & +TTA & \textbf{Ours} & no TTA & +TTA & \textbf{Ours} & no TTA & +TTA & \textbf{Ours} & no TTA & +TTA & \textbf{Ours} \\
\midrule
C100-C & 48.6 & 44.0 & \textbf{60.7} & 64.7 & 46.3 & \textbf{65.9} & 65.6 & 66.5 & \textbf{67.0} & 69.0 & 69.1 & \textbf{69.5} \\
C100-TF & 49.7 & 48.5 & \textbf{66.4} & 63.1 & 51.6 & \textbf{68.9} & 73.4 & 73.8 & \textbf{74.6} & 75.2 & 75.7 & \textbf{76.4} \\
C20 & 63.7 & \textbf{64.1} & 63.9 & 68.1 & 68.1 & \textbf{68.6} & 76.9 & 77.0 & \textbf{77.9} & 79.1 & 79.2 & \textbf{79.9} \\
DomainNet & 39.1 & 37.2 & \textbf{43.5} & \textbf{51.1} & 39.3 & \textbf{51.1} & 62.8 & 62.7 & \textbf{63.4} & 63.9 & 64.0 & \textbf{64.7} \\
Inet-TF & 42.1 & 40.7 & \textbf{76.7} & 70.8 & 42.7 & \textbf{80.0} & 95.7 & 95.8 & \textbf{96.1} & 96.0 & 95.9 & \textbf{96.3} \\
IN-C & 17.1 & 15.3 & \textbf{22.3} & 31.8 & 15.9 & \textbf{34.3} & 50.7 & 52.6 & \textbf{52.8} & 56.4 & 56.7 & \textbf{57.3} \\
\midrule
Mean & 43.4 & 41.6 & \textbf{55.6} & 58.3 & 44.0 & \textbf{61.4} & 70.8 & 71.4 & \textbf{72.0} & 73.3 & 73.4 & \textbf{74.0} \\
\bottomrule
\end{tabular}

\vspace{-2mm}
\end{table*}
\subsection{Experimental setup}
\label{sec:exp_setup}
\paragraph{Benchmarks.}
Since no existing benchmark provides the three disjoint sets of each domain (\cref{sec:setup_def}), we build six benchmarks (\appref{tab:data,sec:s_data}) that cover the two kinds of shift of \cref{sec:detector}.
\emph{Low-level shifts} change the statistics of the pixels, which is what TTA was designed for. \textbf{C100-C} and \textbf{IN-C} follow the 15 corruptions of CTTA~\cite{hendrycks2019benchmarking,wang2022cotta} on CIFAR-100 and ImageNet (14 tasks), while \textbf{C100-TF} and \textbf{Inet-TF} (on Imagenette) start from clean images, followed by five strong corruptions (5 tasks).
\emph{Semantic shifts} change the content of the images, as in domain-incremental CL. \textbf{C20} changes the sub-class of each of the 20 super-classes of CIFAR-100 at every task, while \textbf{DomainNet}~\cite{peng2019domainnet} changes the style across six domains. On ViTs, we also use ImageNet-R~\cite{hendrycks2021many} and ImageNet-A~\cite{hendrycks2021nae}.

\paragraph{Backbones and CL methods.}
As backbones, we use ResNets (a ResNeXt-29 and a ResNet-50), whose BatchNorm layers are the source of the damage of \cref{sec:analysis}, and a ViT-B/16, which uses LayerNorm. The main CL methods are experience replay with a buffer of 200 images (\textbf{ER})~\cite{chaudhry2019tiny}, a strong baseline of online CL, and fine-tuning without any buffer (\textbf{FT}, memory-free), for applications that cannot store images (\appref{sec:s_backbones}). %
We select one CL learning rate per backbone on validation streams of other benchmarks, which share neither the images nor the kind of shift of the evaluated one, and keep it fixed during evaluation (\appref{sec:s_lr}), unlike online CL that adapts it~\cite{michel2026from}.
TTA settings and implementation changes are detailed in~\appref{sec:s_tta}.

\paragraph{TTA methods.}
We run 20 TTA methods using implementations adapted to \clp, with method-specific changes documented in \appref{sec:s_tta_impl}.
\Cref{tab:main,tab:story} report the median over the 18 implementations (13 on ViTs) that keep every core component of the original method, and the other analyses average Tent~\cite{wang2021tent} and ROID~\cite{marsden2024roid}, which are typical (\appref{sec:s_full}).

\paragraph{Baselines.}
We compare \ours with two baselines that use the same CL and TTA methods. The CL method alone (\emph{no TTA} in the tables) ignores the preview and predicts every image with the network $\theta_\mathrm{tr}$ that it has trained. The naive combination \cltta (\emph{+TTA} in the tables) runs the TTA method during the preview, predicts every image with the adapted network, as TTA methods are usually deployed, and starts the next task from this network, while \ours starts it from $\theta_\mathrm{tr}$ (\cref{sec:dosing}).

\paragraph{Protocol.}
Unless stated otherwise, we use the batch-independent protocol, and every evaluation batch mixes all evaluated domains (\cref{sec:eval_protocols}). All numbers are averaged over \nSeeds{} seeds, and gains are tested with a paired $t$-test (\appref{tab:signif}).

\subsection{Main results}
\label{sec:main_results}
\paragraph{The naive combination is worse than ignoring the preview.}
\Cref{tab:main} gives the scores, and \appref{sec:s_baselines} gives the numbers behind the trends of \cref{sec:analysis}. On ResNets, \cltta is below the CL method alone by $1.8$ points without memory and by $14.3$ points with a buffer (\cref{tab:main}), and it is below in \nBelowFair{} of the \nCells{} settings, although TTA raises the preview domain by up to $26$ points (\cref{tab:story}). In contrast, \ours is at least as good as the CL method alone in \nWinAlone{} of \nCells{} settings, by \gainAloneResNet{} points on ResNets and \gainAloneViT{} points on ViTs on average, and above \cltta in \nWinPlain{} settings, by \gainPlainResNet{} and \gainPlainViT{} points. Method by method, it is at least as good as \cltta in \nPairWins{} of \nPairCells{} pairs of method and setting, and the medians over TTA methods in \cref{tab:main} are never more than $0.1$ points below the CL method alone.
Individual combinations can perform substantially worse: on C100-C with ViT and CoTTA, \ours is $62.6$ points below CL alone with FT and $9.7$ points below it with ER (\appref{tab:methods_corr}). Taking the median over methods for each seed, the gain is significant ($p<0.05$, paired $t$-test) in \nSigAlone{} of \nSigCells{} settings against the CL method alone and in \nSigPlain{} against \cltta (\appref{tab:signif}).
The appendix also gives the test-batch protocol, held-out and balanced scores, other CL methods and segmentation (\appref{tab:protocols,tab:heldout,sec:s_balanced,tab:othercl,sec:s_seg}).

\paragraph{Where the gain comes from.}
\ours \emph{without a TTA method} (\appref{tab:s_wotta}) applies the recorded statistics, the bank and the prototypes to the network that TTA never adapted, which amounts on ResNets to per-domain BatchNorm with the inferred domain. It gives $3.7$ of the $7.7$ points that \ours gains over the CL method alone on ResNets (up to $20.4$ on Inet-TF without memory), and TTA adds the other $4.0$ (up to $14.2$), while on ViTs the two parts give $0.4$ and $0.5$ points.
TTA provides an additional gain on average, although the benefit depends on the method and setting.
Hard dosing with the inferred domain performs within $0.1$ points of soft dosing in our ablation (\cref{tab:ablation}).
These results support domain-specific statistics and routing as important components, but do not establish a substantial advantage of soft over hard dosing. %
TTA lowers \ours in two settings, by at most $0.4$ points. Notably, Continual Normalization~\cite{pham2022cn}, a BatchNorm-aware CL method, does not replace the recorded statistics (\appref{tab:cn}).

\paragraph{Memory-free CL gains the most.}
Without storing any image, FT with \ours reaches $76.7$ on Inet-TF and $66.4$ on C100-TF, above ER alone with its replay buffer ($70.8$ and $63.1$), since the recorded statistics and the bank give each domain its own statistics. More generally, the gain follows what TTA can learn, with $10.9$ points on corruptions and $2.2$ on the style shift of DomainNet on ResNets. On C20, whose sub-class domains the recorded statistics cannot separate, the bank check drops the bank, and \ours achieves accuracy close to the CL method alone ($63.9$ against $63.7$ without memory).

\paragraph{ViTs show smaller aggregate gains in our setting.}
On ViTs, TTA barely changes the score of the CL method alone ($+0.6$ and $+0.1$ points), while \ours gains $0.7$ to $1.0$ points on every type of shift and ties it on ImageNet-R and ImageNet-A (\appref{tab:inra}). Although TTA gains up to $15$ points on corrupted previews, the preview only weighs $0.17$ to $0.29$ of the score (\appref{sec:vitdata}). \ours runs at $840$ to $2{,}900$ images/s on ResNets (\cltta: $1{,}250$ to $4{,}300$) and $260$ to $269$ on ViTs ($265$ to $275$, \appref{tab:latency}).

\subsection{Ablations}
\label{sec:decomp}
In \Cref{tab:ablation}, we present an ablation of our proposed approach by changing one part of \ours at a time.

\paragraph{Domain detection.}
Without knowing the domain, \ours matches per-domain BatchNorm given the true domain within $0.7$ points, while keeping only the recorded statistics of the latest domain costs $8.3$ points on C100-TF and Inet-TF, since old images are then taken for the preview.

\paragraph{Adaptive model.}
Without a buffer, removing the bank costs $5.9$ points on C100-TF and $19.6$ on Inet-TF, while always keeping it, without the bank check, costs $4.9$ points on C20, which then falls below the CL method alone.

\begin{table}[t]
\centering
\caption{\textbf{Ablations.} Score of \ours, then the difference of each variant to it in points (Tent and ROID, ER and FT averaged, paired by seed). Gray: within $0.5$ points. TF/Inet: C100-TF and Inet-TF. Further ablations are reported in \appref{tab:s_ablation}.}
\label{tab:ablation}
\vspace{-2mm}
\small
\setlength{\tabcolsep}{2.5pt}
\resizebox{\columnwidth}{!}{\begin{tabular}{@{}l ccc cc@{}}
\toprule
& \multicolumn{3}{c}{ResNet} & \multicolumn{2}{c}{ViT-B/16} \\
\cmidrule(lr){2-4}\cmidrule(lr){5-6}
variant & TF/Inet & DN & C20 & TF/Inet/C20 & DN \\
\midrule
\ours (score) & 73.1 & 47.6 & 66.3 & 83.9 & 64.2 \\
\midrule
\multicolumn{6}{@{}l}{\emph{Domain detection}} \\

\quad only the latest domain's recorded statistics & $-$8.3 & $+$1.7 & \sm{$+$0.2} & \sm{$+$0.1} & \sm{$+$0.2} \\
\quad statistics recorded on $\theta_\mathrm{tr}$, every layer & $-$1.1 & \sm{$+$0.1} & \sm{$-$0.1} & \sm{$+$0.3} & \sm{$+$0.3} \\
\quad statistics recorded on the frozen last block & $-$2.0 & $+$0.7 & \sm{0.0} & \sm{$+$0.4} & \sm{$+$0.3} \\
\quad true domain given (oracle) & \sm{$-$0.3} & $-$0.7 & \sm{$+$0.2} & \sm{$+$0.5} & \sm{$+$0.4} \\
\multicolumn{6}{@{}l}{\emph{Adaptive model}} \\
\quad hard dose (per-domain BatchNorm) & \sm{0.0} & \sm{0.0} & \sm{$-$0.1} & \sm{0.0} & \sm{0.0} \\
\quad no bank & $-$6.4 & $-$1.3 & \sm{0.0} & \sm{0.0} & \sm{0.0} \\
\quad bank always, without bank check & \sm{$+$0.3} & $-$1.2 & $-$2.5 & \sm{0.0} & \sm{0.0} \\
\quad no prototype head & \sm{$-$0.2} & \sm{$-$0.2} & \sm{$-$0.3} & $-$0.6 & $-$0.6 \\
\bottomrule
\end{tabular}
}
\vspace{-3mm}
\end{table}

\section{Limitations and conclusion}
\label{sec:limits}

\paragraph{Limitations.}
This work presents an attempt at combining CL and TTA. However, the proposed setup could be expanded further. Namely, we only studied the online case, with one epoch of training and one pass over a preview of a few thousand images per task. This limitation is even more apparent with ViTs, as TTA gains increase with preview length (\appref{fig:previewlen}), or the number of adaptation updates.
Moreover, while the bank stores no image, the bank check that decides whether to use it relies on the last 200 training images of each task, which are not selected and only come from the latest domain. It can therefore drop a bank that would help the older domains, and it cannot see damage that appears on older domains only, as in memory-free segmentation.
In addition, \ours assumes that the preview only holds the new domain. This allows us to isolate domain-specific statistics, but future setups should consider multi-domain previews. Fortunately, \ours remains competitive even in such scenarios as we briefly explored in appendix \appref{sec:s_mix}.
Where domains differ by content rather than appearance, the recorded statistics cannot separate them, and \ours may offer little benefit over the CL method alone.
Finally, \ours only fixes how TTA is combined with CL, and it does not change how the CL method itself forgets. Future work should investigate how to provide an end-to-end method that does both TTA and CL.
\paragraph{Conclusion.}
In this work, we introduced CL-P, a new setup that aims to combine CL and TTA. In real-world applications, a model can see the next domain before its labels arrive. TTA methods can use this preview, but on backbones where every domain shares the BatchNorm statistics it costs the domains the CL method has just learned. \ours controls adaptation using estimated domain membership and improves aggregate performance in most evaluated settings.
Longer previews, sample selection for mixed previews and recorded statistics that see content shifts are the next steps.

\section*{LLM Usage Statement}
We used an LLM to produce part of the code that was fully reviewed. In the paper, the LLM was used to write a placeholder text in order to organize the sections but was fully rewritten. Finally, an LLM was used to format the result tables.
{
    \small
    \bibliographystyle{ieeenat_fullname}
    \bibliography{main}
}

\clearpage
\appendix
\renewcommand{\thesection}{\Alph{section}}
\setcounter{section}{0}
\onecolumn
\raggedbottom
\begin{center}{\Large\bfseries Supplementary material}\end{center}
\vspace{2mm}
\noindent This appendix compares our setup with the closest ones (\cref{sec:s_setups}), lists every component of \ours with its settings and cost (\cref{sec:s_impl}), how the baselines are run (\cref{sec:s_tta}), the selection of the CL learning rates (\cref{sec:s_lr}), the datasets (\cref{sec:s_data}) and complete results (\cref{sec:s_full}). All numbers are percentages unless stated otherwise; ``score'' is $A$ of \cref{eq:score}. The tables are generated from the runs by the released scripts.

\section{Comparison with related setups}
\label{sec:s_setups}
\Cref{tab:s_setups} compares \clp with the setups that are the closest to ours. On the one hand, CTTA, continual unsupervised domain adaptation and initially labeled streams adapt to new domains without labels, but they never receive the labels of these domains, so that no CL method trains on them afterwards. On the other hand, label delay, DoSAPP and ARC combine supervised training with unlabeled data. However, in label delay, the unlabeled data are the latest batches of the same stream, and the model is evaluated on these batches only. In DoSAPP and ARC, the unlabeled data come from classes already seen, and they serve to retain the past. Therefore, \clp is the only setup where the unlabeled data come from a whole domain that the model will train on next, and where the adapted model is evaluated on every trained domain as well as on this domain.

\begin{table}[H]
\centering
\caption{\textbf{Setups close to ours.} Labeled: data used for supervised training. Unlabeled: data the model adapts to without labels. Labels later: whether these data are labeled afterwards. Evaluation: data on which the reported accuracy is measured. Class-inc.: class-incremental.}
\label{tab:s_setups}
\footnotesize
\setlength{\tabcolsep}{4pt}
\renewcommand{\arraystretch}{1.15}
\begin{tabular}{@{}p{2.8cm}p{2.4cm}p{2.8cm}p{1.5cm}p{2.2cm}p{2.9cm}@{}}
\toprule
\raggedright Setup & \raggedright Labeled & \raggedright Unlabeled & \raggedright Labels later & \raggedright Adaptation & \raggedright Evaluation \tabularnewline
\midrule
\raggedright Domain-incremental CL~\cite{vandeven2022three} & \raggedright sequence of domains & \raggedright none & \raggedright n/a & \raggedright n/a & \raggedright all seen domains \tabularnewline
\raggedright CTTA~\cite{wang2022cotta,marsden2024roid} & \raggedright source domain only & \raggedright stream of new domains & \raggedright never & \raggedright online & \raggedright current stream \tabularnewline
\raggedright Continual UDA~\cite{saporta2022muhdi,cho2023codag,sun2025mpm} & \raggedright source domain only & \raggedright sequence of new domains & \raggedright never & \raggedright offline, several epochs & \raggedright seen domains \tabularnewline
\raggedright Initially labeled streams~\cite{dyer2014compose} & \raggedright first batch only & \raggedright drifting stream & \raggedright never & \raggedright online & \raggedright current stream \tabularnewline
\raggedright Label delay~\cite{csaba2024labeldelay} & \raggedright same stream, $d$ steps late & \raggedright latest batches of the stream & \raggedright after $d$ steps & \raggedright online & \raggedright latest batch \tabularnewline
\raggedright DoSAPP~\cite{singh2026dosapp} & \raggedright sequence of tasks (class-inc.) & \raggedright test data of current and seen tasks & \raggedright never & \raggedright unsupervised phase after each task & \raggedright all seen tasks \tabularnewline
\raggedright ARC~\cite{chen2025arc} & \raggedright sequence of tasks (class-inc.) & \raggedright test data of seen classes & \raggedright never & \raggedright at test time & \raggedright all seen classes \tabularnewline
\midrule
\raggedright \clp (ours) & \raggedright sequence of domains & \raggedright next domain, before its labels & \raggedright at the next task & \raggedright online, before training on it & \raggedright all seen domains, including the preview one \tabularnewline
\bottomrule
\end{tabular}
\end{table}

\section{Components, settings and costs of \ours}
\label{sec:s_impl}

\begin{description}\itemsep0pt
\item[CL training.] ER with a reservoir buffer of 200 images (FT: none), one epoch per task, training batch 10, replay batch 64, Adam without weight decay; learning rate $3{\cdot}10^{-5}$ (ResNet), $10^{-5}$ (ViT), selected on validation (\cref{sec:s_lr}), CODA-Prompt $10^{-3}$ (official); ViT: gradient checkpointing, inputs resized to $224$. DER++: $\alpha=0.1$, $\beta=0.5$. TTA stream batch size 100 (ResNet; DomainNet 64) or 64 (ViT); evaluation batch 100 / 64 / 32.
\item[Hardware.] The experiments ran on NVIDIA A4500, RTX 2080 SUPER, RTX 3080, RTX 5070, RTX A1000 and RTX PRO 6000 GPUs with PyTorch 2.6 to 2.8. Systems that are compared share the configuration and the seeds of their runs.
\item[Preview pass.] The TTA method is re-initialized from $\theta_\mathrm{tr}(t)$ and run online, one pass over the shuffled preview, with its official optimizer (\cref{sec:s_tta}).
\item[Recorded statistics (domain posterior).] Frozen copy of the \emph{initial} network, pruned to its first block (ResNet first stage, ViT patch embedding). Per image, the moment vector $s(x)$: channel means and log-variances of the network input and of the output of this block. The recorded statistics of a domain are the running mean and variance of $s(x)$ (parallel update of Chan \etal) over its images: a trained domain's over its last $1000$ training images, accumulated during its training stream with a no-gradient forward; the preview's over the preview stream, on the forward the prototype step already runs, prequentially. Posterior: diagonal Gaussians, uniform prior over the preview, the latest domain and every older domain (\texttt{--dose-rule posterior --dose-sensors stem0 --dose-domains all}).
\item[Dosing (BatchNorm).] \Cref{eq:dose} in every BatchNorm layer, per image: the preview's statistics (exponential average during the preview, momentum $0.1$) with the adapted affine parameters, against the trained domains' statistics mixed by moment matching (mixture mean and variance) with $\theta_\mathrm{tr}$'s affine parameters. Trained domains' statistics: latest domain, $\theta_\mathrm{tr}$'s stored statistics; older domains with a buffer, the current stored statistics; without a buffer, the bank: $\theta_\mathrm{tr}$'s stored statistics at the end of each domain's task, two numbers per channel (0.2 MB per domain on a ResNet-50) (\texttt{--dose-split bn --dose-bank verify}).
\item[Dosing (head and other changes).] Class prototypes built online during the preview ($C\times(d{+}1)$ numbers, prior-corrected soft labels), added as $\log p+\lambda\cos(z,\mu_c)/\tau$, $\tau=0.05$, on every network. Changes of the TTA method outside BatchNorm: shared, or dosed per image after the shared-parameter check (module by module: convolutions and linear layers by their output interpolation, LayerNorm and GroupNorm by the interpolation of the two affine maps; otherwise the log-probabilities of $\theta_\mathrm{tr}$ and of the dosed network mixed with $\lambda$) (\texttt{--dose-shared verify}).
\item[Bank check and shared-parameter check (online).] No image outlives its stream; both checks use 200 images, the size of ER's buffer (\texttt{--dose-check online --dose-check-images 200}). \emph{Bank} (memory-free CL, tasks $t\ge1$): during the last 200 images of domain $t$'s training stream, each batch is scored before the optimizer step that trains on it, in eval mode, by the current network with every BatchNorm normalizing with the posterior mixture of the trained domains' statistics (the recorded statistics of the latest domain, built on this stream, against those of the older domains; the latest domain's statistics are the stored ones, an older domain's come from the bank) and with the stored statistics alone; two hit counts are kept, and the decision applies to the task's deployment. \emph{Shared changes} (when the TTA method changed tensors outside BatchNorm): after the preview, the deployed network with those changes shared or dosed scores the replay buffer (its images in storage order, at most 200); without a buffer the changes are shared. An option is dropped (the bank) or switched (dosed) only if the alternative is significantly more accurate: exact one-sided sign test on the discordant images, level $5\%$. The decisions hold for the task.
\item[Test-batch protocol.] Single-domain batches: every BatchNorm layer on the batch's statistics with the adapted affine parameters (\texttt{--route-adapted-bn batch --route-trained-bn full}). Mixed batches, memory-free: $\theta_\mathrm{tr}$'s branch normalizes each old image with a flat-top kernel average of its batch neighbours' moments (bandwidth: median distance to the $B/8$-th nearest neighbour), shrunk towards the pooled old images with a pseudo-count of one batch, and with the latest domain's stored statistics in proportion of its posterior (\texttt{--dose-protocol official}).
\item[$\theta_\mathrm{tr}$.] Stored as a diff of the tensors the TTA method changed (exact; tested), including the BatchNorm statistics that Tent, EATA, SAR and ROID strip; a full copy when the structure changes (RoTTA, NOTE, ViDA, DePT, DPCore, DAS) or a container owns a changed tensor (CoTTA's ViT position embedding).
\item[Next task.] Training resumes from $\theta_\mathrm{tr}$, the network before TTA, which \ours keeps for its trained branch (\texttt{--train-from snapshot}). From the adapted network, as in \cltta, is an ablation, which changes the score by at most $0.4$ points on the benchmarks of \cref{tab:ablation}, but costs up to $14.1$ points on C100-C without memory (\cref{sec:s_c100c}). Conversely, giving \cltta the same start does not close the gap. We also scored \cltta reset to $\theta_\mathrm{tr}$ after each preview, so that the CL method trains exactly as alone and \cltta only changes the network it deploys. Over the $950$ runs behind \cref{tab:main} in which this variant was logged ($525$ on ResNets and $425$ on ViTs, not logged for Norm, ReservoirTTA and some methods with a buffer), it scores $45.2$ on ResNets and $67.2$ on ViTs, against $45.4$ and $69.0$ for \cltta from its adapted network and $58.3$ and $70.1$ for \ours, which is above it in $90\%$ of these runs. In the test-batch protocol, \ours uses the same test-batch statistics as the baseline (item above).
\item[Overhead at test time.] One pass through the frozen first block per image, the per-image mixing of every BatchNorm layer, and, when the shared changes are dosed and cannot be split by module, a second forward pass of $\theta_\mathrm{tr}$. During the preview: one extra forward pass per batch (recorded statistics, BatchNorm statistics and prototypes on the same forward).
\end{description}

\section{How the baselines are run}
\label{sec:s_tta}
We implement the TTA methods using their official code where available and their published descriptions otherwise.
Optimizer settings are listed in~\cref{tab:s_tta}, and implementation changes are detailed in~\cref{sec:s_tta_impl}. The 20 methods are Tent, EATA, SAR, ROID, CoTTA, RDumb, RoTTA, PETAL, NOTE, LAW, CMF, Surgical, PALM, LCoTTA, POEM, DePT, DPCore, ViDA, DAS, and Norm, and we also run ReservoirTTA around Tent, EATA, SAR, and ROID. %
The main tables report the median over the implementations that keep every core component of the original method, namely all of them except PETAL, ViDA, DePT, DPCore, DAS and POEM (\cref{sec:s_tta_impl}), and \cref{tab:s_subsets} gives it over every method. The adapted network is deployed as in the official code, so Tent, EATA, SAR, ROID, and CoTTA normalize with the statistics of the test batch, and no method gets recalibrated statistics. DER++ replays logits as in the original paper. CODA-Prompt uses the official prefix-tuning prompts without a buffer, with a frozen pretrained ViT and prompt components learned per domain.

\subsection{Implementation details of the TTA methods}
\label{sec:s_tta_impl}
Each implementation is integrated into one module of our framework and checked against available official code or the published equations. The algorithmic code is unchanged, except for the changes below. SAR, Tent, Norm, LAW, and PALM need none of the method-specific ones.

The official code assumes a single adaptation phase on a fixed source model. Four changes are therefore common to all methods.
\begin{description}\itemsep0pt
\item[Reset.] At each preview, the method restarts from the network that the CL method has just trained, $\theta_\mathrm{tr}(t)$. Its source, anchor, and teacher copies, its optimizer state, and its memories are re-initialized from this network.
\item[Source data.] Some methods read source data: EATA for its Fisher information, POEM for its protector, and ReservoirTTA for its style statistics. They read, without labels, the training images of the domain the network was just trained on, at the start of each preview. These additional source-data accesses are exceptions to a strict single-pass protocol.
\item[Next task.] When the next task keeps the adaptation (\cltta, and \ours in the ablation of \cref{tab:s_thetatr}), it starts from the network that the method deploys. For CoTTA, PETAL, RoTTA and ViDA, this is the teacher and not the internal student. On a ViT, CoTTA's student collapses during a preview while its teacher does not, and starting from the student would train the collapse into the CL model.
\item[Class count.] Some entropy margins are hard-coded for the 1000 classes of ImageNet ($0.4\ln 1000$). This is above the maximum entropy of 10 to 100 classes and would disable the filter, so we set them to $0.4\ln C$ (EATA, RDumb, LCoTTA and POEM). The redundancy margins of EATA and RDumb use the official value for the class count ($0.4$ for 10 classes, $0.1$ for 100 and $0.05$ for 1000), log-interpolated to $0.264$ for the 20 classes of C20.
\end{description}
Method-specific changes:
\begin{description}\itemsep0pt
\item[RDumb.] The official code is ResNet-only. On a ViT, our port adapts the LayerNorm affine parameters, as Tent and EATA do. Its reset target is the network at the start of each preview.
\item[ROID, CMF.] Our ROID normalizes by the actual batch size instead of the configured one, and builds its augmentations at the true input resolution. CMF shares the same port and loss code, and its DomainNet WeightNorm branch is not used.
\item[PETAL.] No SWAG checkpoint exists for our backbones. Its mean model is therefore the network before the preview and its covariance is one, so its penalty becomes $\tfrac12\lambda\|\theta-\theta_\mathrm{anchor}\|^2$, which is negligible at the official $\lambda=10^{-8}$. We also clamp a quantile index where the official code reads past the end of an array when the percentile is $1$.
\item[RoTTA.] Its robust BatchNorm replaces BatchNorm at the first adaptation step, because doing it at construction would erase the pretrained statistics before CL training. It is converted back to standard BatchNorm before the next task, with the adapted statistics and affine parameters.
\item[NOTE.] The source model is trained with instance-aware BatchNorm from the start, as NOTE requires. The official code predicts one image at a time. We predict a batch at once in evaluation mode, which gives the same predictions, and insert its images in the memory in order.
\item[Surgical.] It adapts the first block, the choice of its paper for input-level shifts.
\item[LCoTTA.] The subspace is updated at every batch. The official 50 or 100 batches assume domains of about 780 batches, while ours have 8 to 20, so the subspace would never be built.
\item[ViDA.] No ViDA checkpoint pre-trained on the source exists for our backbones, so the adapters start from zero. Its numbers are therefore not directly comparable to those of its paper.
\item[DPCore.] The prompt is added to the first patch tokens instead of prepended, so that it is the identity at zero. Its source statistics come from the training images of the latest domain.
\item[DePT.] No official code is released. Our port follows the paper: prompt tuning on a frozen ViT, pseudo-labels from memory-bank prototypes, and a filter on agreement and confidence. An L2 pull of the prompt stands in for its hierarchical regularizer.
\item[DAS.] It is run on ViTs only, as in the official configurations. Its labeled source batch is drawn from the replay buffer of the CL method, the only labeled data the setup holds, so it runs with ER only. The official method draws it instead from a knowledge base of class images generated beforehand with Stable Diffusion, which our setup does not have.
We use fp32 instead of mixed precision.
\item[ReservoirTTA.] It wraps Tent, EATA, SAR and ROID as in the official variants. Its style threshold comes from at least 1{,}000 batch styles of the training images of the latest domain, and the reservoir is rebuilt at each preview.
\item[POEM.] Its protector uses the entropies of the unadapted network on 2{,}000 unlabeled training images of the latest domain, while the official code holds out $25\%$ of a source validation set.
\end{description}
For \ours, the hooks that record the statistics are re-attached to the network a method deploys before each recorded forward pass, because PETAL and RoTTA create their teacher at their first adaptation step.

\begin{table}[H]
\centering
\caption{Optimizers (SGD: momentum $0.9$; $b$ = batch size). ImageNet-scale: Imagenette, DomainNet, ImageNet-R/A. The other hyperparameters (margins, thresholds, augmentations: 32 views, 8 on 224-pixel datasets) are the official ones; \texttt{src/tta/factory.py} cites the source of each value. Tent on CIFAR: $10^{-3}$, as in the CoTTA code~\cite{wang2022cotta}, while the PALM benchmark configuration~\cite{maharana2025palm} uses $2{\cdot}10^{-5}$.}
\label{tab:s_tta}
\small
\setlength{\tabcolsep}{4pt}
\begin{tabular}{@{}lcc@{}}
\toprule
Method & CIFAR datasets & ImageNet-scale datasets \\
\midrule
Tent & Adam $10^{-3}$ & SGD $2.5{\cdot}10^{-4}$ \\
EATA, SAR, ROID, CMF & Adam $10^{-3}$ & SGD $2.5{\cdot}10^{-4}$ \\
CoTTA & Adam $10^{-3}$ & SGD $10^{-2}$ \\
RoTTA, PETAL & Adam $10^{-3}$ & Adam $10^{-3}$ \\
LAW, PALM, NOTE & Adam $5{\cdot}10^{-5}$, $5{\cdot}10^{-4}$, $10^{-4}$ & same \\
LCoTTA & SGD $1.5{\cdot}10^{-4}$ (ViT $5{\cdot}10^{-4}$) & same \\
ViDA & Adam $3{\cdot}10^{-4}$ (adapters $10^{-6}$) & SGD $5{\cdot}10^{-7}$ ($2{\cdot}10^{-7}$) \\
DPCore, DePT, DAS & $10^{-2}$, $10^{-3}$, Adam $10^{-5}$ & same \\
Tent, EATA, SAR on ViT & \multicolumn{2}{c}{SGD $0.001\,b/64$ for every dataset} \\
\bottomrule
\end{tabular}
\end{table}

\section{Selection of the CL learning rates}
\label{sec:s_lr}
No test image is used. For every domain, $10\%$ of its training images are held out and never trained on; those of domain $t$ play the role of $\mathcal{D}^\mathrm{te}_t$ and those of domain $t{+}1$ of the preview. The last task is dropped. Learning rates can also be tuned online during the stream, for instance with hypergradients~\cite{michel2026from}, which we do not use. %
We run the CL method alone for each rate with seed 0 (\cref{tab:s_lr}). The rate of each benchmark is the best average validation score over other benchmarks, which share neither its images nor its kind of shift (\cref{tab:s_lr_dev}), so that its own domains take no part in the choice, and it is then fixed during evaluation. Every benchmark gets $3{\cdot}10^{-5}$ on ResNets and $10^{-5}$ on ViTs, which is also the best average over all five benchmarks.

\begin{table}[H]
\centering
\caption{CL learning rate, the CL method alone, seed 0, validation score. Selected: $3{\cdot}10^{-5}$ (ResNet), $10^{-5}$ (ViT). Validation accuracies on the CIFAR benchmarks are inflated because the RobustBench ResNet was trained on the CIFAR-100 \trn{training} images from which the held-out images come.}
\label{tab:s_lr}
\small
\setlength{\tabcolsep}{3pt}
\begin{tabular}{@{}lcccc@{}}
\toprule
ResNet, validation & $10^{-5}$ & $3{\cdot}10^{-5}$ & $10^{-4}$ & $3{\cdot}10^{-4}$ \\
\midrule
CIFAR-100-C   & \textbf{89.4} & 88.3 & 77.2 & 50.0 \\
CIFAR-100-TF  & \textbf{93.0} & 91.3 & 82.5 & 58.6 \\
CIFAR-20      & 68.4 & \textbf{72.7} & 70.1 & 58.3 \\
Imagenette-TF & 92.1 & \textbf{93.0} & 91.3 & 82.0 \\
DomainNet     & 41.4 & \textbf{48.6} & 42.9 & 29.3 \\
\midrule
average       & 76.9 & \textbf{78.8} & 72.8 & 55.6 \\
\bottomrule
\end{tabular}
\hspace{8mm}
\begin{tabular}{@{}lcccc@{}}
\toprule
ViT, validation & $3{\cdot}10^{-6}$ & $10^{-5}$ & $3{\cdot}10^{-5}$ & $10^{-4}$ \\
\midrule
CIFAR-100-C   & 52.4 & \textbf{65.9} & 63.5 & 46.7 \\
CIFAR-100-TF  & 75.5 & \textbf{80.8} & 75.2 & 60.3 \\
CIFAR-20      & \textbf{79.2} & 78.6 & 73.1 & 60.1 \\
DomainNet     & 53.3 & \textbf{61.5} & 59.2 & 42.6 \\
Imagenette-TF & \textbf{100.0} & 99.5 & 98.4 & 91.8 \\
\midrule
average       & 72.1 & \textbf{77.3} & 73.9 & 60.3 \\
\bottomrule
\end{tabular}
\end{table}

\paragraph{Separate development domains.} A selection that only leaves the evaluated benchmark out gives the same rates (\cref{tab:s_lr_dev}). %
The development benchmarks are the other benchmarks of this paper, scored on their held-out validation images, and no benchmark outside the paper was used. On ResNets, the rate is shared by the ResNeXt-29 of the CIFAR benchmarks and the ResNet-50 of the others, so that C100-C and C100-TF get the rate selected with a ResNet-50 on DomainNet.

\begin{table}[H]
\centering
\caption{CL learning rate selected for each benchmark on separate development benchmarks only (best average validation score of \cref{tab:s_lr}). Development benchmarks share neither the images nor the kind of shift of the evaluated one. Leave one out: every other benchmark.}
\label{tab:s_lr_dev}
\small
\setlength{\tabcolsep}{4pt}
\begin{tabular}{@{}llcccc@{}}
\toprule
& & \multicolumn{2}{c}{ResNet} & \multicolumn{2}{c}{ViT-B/16} \\
\cmidrule(lr){3-4}\cmidrule(lr){5-6}
Evaluated & Development benchmarks & separate & leave one out & separate & leave one out \\
\midrule
C100-C, C100-TF & DomainNet & $3{\cdot}10^{-5}$ & $3{\cdot}10^{-5}$ & $10^{-5}$ & $10^{-5}$ \\
C20 & Imagenette-TF, DomainNet & $3{\cdot}10^{-5}$ & $3{\cdot}10^{-5}$ & $10^{-5}$ & $10^{-5}$ \\
Inet-TF, IN-C & CIFAR-20, DomainNet & $3{\cdot}10^{-5}$ & $3{\cdot}10^{-5}$ & $10^{-5}$ & $10^{-5}$ \\
DomainNet & the four others & $3{\cdot}10^{-5}$ & $3{\cdot}10^{-5}$ & $10^{-5}$ & $10^{-5}$ \\
\bottomrule
\end{tabular}
\end{table}

\section{Backbones and CL methods}
\label{sec:s_backbones}
On C100-TF and C100-C we use the AugMix ResNeXt-29~\cite{hendrycks2020augmix} from RobustBench~\cite{croce2021robustbench}. On Inet-TF, DomainNet, IN-C and C20 we use the torchvision ImageNet ResNet-50~\cite{he2016resnet}, with CIFAR images upsampled to 112 pixels. C20 needs a backbone that has never seen the fine classes of CIFAR-100. The ViT is a ViT-B/16~\cite{dosovitskiy2021vit} from timm.
The CL methods are experience replay with a buffer of 200 images (\textbf{ER}), fine-tuning without any buffer (\textbf{FT}, memory-free), DER++~\cite{buzzega2020derpp} and CODA-Prompt~\cite{smith2023coda}. Each task is trained for one epoch with Adam. The learning rates, $3{\cdot}10^{-5}$ for ResNets and $10^{-5}$ for ViTs, are selected on validation, and CODA-Prompt keeps its own $10^{-3}$.

\section{Datasets}
\label{sec:s_data}
We use six domain-incremental benchmarks with a fixed label space.
\textbf{C100-TF} and \textbf{Inet-TF} (Imagenette) have 5 tasks each: a clean domain followed by corrupted ones, generated once per image with the official code for Cifar100-C and ImageNet-C~\cite{hendrycks2019benchmarking} at severity 5.
\textbf{C100-C} (14 tasks) uses the 15 corruptions of CIFAR-100-C. Each domain has its own fifteenth of the CIFAR-100 training images, corrupted with the official generator. Its preview and evaluation images are disjoint halves of the released test images, which every corruption reuses as in all CIFAR-100-C benchmarks.
\textbf{C20} (4 tasks) uses the 20 super-classes of CIFAR-100. Each task uses a different fine class of every super-class, so the shift is semantic rather than statistical.
\textbf{DomainNet}~\cite{peng2019domainnet} (5 tasks) uses the 100 most frequent classes over clipart, infograph, painting, quickdraw, real and sketch, a shift of style.
\textbf{IN-C} (14 tasks) uses the 15 corruptions of ImageNet-C at severity 5.
On ViTs we also use ImageNet-R~\cite{hendrycks2021many} and ImageNet-A~\cite{hendrycks2021nae} (200 classes), whose test sets are about ten times smaller.
\Cref{tab:data} gives the number of images per task. The training images of a task are labeled and seen once. The preview is the unlabeled stream that TTA adapts to, and the evaluation set holds the held-out images of a domain. Training, preview and evaluation images never overlap, under any corruption.

\paragraph{Data splits.}
For each domain, training, preview and evaluation images form three separate sets. In particular, the preview stream of domain $t{+}1$ and the labeled training set of task $t{+}1$ are two disjoint sets of images of the same domain. The next task thus trains on the preview domain, but never on the preview images themselves, so that their evaluation does not depend on what the next task learns.

\begin{table}[H]
\centering
\caption{\textbf{Images per task.} \trn{Training} images are labeled, \prv{preview} images are not, and the \evl{evaluation} set of a domain is scored at every later task. ImageNet-R and ImageNet-A (ViT only) also add a corruption at each task. DomainNet caps each class, and some classes have fewer images in some domains.}
\label{tab:data}
\vspace{-2mm}
\small
\setlength{\tabcolsep}{4pt}
\begin{tabular}{@{}l l cc rr rr@{}}
\toprule
 & shift & tasks & classes & \multicolumn{2}{c}{training} & preview & eval. \\
\cmidrule(lr){5-6}
 & & & & images & per class & & \\
\midrule
C100-TF & corruption & 5 & 100 & 10{,}000 & 100 & 1{,}000 & 1{,}000 \\
Inet-TF & corruption & 5 & 10 & ${\approx}$1{,}900 & ${\approx}$190 & ${\approx}$390 & ${\approx}$390 \\
C100-C & corruption & 14 & 100 & 3{,}333 & 33 & 5{,}000 & 2{,}000 \\
IN-C & corruption & 14 & 1{,}000 & 25{,}000 & 25 & 20{,}000 & 5{,}000 \\
C20 & sub-class & 4 & 20 & 10{,}000 & 500 & 1{,}000 & 1{,}000 \\
DomainNet & style & 5 & 100 & ${\le}$10{,}000 & ${\le}$100 & ${\le}$2{,}000 & ${\le}$1{,}000 \\
\midrule
ImageNet-R & rendition & 5 & 200 & 4{,}800 & 24 & 600 & 600 \\
ImageNet-A & adversarial & 5 & 200 & 1{,}200 & 6 & 150 & 150 \\
\bottomrule
\end{tabular}
\vspace{-3mm}
\end{table}

\textbf{C100-TF, Inet-TF}: the clean data followed by corrupted copies (gaussian noise, defocus blur, snow, brightness and contrast for C100-TF, and gaussian noise, defocus blur, brightness, contrast and shot noise for Inet-TF); each image of a domain is corrupted once with the official CIFAR-C / ImageNet-C generator at severity 5 and cached; the images of a domain are a fixed per-class chunk of the data, disjoint across domains and across training, preview and evaluation. \textbf{C100-C}: 14 tasks; domain $d$ of the training set is the CIFAR-100 training images whose position in their class is $d \bmod 15$ ($\approx 3{,}333$ images), corrupted by the official generator; evaluation: even indices of the released CIFAR-100-C test set of corruption $t$; preview: odd indices of corruption $t{+}1$ (the same 5{,}000 images under every corruption, as in every CIFAR-C CTTA benchmark; an evaluated image is never in a stream). \textbf{C20}: task $i$ trains on fine-class slot $i$ of every super-class and the preview shows slot $i{+}1$. \textbf{DomainNet}: the 100 classes with most images over the six domains, official train/test splits, 100 training, 10 evaluation and 20 preview images per class and domain. \textbf{ImageNet-R / ImageNet-A} (ViT only): 200 classes, every fifth image per class held out for evaluation and preview, the rest for training, cut into 5 consecutive chunks, each chunk corrupted (domain 0 clean) as for Inet-TF; we report them on ViTs only, since the ResNet at 112 pixels reaches only $2$ to $19\%$ on them.

\section{Complete results}
\label{sec:s_full}

\paragraph{Standard deviations and seeds.}
All numbers are means over \nSeeds{} seeds. For the entries of \ours in \cref{tab:main}, the standard deviation over seeds (of the per-seed median over TTA methods) has a median of \medStd{} and a maximum of \maxStd{} points (\maxStdWhere).

\subsection{Why the baselines fall short}
\label{sec:s_baselines}
This subsection gives the numbers behind the trends of \cref{sec:analysis}, from \cref{tab:story,tab:oldnew}. On ResNets, the CL methods alone only reach $30.4\%$ on the preview domain without memory and $43.2\%$ with a buffer, while the same domain reaches $71$ to $72\%$ once it has been trained on with labels (\cref{tab:story}). While ViTs generalize better across corruptions, the gap still remains $18$ points ($63$ to $65\%$ against $82\%$, \cref{tab:oldnew}). With TTA, the accuracy on the preview domain rises to $56\%$ ($+26.0$ and $+12.5$ points), but the old domains drop by $11$ and $24$ points, and the domain just learned by $25$ to $28$ points (up to $65$ on Inet-TF). Forgetting also rises, from $34.1$ to $36.6$ points without memory and from $11.5$ to $32.9$ with a buffer. This forgetting is measured on the deployed network at the last task, so that it includes the damage of the adapted network on the old domains. Before TTA, the network forgets $32.9$ points without memory and $9.0$ with a buffer, against $33.5$ and $12.2$ for the CL method alone (seed 0 of Tent, SAR, EATA and ROID on the six benchmarks).
With test-batch statistics, the damage falls to about $8$ points without memory and $4$ with a buffer, and with a buffer the old domains remain below the CL method alone ($62.6$ against $65.9$). In contrast, on single-domain test batches (Oracle), \cltta only loses $0.6$ points on the domain just learned without memory and $2.6$ with a buffer, and the forgetting of memory-free CL drops from $34$ to $10.9$ points, close to that of a ViT ($11.6$).

\subsection{Every benchmark}
\Cref{tab:oldnew} splits the score of every benchmark into its old part, its new part and the latest-domain damage, and \cref{tab:signif} tests the gains of \cref{tab:main} over seeds. Both take the median over the TTA methods of \cref{tab:main}, as \cref{tab:main} and \cref{tab:story} do.

\begin{table}[H]
\centering
\caption{\textbf{Old domains, \prv{preview} domain and latest-domain damage, every benchmark} (accuracy and points, mixed batches, median over the TTA methods of \cref{tab:main}, per column; ``--'': the two protocols coincide on ViTs). +TTA: \cltta with its own statistics (batch-independent); +TTA$^\mathrm{batch}$: with the test batch's (test-batch protocol; on ViTs both coincide). \Cref{tab:story} averages the ResNet rows.}
\label{tab:oldnew}
\vspace{-2mm}
\small
\setlength{\tabcolsep}{3pt}
\fitwidth{\begin{tabular}{@{}ll cccc cccc cccc@{}}
\toprule
 & & \multicolumn{4}{c}{Old domains} & \multicolumn{4}{c}{Preview domain} & \multicolumn{4}{c}{Latest-domain damage} \\
\cmidrule(lr){3-6}\cmidrule(lr){7-10}\cmidrule(lr){11-14}
 & & no TTA & +TTA & {\scriptsize +TTA$^\mathrm{batch}$} & \textbf{Ours} & no TTA & +TTA & {\scriptsize +TTA$^\mathrm{batch}$} & \textbf{Ours} & no TTA & +TTA & {\scriptsize +TTA$^\mathrm{batch}$} & \textbf{Ours} \\
\midrule
ResNet, FT & C100-C & 49.1 & 40.0 & 57.8 & 60.1 & 45.8 & 68.7 & 63.3 & 66.5 & 0 & $-$22.0 & $-$9.4 & $-$0.6 \\
 & C100-TF & 59.6 & 41.5 & 64.9 & 67.5 & 32.7 & 69.2 & 61.2 & 67.9 & 0 & $-$31.7 & $-$6.1 & $-$0.2 \\
 & C20 & 75.9 & 76.0 & 76.2 & 75.7 & 43.8 & 44.7 & 44.6 & 44.7 & 0 & 0 & 0 & $-$0.3 \\
 & DomainNet & 47.4 & 39.6 & 51.3 & 48.7 & 24.4 & 39.7 & 38.6 & 37.5 & 0 & $-$20.5 & $-$5.3 & $-$1.3 \\
 & Inet-TF & 53.7 & 26.9 & 75.1 & 79.4 & 19.3 & 75.1 & 62.8 & 74.8 & 0 & $-$64.7 & $-$9.1 & $-$0.1 \\
 & IN-C & 18.2 & 11.7 & 21.0 & 20.5 & 16.2 & 40.7 & 32.6 & 39.2 & 0 & $-$26.4 & $-$17.5 & $-$0.8 \\
\midrule
ResNet, ER & C100-C & 65.8 & 42.7 & 62.1 & 65.6 & 57.4 & 67.5 & 63.8 & 65.1 & 0 & $-$16.5 & $-$3.1 & $-$0.1 \\
 & C100-TF & 71.1 & 46.1 & 68.6 & 70.7 & 49.2 & 68.0 & 60.8 & 66.9 & 0 & $-$26.5 & $-$2.6 & 0 \\
 & C20 & 81.1 & 80.8 & 80.6 & 81.3 & 46.0 & 45.9 & 45.6 & 47.1 & 0 & $-$0.8 & $-$0.9 & $-$0.1 \\
 & DomainNet & 59.5 & 42.1 & 55.7 & 58.6 & 38.1 & 40.3 & 40.2 & 40.6 & 0 & $-$21.4 & $-$4.8 & $-$0.8 \\
 & Inet-TF & 83.5 & 29.3 & 80.6 & 83.5 & 48.1 & 75.1 & 65.4 & 75.5 & 0 & $-$62.6 & $-$4.4 & 0 \\
 & IN-C & 34.4 & 12.8 & 27.7 & 34.3 & 20.4 & 37.3 & 31.2 & 35.2 & 0 & $-$23.4 & $-$6.5 & $-$0.2 \\
\midrule
ViT-B/16, FT & C100-C & 65.9 & 66.3 & -- & 66.7 & 64.2 & 68.2 & -- & 68.5 & 0 & $-$0.6 & -- & $-$0.2 \\
 & C100-TF & 78.4 & 78.4 & -- & 78.4 & 67.0 & 68.2 & -- & 69.9 & 0 & $-$0.1 & -- & 0 \\
 & C20 & 89.6 & 89.6 & -- & 89.4 & 54.1 & 54.6 & -- & 56.8 & 0 & 0 & -- & 0 \\
 & DomainNet & 69.8 & 69.8 & -- & 69.5 & 53.7 & 53.6 & -- & 56.1 & 0 & 0 & -- & $-$0.3 \\
 & Inet-TF & 97.5 & 97.4 & -- & 97.3 & 93.0 & 93.2 & -- & 94.3 & 0 & 0 & -- & 0 \\
 & IN-C & 51.6 & 51.9 & -- & 52.0 & 48.7 & 58.3 & -- & 56.0 & 0 & $-$2.2 & -- & $-$1.5 \\
\midrule
ViT-B/16, ER & C100-C & 69.5 & 69.3 & -- & 69.5 & 66.1 & 69.0 & -- & 69.5 & 0 & $-$0.3 & -- & $-$0.1 \\
 & C100-TF & 80.5 & 80.5 & -- & 80.5 & 68.0 & 69.1 & -- & 71.4 & 0 & $-$0.1 & -- & 0 \\
 & C20 & 91.9 & 92.1 & -- & 91.8 & 55.9 & 56.1 & -- & 60.4 & 0 & 0 & -- & 0 \\
 & DomainNet & 71.2 & 71.3 & -- & 71.1 & 53.7 & 54.2 & -- & 57.0 & 0 & 0 & -- & $-$0.2 \\
 & Inet-TF & 97.6 & 97.7 & -- & 97.7 & 93.7 & 93.1 & -- & 94.1 & 0 & 0 & -- & 0 \\
 & IN-C & 57.9 & 56.5 & -- & 57.2 & 50.2 & 58.7 & -- & 56.9 & 0 & $-$1.9 & -- & $-$1.4 \\
\bottomrule
\end{tabular}
}
\vspace{-2mm}
\end{table}

\begin{table}[H]
\centering
\caption{\textbf{Paired differences over seeds} (batch-independent protocol, mixed batches). For each seed, the median over the TTA methods of \cref{tab:main} of the gain of \ours over the CL method alone and over \cltta, then its mean over seeds and, in brackets, the $95\%$ confidence interval of this mean (Student $t$ with $n-1$ degrees of freedom). Each mean is over $n=5$ seeds. $^{*}$: $p<0.05$, one-sided paired $t$-test.}
\label{tab:signif}
\vspace{-2mm}
\small
\setlength{\tabcolsep}{4pt}
\fitwidth{\begin{tabular}{@{}l cc cc cc cc @{}}
\toprule
 & \multicolumn{2}{c}{ResNet, FT} & \multicolumn{2}{c}{ResNet, ER} & \multicolumn{2}{c}{ViT-B/16, FT} & \multicolumn{2}{c}{ViT-B/16, ER} \\
\cmidrule(lr){2-3}\cmidrule(lr){4-5}\cmidrule(lr){6-7}\cmidrule(lr){8-9}
 & $-$no TTA & $-$+TTA & $-$no TTA & $-$+TTA & $-$no TTA & $-$+TTA & $-$no TTA & $-$+TTA \\
\midrule
C100-C & $+$12.2\,{\scriptsize[$+$11.5, $+$12.9]}$^{*}$ & $+$16.5\,{\scriptsize[$+$16.1, $+$16.8]}$^{*}$ & $+$1.1\,{\scriptsize[$+$1.0, $+$1.2]}$^{*}$ & $+$19.5\,{\scriptsize[$+$19.3, $+$19.6]}$^{*}$ & $+$1.4\,{\scriptsize[$+$0.9, $+$1.9]}$^{*}$ & $+$0.7\,{\scriptsize[$+$0.5, $+$0.9]}$^{*}$ & $+$0.5\,{\scriptsize[$+$0.1, $+$0.9]}$^{*}$ & $+$0.5\,{\scriptsize[$+$0.2, $+$0.8]}$^{*}$ \\
C100-TF & $+$16.7\,{\scriptsize[$+$16.4, $+$17.0]}$^{*}$ & $+$17.9\,{\scriptsize[$+$17.3, $+$18.5]}$^{*}$ & $+$5.8\,{\scriptsize[$+$5.3, $+$6.3]}$^{*}$ & $+$17.4\,{\scriptsize[$+$16.6, $+$18.1]}$^{*}$ & $+$1.2\,{\scriptsize[$+$0.6, $+$1.7]}$^{*}$ & $+$0.6\,{\scriptsize[$+$0.3, $+$0.8]}$^{*}$ & $+$1.2\,{\scriptsize[$+$0.8, $+$1.6]}$^{*}$ & $+$0.7\,{\scriptsize[$+$0.5, $+$1.0]}$^{*}$ \\
C20 & $+$0.1\,{\scriptsize[$-$0.3, $+$0.5]} & $-$0.2\,{\scriptsize[$-$0.5, $+$0.1]} & $+$0.4\,{\scriptsize[$\pm$0.0, $+$0.8]}$^{*}$ & $+$0.5\,{\scriptsize[$-$0.1, $+$1.0]}$^{*}$ & $+$1.0\,{\scriptsize[$+$0.4, $+$1.6]}$^{*}$ & $+$1.0\,{\scriptsize[$+$0.5, $+$1.5]}$^{*}$ & $+$0.8\,{\scriptsize[$+$0.4, $+$1.2]}$^{*}$ & $+$0.7\,{\scriptsize[$+$0.1, $+$1.3]}$^{*}$ \\
DomainNet & $+$4.4\,{\scriptsize[$+$3.3, $+$5.4]}$^{*}$ & $+$6.0\,{\scriptsize[$+$5.5, $+$6.6]}$^{*}$ & $\pm$0.0\,{\scriptsize[$-$0.4, $+$0.3]} & $+$11.8\,{\scriptsize[$+$11.3, $+$12.3]}$^{*}$ & $+$0.6\,{\scriptsize[$+$0.4, $+$0.9]}$^{*}$ & $+$0.7\,{\scriptsize[$+$0.6, $+$0.8]}$^{*}$ & $+$0.8\,{\scriptsize[$+$0.7, $+$1.0]}$^{*}$ & $+$0.7\,{\scriptsize[$+$0.6, $+$0.8]}$^{*}$ \\
Inet-TF & $+$34.2\,{\scriptsize[$+$32.9, $+$35.6]}$^{*}$ & $+$35.8\,{\scriptsize[$+$35.4, $+$36.1]}$^{*}$ & $+$9.1\,{\scriptsize[$+$8.2, $+$10.0]}$^{*}$ & $+$37.2\,{\scriptsize[$+$36.5, $+$38.0]}$^{*}$ & $+$0.4\,{\scriptsize[$-$0.1, $+$0.9]}$^{*}$ & $+$0.3\,{\scriptsize[$+$0.1, $+$0.5]}$^{*}$ & $+$0.2\,{\scriptsize[$-$0.1, $+$0.5]} & $+$0.3\,{\scriptsize[$+$0.1, $+$0.4]}$^{*}$ \\
IN-C & $+$5.2\,{\scriptsize[$+$5.0, $+$5.4]}$^{*}$ & $+$6.8\,{\scriptsize[$+$6.5, $+$7.1]}$^{*}$ & $+$2.4\,{\scriptsize[$+$2.2, $+$2.7]}$^{*}$ & $+$18.3\,{\scriptsize[$+$18.1, $+$18.4]}$^{*}$ & $+$2.0\,{\scriptsize[$+$1.8, $+$2.1]}$^{*}$ & $+$0.5\,{\scriptsize[$+$0.4, $+$0.7]}$^{*}$ & $+$0.9\,{\scriptsize[$+$0.7, $+$1.1]}$^{*}$ & $+$0.5\,{\scriptsize[$+$0.2, $+$0.7]}$^{*}$ \\
\bottomrule
\end{tabular}
}
\vspace{-2mm}
\end{table}

\subsection{Every TTA method}
\paragraph{Subsets of TTA methods.} Some implementations miss or replace a core component of the original method, and some read training images again at the start of each \prv{preview}, which a strict single-pass protocol excludes (\cref{sec:s_tta_impl}). \Cref{tab:main} therefore takes the median over the implementations that keep every core component, namely all except PETAL (no SWAG posterior), ViDA (no source-pretrained adapters), DePT (no official code, an L2 pull instead of its regularizer), DPCore (prompt added instead of prepended), DAS (images of the replay buffer instead of a knowledge base of generated images) and POEM (protector on training images). \Cref{tab:s_subsets} gives the same medians over every method and over the strict single-pass methods, on the same runs and seeds. Over every method, \ours is at least as good as the CL method alone in \nAllAlone{} of the \nCells{} settings and above \cltta in \nAllPlain{}. Without the methods that read training or buffer images again (EATA, POEM, DPCore, DAS and ReservoirTTA), it is at least as good in \nStrictAlone{} and above in \nStrictPlain{}. Every mean stays within $0.5$ points of \cref{tab:main}.

\begin{table}[H]
\centering
\caption{\textbf{\Cref{tab:main} over other sets of TTA methods} (score, \%, medians over the methods of each set, same runs and seeds). Top: every method (\nAllMin{} to \nAllMax{} per setting). Bottom: strict single pass, without EATA, POEM, DPCore, DAS and ReservoirTTA, which read training or buffer images again during the \prv{preview} (\nStrictMin{} to \nStrictMax{} per setting). \Cref{tab:main} itself uses the implementations that keep every core component (\nCoreMin{} to \nCoreMax{} per setting). Bold: best method.}
\label{tab:s_subsets}
\vspace{-2mm}
\small
\setlength{\tabcolsep}{4pt}
\fitwidth{\begin{tabular}{@{}l ccc ccc ccc ccc@{}}
\toprule
 & \multicolumn{3}{c}{ResNet, FT} & \multicolumn{3}{c}{ResNet, ER} & \multicolumn{3}{c}{ViT-B/16, FT} & \multicolumn{3}{c}{ViT-B/16, ER} \\
\cmidrule(lr){2-4}\cmidrule(lr){5-7}\cmidrule(lr){8-10}\cmidrule(lr){11-13}
 & no TTA & +TTA & \textbf{Ours} & no TTA & +TTA & \textbf{Ours} & no TTA & +TTA & \textbf{Ours} & no TTA & +TTA & \textbf{Ours} \\
\midrule
\multicolumn{13}{@{}l}{\emph{Every TTA method}} \\
C100-C & 48.6 & 44.2 & \textbf{60.7} & 64.7 & 45.2 & \textbf{65.8} & 65.6 & 66.0 & \textbf{66.5} & 69.0 & 68.8 & \textbf{69.3} \\
C100-TF & 49.7 & 48.5 & \textbf{66.4} & 63.1 & 51.6 & \textbf{68.9} & 73.4 & 73.8 & \textbf{74.3} & 75.2 & 75.6 & \textbf{76.3} \\
C20 & 63.7 & \textbf{64.1} & 63.9 & 68.1 & 68.1 & \textbf{68.6} & 76.9 & 77.0 & \textbf{77.8} & 79.1 & 79.2 & \textbf{79.9} \\
DomainNet & 39.1 & 37.2 & \textbf{43.5} & \textbf{51.1} & 39.3 & \textbf{51.1} & 62.8 & 62.7 & \textbf{63.3} & 63.9 & 64.0 & \textbf{64.6} \\
Inet-TF & 42.1 & 40.7 & \textbf{76.7} & 70.8 & 42.7 & \textbf{80.0} & 95.7 & 95.8 & \textbf{96.1} & 96.0 & 95.9 & \textbf{96.2} \\
IN-C & 17.1 & 15.3 & \textbf{22.4} & 31.8 & 15.7 & \textbf{34.3} & 50.7 & \textbf{51.4} & 51.3 & 56.4 & 56.4 & \textbf{57.1} \\
Mean & 43.4 & 41.7 & \textbf{55.6} & 58.3 & 43.8 & \textbf{61.4} & 70.8 & 71.1 & \textbf{71.6} & 73.3 & 73.3 & \textbf{73.9} \\
\midrule
\multicolumn{13}{@{}l}{\emph{Strict single pass: no method reads training or buffer images again}} \\
C100-C & 48.6 & 43.7 & \textbf{60.7} & 64.7 & 45.2 & \textbf{65.8} & 65.6 & 66.0 & \textbf{66.3} & 69.0 & 68.9 & \textbf{69.2} \\
C100-TF & 49.7 & 48.0 & \textbf{66.2} & 63.1 & 51.4 & \textbf{68.9} & 73.4 & 73.7 & \textbf{74.1} & 75.2 & 75.5 & \textbf{76.2} \\
C20 & 63.7 & \textbf{64.1} & 63.8 & 68.1 & 68.0 & \textbf{68.5} & 76.9 & 76.8 & \textbf{77.8} & 79.1 & 79.2 & \textbf{79.9} \\
DomainNet & 39.1 & 37.3 & \textbf{43.8} & \textbf{51.1} & 39.5 & \textbf{51.1} & 62.8 & 62.7 & \textbf{63.4} & 63.9 & 63.9 & \textbf{64.6} \\
Inet-TF & 42.1 & 40.8 & \textbf{76.8} & 70.8 & 42.7 & \textbf{79.9} & 95.7 & 95.7 & \textbf{96.0} & 96.0 & 95.9 & \textbf{96.1} \\
IN-C & 17.1 & 15.3 & \textbf{22.3} & 31.8 & 15.7 & \textbf{34.3} & 50.7 & \textbf{52.0} & 51.3 & 56.4 & 56.5 & \textbf{57.1} \\
Mean & 43.4 & 41.5 & \textbf{55.6} & 58.3 & 43.7 & \textbf{61.4} & 70.8 & 71.2 & \textbf{71.5} & 73.3 & 73.3 & \textbf{73.9} \\
\bottomrule
\end{tabular}
}
\vspace{-2mm}
\end{table}

\paragraph{Every method.} \Cref{tab:methods_ours} gives the gain of \ours over \cltta for every TTA method in our protocol, and \cref{tab:main} reports its medians over the methods that keep every core component. Tent and ROID, which the analyses average, are typical of the other methods. Over the settings of \cref{tab:main}, the gain of \ours over \cltta is $0.1$ points above the median over every method with Tent and $0.2$ points below it with ROID. \Cref{tab:methods,tab:methods_ft,tab:methods_corr} compare \ours with \cltta in the test-batch protocol, with ER, with FT and on the corruption benchmarks. %
Of the 20 TTA methods, DAS, DePT, DPCore and ViDA only run on ViTs, while Norm, NOTE, PALM, RoTTA and Surgical only run on ResNets, so that 16 methods run on ResNets and 15 on ViTs. Every method was also run in the test-batch protocol, on every benchmark where it runs in ours (\cref{tab:methods_ours}, ``--'': not run). Notably, a ViT has no BatchNorm statistics, so that its two protocols coincide, and its test-batch columns repeat those of \cref{tab:methods_ours}. The comparison is split over four tables. \Cref{tab:methods} gives ER on C100-TF, Inet-TF and DomainNet for ResNets, and on C100-TF, C20, Inet-TF and DomainNet for ViTs. \Cref{tab:methods_ft} gives FT, as well as ER on ResNet C20. \Cref{tab:methods_corr} gives C100-C and IN-C, where nine methods run, except Tent and ROID. Tent and ROID are given on every benchmark in \cref{tab:protocols,tab:oldnew}.
Over these tables, with ER, \ours is at least as good as \cltta in the test-batch protocol in \nTbErWins{} of \nTbErCells{} pairs of method and benchmark, including ReservoirTTA~\cite{vray2025reservoirtta} around Tent, EATA, SAR and ROID, with a median gain of \medResNet{} points on ResNets and \medViT{} on ViTs in \cref{tab:methods}. Its three losses are PALM~\cite{maharana2025palm} on ResNet DomainNet and C20 ($-0.3$ and $-0.8$) and EATA on ViT IN-C ($-0.1$). Without a buffer, \ours is at least as good in \nTbFtWins{} of \nTbFtCells{} pairs. On ResNets, it is at least as good on C100-TF and Inet-TF with every method, but it falls below on DomainNet with every method, on C20 with 16 of 18 methods, and on C100-C and IN-C with about half of them, where test-batch statistics recover what memory-free CL has lost.
\Cref{tab:vitgain} gives what each method gains on the preview of a ViT, and \cref{tab:x10} the scores at ten times the official TTA learning rate. %
With ER, \ours is above the CL method alone in 103 of the 122 pairs of \cref{tab:methods}. In 12 others, the TTA method itself damages the network (\cltta loses up to $12$ points), and \ours recovers part of the loss.

\begin{table}[H]
\centering
\caption{\textbf{\ours $-$ \cltta for every TTA method, in our batch-independent protocol} (score, points; mixed batches; top: ResNets, bottom: ViT-B/16; ``--'': not run on this benchmark). \Cref{tab:main} gives the medians of these columns.}
\label{tab:methods_ours}
\small
\setlength{\tabcolsep}{3pt}
\fitwidth{\begin{tabular}{@{}c@{}}\begin{tabular}{@{}l cccccc cccccc@{}}
\toprule
 & \multicolumn{6}{c}{FT} & \multicolumn{6}{c}{ER} \\
\cmidrule(lr){2-7}\cmidrule(lr){8-13}
 & C100-C & C100-TF & C20 & DN & Inet-TF & IN-C & C100-C & C100-TF & C20 & DN & Inet-TF & IN-C \\
\midrule
CMF & $+$16.0 & $+$17.6 & $-$0.4 & $+$4.8 & $+$35.5 & $+$9.7 & $+$16.2 & $+$16.7 & $+$0.4 & $+$10.5 & $+$37.8 & $+$21.2 \\
CoTTA & $+$16.7 & $+$19.0 & $-$0.2 & $+$7.0 & $+$36.2 & $+$6.4 & $+$21.0 & $+$18.4 & $+$0.5 & $+$12.1 & $+$37.2 & $+$16.9 \\
EATA & $+$16.4 & $+$18.6 & $-$0.2 & $+$6.3 & $+$36.1 & $+$7.4 & $+$17.3 & $+$17.7 & $+$0.4 & $+$11.6 & $+$37.6 & $+$18.5 \\
LAW & -- & $+$19.1 & $-$1.1 & $+$5.7 & $+$36.3 & -- & -- & $+$18.1 & $+$0.5 & $+$11.1 & $+$37.6 & -- \\
LCoTTA & -- & $+$19.0 & $-$0.1 & $+$6.9 & $+$36.3 & -- & -- & $+$18.3 & $+$0.6 & $+$11.9 & $+$37.2 & -- \\
Norm & $+$17.5 & $+$19.2 & $-$0.3 & $+$7.0 & $+$36.1 & $+$6.6 & $+$23.2 & $+$18.4 & $+$0.6 & $+$12.2 & $+$37.3 & $+$18.0 \\
NOTE & -- & $+$0.3 & $+$0.2 & $-$0.1 & $+$0.8 & -- & -- & $+$0.2 & $\pm$0.0 & $+$0.1 & $+$0.6 & -- \\
PALM & -- & $+$17.7 & $-$0.9 & $+$1.7 & $+$35.6 & -- & -- & $+$16.9 & $-$0.6 & $+$9.0 & $+$36.2 & -- \\
PETAL & -- & $+$3.3 & $-$0.2 & $-$1.3 & $+$6.5 & -- & -- & $+$3.3 & $+$0.8 & $+$1.8 & $+$4.9 & -- \\
POEM & $+$15.2 & $+$19.0 & $-$0.2 & $+$6.3 & $+$35.9 & $+$8.9 & $+$20.7 & $+$18.5 & $+$0.6 & $+$12.0 & $+$37.5 & $+$20.6 \\
RDumb & -- & $+$18.9 & $-$0.2 & $+$6.6 & $+$36.0 & -- & -- & $+$18.1 & $+$0.4 & $+$11.5 & $+$37.6 & -- \\
ROID & $+$15.6 & $+$17.4 & $-$0.1 & $+$4.7 & $+$35.9 & $+$8.7 & $+$15.8 & $+$16.9 & $+$0.4 & $+$10.4 & $+$37.5 & $+$19.8 \\
RoTTA & $-$0.7 & $-$0.1 & $+$0.5 & $+$0.3 & $-$0.3 & $+$0.4 & $+$23.0 & $+$21.2 & $\pm$0.0 & $+$10.5 & $+$27.0 & $+$18.7 \\
SAR & $+$16.9 & $+$17.8 & $-$0.2 & $+$6.8 & $+$36.2 & $+$6.6 & $+$18.5 & $+$16.8 & $+$0.5 & $+$12.1 & $+$37.2 & $+$17.6 \\
Surgical & -- & $+$17.4 & $-$3.8 & $-$6.4 & $+$37.6 & -- & -- & $+$18.9 & $+$2.6 & $+$14.0 & $+$39.7 & -- \\
Tent & $+$18.7 & $+$17.2 & $-$0.2 & $+$6.9 & $+$35.4 & $+$7.0 & $+$20.6 & $+$17.0 & $+$0.6 & $+$11.9 & $+$37.6 & $+$18.0 \\
\midrule
Reservoir+EATA & -- & $+$17.9 & $-$0.8 & $+$5.9 & $+$35.4 & -- & -- & $+$17.4 & $+$0.7 & $+$12.0 & $+$36.8 & -- \\
Reservoir+ROID & -- & $+$17.7 & $-$1.2 & $+$5.8 & $+$35.5 & -- & -- & $+$16.6 & $+$0.2 & $+$11.8 & $+$37.2 & -- \\
Reservoir+SAR & -- & $+$17.8 & $-$0.1 & $+$6.2 & $+$35.9 & -- & -- & $+$16.7 & $+$0.5 & $+$12.1 & $+$37.2 & -- \\
Reservoir+Tent & -- & $+$17.5 & $\pm$0.0 & $+$6.4 & $+$35.8 & -- & -- & $+$17.1 & $+$0.5 & $+$12.2 & $+$37.0 & -- \\
\bottomrule
\end{tabular}
\\[2mm]
\begin{tabular}{@{}l cccccc cccccc@{}}
\toprule
 & \multicolumn{6}{c}{FT} & \multicolumn{6}{c}{ER} \\
\cmidrule(lr){2-7}\cmidrule(lr){8-13}
 & C100-C & C100-TF & C20 & DN & Inet-TF & IN-C & C100-C & C100-TF & C20 & DN & Inet-TF & IN-C \\
\midrule
CMF & $+$1.5 & $+$0.6 & $+$0.8 & $+$0.5 & $+$0.1 & $+$1.1 & $+$0.4 & $+$0.8 & $+$0.7 & $+$0.9 & $+$0.6 & $+$0.8 \\
CoTTA & $-$0.4 & $\pm$0.0 & $+$1.0 & $+$0.7 & $+$0.3 & $-$9.0 & $+$56.5 & $+$0.6 & $+$0.9 & $+$0.9 & $+$0.2 & $-$1.6 \\
DAS & -- & -- & -- & -- & -- & -- & -- & $+$0.3 & $\pm$0.0 & $+$0.1 & $+$0.2 & -- \\
DePT & -- & $+$0.7 & $+$1.0 & $+$0.6 & $+$0.4 & -- & -- & $+$0.6 & $+$0.6 & $+$0.6 & $\pm$0.0 & -- \\
DPCore & $+$0.3 & $+$0.7 & $+$0.9 & $+$0.6 & $+$0.3 & $\pm$0.0 & $+$0.5 & $+$0.6 & $+$0.7 & $+$0.7 & $+$0.2 & $+$0.1 \\
EATA & $+$0.6 & $+$0.5 & $+$1.2 & $+$0.7 & $+$0.3 & $+$0.9 & $+$0.3 & $+$0.8 & $+$0.7 & $+$0.7 & $+$0.2 & $+$0.9 \\
LAW & -- & $+$0.9 & $+$0.5 & $+$1.2 & $+$0.7 & -- & -- & $+$0.7 & $+$1.0 & $+$1.8 & $+$0.5 & -- \\
LCoTTA & -- & $+$0.4 & $+$1.0 & $+$0.6 & $+$0.2 & -- & -- & $+$0.4 & $+$0.6 & $+$0.9 & $+$0.1 & -- \\
PETAL & -- & $+$0.8 & $+$1.5 & $-$0.3 & $+$0.2 & -- & -- & $+$2.1 & $+$2.0 & $+$5.2 & $\pm$0.0 & -- \\
POEM & $+$0.5 & $+$0.5 & $+$1.3 & $+$2.2 & $+$0.4 & $+$1.1 & $+$0.3 & $+$0.8 & $+$0.7 & $+$1.0 & $+$0.4 & $+$8.7 \\
RDumb & -- & $+$0.4 & $+$1.0 & $+$0.7 & $+$0.3 & -- & -- & $+$0.8 & $+$0.7 & $+$0.7 & $+$0.1 & -- \\
ROID & $+$1.8 & $+$0.8 & $+$1.1 & $+$0.6 & $+$0.1 & $+$0.6 & $+$0.5 & $+$0.6 & $+$0.7 & $+$0.6 & $+$0.6 & $+$0.4 \\
SAR & $+$0.3 & $+$0.3 & $+$0.7 & $+$0.7 & $+$0.4 & $+$0.5 & $+$0.5 & $+$0.8 & $+$0.9 & $+$0.8 & $+$0.2 & $+$0.2 \\
Tent & $+$0.2 & $+$0.5 & $+$0.9 & $+$0.6 & $+$0.3 & $\pm$0.0 & $+$0.2 & $+$0.6 & $+$0.9 & $+$0.7 & $+$0.2 & $+$1.1 \\
ViDA & $-$2.4 & $+$1.2 & $+$1.0 & $+$0.5 & $+$0.3 & $+$0.2 & $+$6.2 & $+$0.8 & $+$0.8 & $+$0.6 & $+$0.2 & $+$0.1 \\
\midrule
Reservoir+EATA & -- & $+$0.8 & $+$0.8 & $+$0.9 & $+$0.1 & -- & -- & $+$0.7 & $+$0.4 & $+$0.7 & $+$0.5 & -- \\
Reservoir+ROID & -- & $+$0.6 & $+$1.0 & $+$0.6 & $+$0.7 & -- & -- & $+$0.7 & $+$0.6 & $+$0.6 & $+$0.1 & -- \\
Reservoir+SAR & -- & $+$1.0 & $+$1.0 & $+$0.6 & $+$0.3 & -- & -- & $+$0.9 & $+$0.7 & $+$0.6 & $+$0.4 & -- \\
Reservoir+Tent & -- & $+$0.9 & $+$0.8 & $+$0.6 & $+$0.5 & -- & -- & $+$0.6 & $+$0.6 & $+$0.7 & $+$0.2 & -- \\
\bottomrule
\end{tabular}
\end{tabular}}
\end{table}

\begin{table}[H]
\centering
\caption{\textbf{\ours $-$ \cltta for every TTA method} (score, points; ER, mixed batches; \ours in the batch-independent protocol, \cltta in the test-batch one; ``--'': not applicable). For CoTTA, PETAL, RoTTA and ViDA, both systems start each task from the deployed teacher. \ours is at least as good in \nMethodWins{} of \nMethodCells{} entries.}
\label{tab:methods}
\small
\setlength{\tabcolsep}{4pt}
\fitwidth{\begin{tabular}{@{}l ccc cccc@{}}
\toprule
 & \multicolumn{3}{c}{ResNet (ER)} & \multicolumn{4}{c}{ViT-B/16 (ER)} \\
\cmidrule(lr){2-4}\cmidrule(lr){5-8}
 & C100-TF & Inet-TF & DomainNet & C100-TF & C20 & Inet-TF & DomainNet \\
\midrule
CMF & $+$2.7 & $+$4.9 & $+$1.7 & $+$0.9 & $+$0.8 & $+$0.6 & $+$0.7 \\
CoTTA & $+$3.2 & $+$5.0 & $+$1.9 & $+$0.6 & $+$0.9 & $+$0.2 & $+$0.9 \\
DAS & -- & -- & -- & $+$0.3 & $\pm$0.0 & $+$0.2 & $+$0.1 \\
DePT & -- & -- & -- & $+$0.6 & $+$0.6 & $\pm$0.0 & $+$0.6 \\
DPCore & -- & -- & -- & $+$0.6 & $+$0.7 & $+$0.2 & $+$0.7 \\
EATA & $+$3.1 & $+$4.8 & $+$1.9 & $+$0.8 & $+$0.7 & $+$0.2 & $+$0.7 \\
LAW & $+$3.3 & $+$5.2 & $+$1.6 & $+$0.7 & $+$1.0 & $+$0.5 & $+$1.8 \\
LCoTTA & $+$3.3 & $+$4.9 & $+$1.9 & $+$0.4 & $+$0.6 & $+$0.1 & $+$0.9 \\
Norm & $+$3.4 & $+$4.8 & $+$1.9 & -- & -- & -- & -- \\
NOTE & $+$0.1 & $+$0.7 & $+$0.1 & -- & -- & -- & -- \\
PALM & $+$2.9 & $+$3.4 & $-$0.4 & -- & -- & -- & -- \\
PETAL & $+$3.2 & $+$4.9 & $+$1.8 & $+$2.1 & $+$2.0 & $\pm$0.0 & $+$5.2 \\
POEM & $+$3.3 & $+$5.1 & $+$2.0 & $+$0.6 & $+$1.0 & $+$0.4 & $+$0.9 \\
RDumb & $+$3.3 & $+$4.8 & $+$1.8 & $+$0.8 & $+$0.7 & $+$0.1 & $+$0.7 \\
RoTTA & $+$20.9 & $+$27.5 & $+$10.5 & -- & -- & -- & -- \\
SAR & $+$2.8 & $+$4.8 & $+$1.9 & $+$0.8 & $+$0.9 & $+$0.2 & $+$0.8 \\
Surgical & $+$3.9 & $+$6.7 & $+$8.3 & -- & -- & -- & -- \\
ViDA & -- & -- & -- & $+$0.8 & $+$0.8 & $+$0.2 & $+$0.6 \\
\midrule
Reservoir+EATA & $+$2.7 & $+$5.0 & $+$1.9 & $+$0.7 & $+$0.4 & $+$0.5 & $+$0.7 \\
Reservoir+ROID & $+$3.0 & $+$4.3 & $+$1.3 & $+$0.7 & $+$0.6 & $+$0.1 & $+$0.6 \\
Reservoir+SAR & $+$2.9 & $+$5.0 & $+$1.9 & $+$0.9 & $+$0.7 & $+$0.4 & $+$0.6 \\
Reservoir+Tent & $+$3.0 & $+$5.0 & $+$1.9 & $+$0.6 & $+$0.6 & $+$0.2 & $+$0.7 \\
\bottomrule
\end{tabular}
}
\end{table}

\begin{table}[H]
\centering
\caption{\textbf{\ours $-$ \cltta for every TTA method, with FT} (and ER on C20; score, points; mixed batches; \ours in the batch-independent protocol, \cltta in the test-batch one; ``--'': not applicable). Gray: \ours below the CL method alone. \ours is at least as good as \cltta in \nFtWins{} of \nFtCells{} entries and as the CL method alone in \nFtAlone{}. Without a buffer on DomainNet, the test-batch protocol gives \cltta about $2$ points.}
\label{tab:methods_ft}
\small
\setlength{\tabcolsep}{4pt}
\fitwidth{\begin{tabular}{@{}l cccc c ccc@{}}
\toprule
 & \multicolumn{4}{c}{ResNet (FT)} & ResNet (ER) & \multicolumn{3}{c}{ViT-B/16 (FT)} \\
\cmidrule(lr){2-5}\cmidrule(lr){6-6}\cmidrule(lr){7-9}
 & C100-TF & Inet-TF & DomainNet & C20 & C20 & C100-TF & C20 & Inet-TF \\
\midrule
CMF & $+$3.1 & $+$6.2 & $-$3.1 & $-$0.2 & $+$0.6 & $+$0.6 & $+$0.8 & $+$0.1 \\
CoTTA & $+$3.3 & $+$6.4 & $-$1.6 & \textcolor{black!50}{$\pm$0.0} & $+$0.8 & \textcolor{black!50}{$\pm$0.0} & \textcolor{black!50}{$+$1.0} & $+$0.3 \\
DePT & -- & -- & -- & -- & -- & $+$0.7 & $+$1.0 & $+$0.4 \\
DPCore & -- & -- & -- & -- & -- & \textcolor{black!50}{$+$0.7} & $+$0.9 & \textcolor{black!50}{$+$0.3} \\
EATA & $+$3.4 & $+$6.7 & $-$2.0 & \textcolor{black!50}{$-$0.2} & $+$0.7 & $+$0.5 & $+$1.2 & $+$0.3 \\
LAW & $+$3.4 & $+$6.2 & $-$2.6 & \textcolor{black!50}{$-$1.1} & $+$0.6 & $+$0.9 & $+$0.5 & \textcolor{black!50}{$+$0.7} \\
LCoTTA & $+$3.6 & $+$6.5 & $-$1.6 & \textcolor{black!50}{$-$0.3} & $+$0.7 & $+$0.4 & $+$1.0 & $+$0.2 \\
Norm & $+$3.7 & $+$6.5 & $-$1.6 & $-$0.4 & $+$0.8 & -- & -- & -- \\
NOTE & $+$0.3 & $+$0.8 & \textcolor{black!50}{$-$0.1} & \textcolor{black!50}{$+$0.2} & \textcolor{black!50}{$\pm$0.0} & -- & -- & -- \\
PALM & $+$3.1 & $+$4.0 & \textcolor{black!50}{$-$6.0} & \textcolor{black!50}{$-$2.3} & \textcolor{black!50}{$-$0.7} & -- & -- & -- \\
PETAL & $+$3.2 & $+$6.5 & $-$1.4 & \textcolor{black!50}{$-$0.1} & $+$0.8 & \textcolor{black!50}{$+$0.8} & \textcolor{black!50}{$+$1.5} & $+$0.2 \\
POEM & $+$3.5 & $+$6.4 & $-$2.2 & \textcolor{black!50}{$-$1.6} & $+$0.7 & $+$0.4 & $+$0.7 & $+$0.3 \\
RDumb & $+$3.6 & $+$6.4 & $-$1.8 & \textcolor{black!50}{$-$0.3} & $+$0.6 & $+$0.4 & $+$1.0 & $+$0.3 \\
RoTTA & \textcolor{black!50}{$-$0.1} & $-$0.3 & \textcolor{black!50}{$+$0.3} & $+$0.4 & $+$0.3 & -- & -- & -- \\
SAR & $+$3.1 & $+$6.4 & $-$1.7 & \textcolor{black!50}{$-$0.3} & $+$0.8 & $+$0.3 & $+$0.7 & $+$0.4 \\
Surgical & $+$6.1 & $+$10.4 & \textcolor{black!50}{$-$11.3} & \textcolor{black!50}{$-$5.6} & \textcolor{black!50}{$+$0.9} & -- & -- & -- \\
ViDA & -- & -- & -- & -- & -- & $+$1.2 & $+$1.0 & $+$0.3 \\
\midrule
Reservoir+EATA & $+$3.2 & $+$6.3 & $-$2.4 & $-$0.8 & $+$0.9 & $+$0.8 & $+$0.8 & $+$0.1 \\
Reservoir+ROID & $+$3.8 & $+$5.7 & $-$2.8 & $-$1.2 & $+$0.7 & $+$0.6 & $+$1.0 & $+$0.7 \\
Reservoir+SAR & $+$3.3 & $+$6.2 & $-$2.4 & \textcolor{black!50}{$-$0.3} & $+$0.7 & $+$1.0 & $+$1.0 & $+$0.3 \\
Reservoir+Tent & $+$3.2 & $+$6.2 & $-$2.3 & $-$0.2 & $+$0.7 & $+$0.9 & $+$0.8 & $+$0.5 \\
\bottomrule
\end{tabular}
}
\end{table}

\begin{table}[H]
\centering
\caption{\textbf{Seven TTA methods per backbone on the corruption benchmarks} (score, points; mixed batches; \ours in the batch-independent protocol, \cltta in the test-batch one; Tent and ROID in \cref{tab:protocols,tab:oldnew}). Without a buffer on C100-C, \cltta with test-batch statistics is $9$ to $11$ points above \ours with EATA, SAR and CMF, as for Tent and ROID in \cref{tab:main}.}
\label{tab:methods_corr}
\small
\setlength{\tabcolsep}{3pt}
\fitwidth{\begin{tabular}{@{}l cc cc cc cc@{}}
\toprule
 & \multicolumn{4}{c}{ResNet} & \multicolumn{4}{c}{ViT-B/16} \\
\cmidrule(lr){2-5}\cmidrule(lr){6-9}
 & \multicolumn{2}{c}{C100-C} & \multicolumn{2}{c}{IN-C} & \multicolumn{2}{c}{C100-C} & \multicolumn{2}{c}{IN-C} \\
 & FT & ER & FT & ER & FT & ER & FT & ER \\
\midrule
\multicolumn{9}{@{}l}{\emph{\ours $-$ \cltta}} \\
CMF & $+$2.3 & $+$3.3 & $+$5.0 & $+$14.0 & $+$1.5 & $+$0.4 & $+$1.1 & $+$0.8 \\
CoTTA & $+$0.2 & $+$1.2 & $-$1.7 & $+$2.8 & $-$0.4 & $+$56.5 & $-$9.0 & $-$1.6 \\
DPCore & -- & -- & -- & -- & $+$0.3 & $+$0.5 & $\pm$0.0 & $+$0.1 \\
EATA & $+$2.4 & $+$3.6 & $+$0.7 & $+$6.9 & $+$0.6 & $+$0.3 & $+$0.9 & $+$0.9 \\
Norm & $+$0.7 & $+$1.9 & $-$1.1 & $+$4.1 & -- & -- & -- & -- \\
POEM & $-$1.2 & $+$1.5 & $+$1.1 & $+$8.9 & $+$0.2 & $+$0.4 & $\pm$0.0 & $+$4.9 \\
RoTTA & $-$1.0 & $+$23.6 & $+$0.4 & $+$18.7 & -- & -- & -- & -- \\
SAR & $+$3.9 & $+$5.7 & $-$0.9 & $+$5.2 & $+$0.3 & $+$0.5 & $+$0.5 & $+$0.2 \\
ViDA & -- & -- & -- & -- & $-$2.4 & $+$6.2 & $+$0.2 & $+$0.1 \\
\midrule
\multicolumn{9}{@{}l}{\emph{\ours $-$ no TTA}} \\
CMF & $+$12.6 & $+$1.3 & $+$4.9 & $+$2.3 & $+$3.0 & $+$1.4 & $+$3.2 & $+$1.1 \\
CoTTA & $+$12.4 & $+$1.0 & $+$5.8 & $+$2.5 & $-$62.6 & $-$9.7 & $-$7.2 & $-$1.5 \\
DPCore & -- & -- & -- & -- & $-$0.1 & $+$0.3 & $-$0.6 & $-$0.6 \\
EATA & $+$12.2 & $+$1.2 & $+$5.7 & $+$2.9 & $+$1.8 & $+$0.6 & $+$3.2 & $+$1.2 \\
Norm & $+$12.0 & $+$0.6 & $+$4.8 & $+$1.9 & -- & -- & -- & -- \\
POEM & $+$11.2 & $+$1.0 & $+$1.9 & $-$2.2 & $+$0.7 & $+$0.2 & $-$0.5 & $-$3.7 \\
RoTTA & $-$7.6 & $\pm$0.0 & $-$1.7 & $+$2.0 & -- & -- & -- & -- \\
SAR & $+$12.1 & $+$1.2 & $+$5.3 & $+$2.6 & $+$1.0 & $+$0.3 & $+$1.2 & $+$0.7 \\
ViDA & -- & -- & -- & -- & $-$9.6 & $-$3.1 & $\pm$0.0 & $+$0.1 \\
\bottomrule
\end{tabular}
}
\end{table}

\begin{table}[H]
\centering
\caption{\textbf{Gain of each TTA method on the \prv{preview} of a ViT} (ER, \cltta, mixed batches, \nSeeds{} seeds): accuracy of the adapted network on the \prv{preview} stream after the pass minus the accuracy before adaptation, in points; ``--'': not run. Bottom: the methods built for ViTs that learn new parameters (prompts, adapters).}
\label{tab:vitgain}
\vspace{-2mm}
\small
\fitwidth{\begin{tabular}{@{}l cccccc@{}}
\toprule
 & C100-C & C100-TF & IN-C & C20 & DomainNet & Inet-TF \\
\midrule
Tent & $+$1.6 & $+$0.8 & $+$5.2 & $\pm$0.0 & $-$0.2 & $+$0.1 \\
EATA & -- & $+$2.1 & $+$13.9 & $+$0.1 & $+$0.8 & $+$0.1 \\
SAR & -- & $+$0.9 & $+$7.2 & $\pm$0.0 & $-$0.2 & $\pm$0.0 \\
ROID & $+$6.2 & $+$5.9 & $+$9.8 & $+$1.1 & $+$1.0 & $-$0.3 \\
CMF & -- & $+$6.0 & $+$12.8 & $+$1.1 & $+$1.1 & $-$0.7 \\
DAS & -- & $+$6.2 & -- & $+$2.5 & $+$1.4 & $+$2.2 \\
\midrule
DPCore & -- & $+$0.6 & -- & $-$0.5 & $-$0.2 & $-$0.1 \\
DePT & -- & $\pm$0.0 & -- & $\pm$0.0 & $\pm$0.0 & $\pm$0.0 \\
ViDA & -- & $+$0.2 & -- & $-$0.6 & $\pm$0.0 & $\pm$0.0 \\
\bottomrule
\end{tabular}
}
\vspace{-2mm}
\end{table}

\begin{table}[H]
\centering
\caption{\textbf{TTA learning rate $\times10$}: score of \cltta / \ours (ER, mixed batches). \cltta: test-batch protocol, next task from the network it deploys (its teacher for CoTTA), \nSeeds{} seeds; \ours: batch-independent protocol, next task from $\theta_\mathrm{tr}$, \nSeeds{} seeds. \ours is above \cltta in \nXTenWin{} of \nXTen{} pairs and at least as good as the CL method alone in \nXTenAlone{}.}
\label{tab:x10}
\vspace{-2mm}
\small
\setlength{\tabcolsep}{4pt}
\fitwidth{\begin{tabular}{@{}l cc ccc@{}}
\toprule
 & \multicolumn{2}{c}{ResNet} & \multicolumn{3}{c}{ViT-B/16} \\
\cmidrule(lr){2-3}\cmidrule(lr){4-6}
 & C100-TF & Inet-TF & C100-TF & C20 & Inet-TF \\
\midrule
no TTA & 63.1 & 70.8 & 75.2 & 79.1 & 96.0 \\
\midrule
CoTTA & 65.4 / \textbf{68.0} & 74.9 / \textbf{80.0} & 1.6 / \textbf{56.1} & 5.9 / \textbf{59.2} & 91.4 / \textbf{93.4} \\
EATA & 44.0 / \textbf{59.1} & 75.5 / \textbf{80.2} & 76.5 / \textbf{77.0} & 79.4 / 79.4 & 96.3 / \textbf{96.4} \\
ROID & 42.4 / \textbf{59.8} & 73.1 / \textbf{78.4} & 73.1 / \textbf{75.6} & 74.5 / \textbf{76.5} & 93.3 / 92.8 \\
SAR & 30.7 / \textbf{57.6} & 75.3 / \textbf{80.0} & 75.8 / \textbf{76.6} & 78.8 / \textbf{79.9} & 95.7 / \textbf{96.2} \\
Tent & 28.3 / \textbf{57.4} & 75.2 / \textbf{79.9} & 75.9 / \textbf{76.4} & 78.6 / \textbf{79.3} & 96.2 / \textbf{96.4} \\
\bottomrule
\end{tabular}
}
\vspace{-3mm}
\end{table}

\providecommand{\nXOne}{25}\providecommand{\nXOneWin}{25}\providecommand{\nXOneAlone}{25}

\begin{table}[H]
\centering
\caption{\textbf{TTA learning rate $\div10$}: same setting as \cref{tab:x10} with the official rate divided by ten, score of \cltta / \ours (ER, mixed batches, \nSeeds{} seeds). \ours is above \cltta in \nXOneWin{} of \nXOne{} pairs and at least as good as the CL method alone in \nXOneAlone{}.}
\label{tab:x01}
\vspace{-2mm}
\small
\setlength{\tabcolsep}{4pt}
\fitwidth{\begin{tabular}{@{}l cc ccc@{}}
\toprule
 & \multicolumn{2}{c}{ResNet} & \multicolumn{3}{c}{ViT-B/16} \\
\cmidrule(lr){2-3}\cmidrule(lr){4-6}
 & C100-TF & Inet-TF & C100-TF & C20 & Inet-TF \\
\midrule
no TTA & 63.1 & 70.8 & 75.2 & 79.1 & 96.0 \\
\midrule
CoTTA & 65.4 / \textbf{68.8} & 75.0 / \textbf{79.9} & 75.1 / \textbf{75.8} & 79.4 / \textbf{80.1} & 95.9 / \textbf{96.2} \\
EATA & 65.6 / \textbf{68.9} & 75.0 / \textbf{80.1} & 75.4 / \textbf{76.1} & 79.4 / \textbf{80.0} & 95.9 / \textbf{96.5} \\
ROID & 65.7 / \textbf{68.8} & 75.0 / \textbf{79.9} & 75.7 / \textbf{76.7} & 79.3 / \textbf{80.7} & 95.8 / \textbf{96.3} \\
SAR & 65.5 / \textbf{68.9} & 74.9 / \textbf{79.9} & 75.3 / \textbf{76.2} & 79.1 / \textbf{80.0} & 96.0 / \textbf{96.1} \\
Tent & 65.5 / \textbf{68.9} & 75.1 / \textbf{79.9} & 75.2 / \textbf{76.0} & 79.1 / \textbf{80.1} & 95.9 / \textbf{96.2} \\
\bottomrule
\end{tabular}
}
\vspace{-3mm}
\end{table}

\subsection{Ablations and stored statistics}
\Cref{tab:s_ablation} gives every variant of \ours on every benchmark, of which \cref{tab:ablation} keeps the most informative ones. \Cref{tab:s_bn} shows why \cltta cannot simply keep the statistics it estimated during TTA: applying the statistics of the preview to every image collapses the old domains, whichever way they are recorded.

Notably, a hard dose, namely per-domain BatchNorm with the inferred domain, scores the same as the soft dose within $0.1$ points. The posterior finds the domain of $97\%$ of the Inet-TF images, but only $27\%$ for the sub-class domains of C20, where chance is $25\%$ (\cref{fig:tsne}). At the last task, the bank check keeps the bank on Inet-TF and drops it on C20. The shared-parameter check doses the changes of RoTTA~\cite{yuan2023rotta}, Surgical~\cite{lee2023surgical} and PETAL~\cite{brahma2023petal} in $86$, $62$ and $10\%$ of the runs, while it shares those of every other method.

\Cref{tab:s_wotta} gives \ours without a TTA method on every benchmark of \cref{tab:main}. The preview is then only used for the recorded statistics of the new domain and for the prototypes, which the unadapted network builds from its own predictions, while the images of the new domain keep the BatchNorm statistics of $\theta_\mathrm{tr}$.

\begin{table}[H]
\centering
\caption{\textbf{\ours without a TTA method} (score $A$, \%, \nSeeds{} seeds): the recorded statistics, the bank and the prototypes on the network that TTA never adapted. No TTA and Ours: as in \cref{tab:main}. Bold: best.}
\label{tab:s_wotta}
\vspace{-2mm}
\small
\setlength{\tabcolsep}{4pt}
\fitwidth{\begin{tabular}{@{}l ccc ccc ccc ccc@{}}
\toprule
 & \multicolumn{3}{c}{ResNet, FT} & \multicolumn{3}{c}{ResNet, ER} & \multicolumn{3}{c}{ViT-B/16, FT} & \multicolumn{3}{c}{ViT-B/16, ER} \\
\cmidrule(lr){2-4}\cmidrule(lr){5-7}\cmidrule(lr){8-10}\cmidrule(lr){11-13}
 & \begin{tabular}{@{}c@{}}no\\TTA\end{tabular} & \begin{tabular}{@{}c@{}}w/o TTA\\method\end{tabular} & \textbf{Ours} & \begin{tabular}{@{}c@{}}no\\TTA\end{tabular} & \begin{tabular}{@{}c@{}}w/o TTA\\method\end{tabular} & \textbf{Ours} & \begin{tabular}{@{}c@{}}no\\TTA\end{tabular} & \begin{tabular}{@{}c@{}}w/o TTA\\method\end{tabular} & \textbf{Ours} & \begin{tabular}{@{}c@{}}no\\TTA\end{tabular} & \begin{tabular}{@{}c@{}}w/o TTA\\method\end{tabular} & \textbf{Ours} \\
\midrule
C100-C & 48.6 & 58.2 & \textbf{60.7} & 64.7 & 64.8 & \textbf{65.9} & 65.6 & 66.0 & \textbf{67.0} & 69.0 & 69.3 & \textbf{69.5} \\
C100-TF & 49.7 & 56.0 & \textbf{66.4} & 63.1 & 63.4 & \textbf{68.9} & 73.4 & 73.8 & \textbf{74.6} & 75.2 & 75.9 & \textbf{76.4} \\
C20 & 63.7 & \textbf{63.9} & \textbf{63.9} & 68.1 & 68.5 & \textbf{68.6} & 76.9 & 77.5 & \textbf{77.9} & 79.1 & \textbf{80.0} & 79.9 \\
DomainNet & 39.1 & 42.3 & \textbf{43.5} & 51.1 & \textbf{51.5} & 51.1 & 62.8 & 63.3 & \textbf{63.4} & 63.9 & \textbf{64.7} & \textbf{64.7} \\
Inet-TF & 42.1 & 62.5 & \textbf{76.7} & 70.8 & 71.7 & \textbf{80.0} & 95.7 & \textbf{96.1} & \textbf{96.1} & 96.0 & 96.1 & \textbf{96.3} \\
IN-C & 17.1 & 19.2 & \textbf{22.3} & 31.8 & 31.8 & \textbf{34.3} & 50.7 & 50.8 & \textbf{52.8} & 56.4 & 56.4 & \textbf{57.3} \\
\midrule
Mean & 43.4 & 50.4 & \textbf{55.6} & 58.3 & 58.6 & \textbf{61.4} & 70.8 & 71.3 & \textbf{72.0} & 73.3 & 73.7 & \textbf{74.0} \\
\bottomrule
\end{tabular}
}
\vspace{-2mm}
\end{table}

\begin{table}[H]
\centering
\caption{\textbf{Ablations, every cell} (score difference to \ours, points, mixed batches, Tent/ROID, paired by seed).}
\label{tab:s_ablation}
\vspace{-2mm}
\small
\setlength{\tabcolsep}{3.5pt}
\fitwidth{\begin{tabular}{@{}l cccccccccccccccc@{}}
\toprule
 & \multicolumn{8}{c}{ResNet} & \multicolumn{8}{c}{ViT-B/16} \\
\cmidrule(lr){2-9}\cmidrule(lr){10-17}
 & \multicolumn{2}{c}{C100-TF} & \multicolumn{2}{c}{Inet-TF} & \multicolumn{2}{c}{C20} & \multicolumn{2}{c}{DomainNet} & \multicolumn{2}{c}{C100-TF} & \multicolumn{2}{c}{Inet-TF} & \multicolumn{2}{c}{C20} & \multicolumn{2}{c}{DomainNet} \\
 & FT & ER & FT & ER & FT & ER & FT & ER & FT & ER & FT & ER & FT & ER & FT & ER \\
\midrule
\multicolumn{17}{@{}l}{\emph{Records and posterior}} \\
\quad only the latest domain's recorded statistics & $-$5.6 & $-$4.1 & $-$13.1 & $-$10.4 & \sm{$+$0.3} & \sm{$+$0.1} & $+$3.6 & \sm{$-$0.3} & \sm{$+$0.2} & \sm{0.0} & \sm{$+$0.1} & \sm{0.0} & \sm{$+$0.1} & \sm{$+$0.1} & \sm{$+$0.3} & \sm{0.0} \\
\quad statistics recorded on $\theta_\mathrm{tr}$, first layer & $-$4.4 & $-$0.7 & \sm{$-$0.2} & \sm{$-$0.1} & \sm{$-$0.3} & \sm{$+$0.2} & \sm{$+$0.4} & $-$0.6 & \sm{$-$0.1} & \sm{0.0} & \sm{$+$0.1} & \sm{0.0} & \sm{$+$0.4} & $+$0.8 & \sm{$+$0.2} & \sm{$+$0.1} \\
\quad statistics recorded on $\theta_\mathrm{tr}$, first quarter & $-$2.4 & $-$1.7 & $-$1.0 & \sm{$-$0.3} & \sm{$-$0.5} & \sm{$-$0.1} & \sm{$+$0.2} & $-$1.2 & \sm{$+$0.1} & \sm{$+$0.1} & \sm{$+$0.1} & \sm{0.0} & $+$0.7 & $+$0.9 & \sm{$+$0.3} & \sm{$+$0.1} \\
\quad statistics recorded on $\theta_\mathrm{tr}$, every layer & $-$2.0 & $-$1.6 & $-$0.7 & \sm{$-$0.2} & \sm{$-$0.1} & \sm{$-$0.1} & $+$1.2 & $-$1.0 & \sm{$+$0.1} & \sm{$+$0.1} & \sm{$+$0.1} & \sm{0.0} & $+$0.8 & $+$1.0 & \sm{$+$0.4} & \sm{$+$0.2} \\
\quad typicality rule, not the posterior & $-$7.0 & $-$2.2 & $-$3.9 & $-$1.1 & \sm{$+$0.5} & \sm{$-$0.1} & $+$2.8 & $-$0.7 & \sm{$-$0.1} & \sm{$+$0.1} & \sm{0.0} & \sm{0.0} & $+$1.0 & $+$1.1 & $+$0.6 & \sm{$+$0.4} \\
\quad true domain given (oracle) & $-$0.9 & $-$0.6 & \sm{$+$0.1} & \sm{0.0} & \sm{$+$0.1} & \sm{$+$0.2} & \sm{$-$0.5} & $-$0.8 & \sm{$+$0.1} & \sm{$-$0.1} & \sm{$+$0.1} & \sm{0.0} & $+$1.3 & $+$1.4 & \sm{$+$0.4} & \sm{$+$0.4} \\
\multicolumn{17}{@{}l}{\emph{Dose}} \\
\quad statistics only, affine shared & \sm{$-$0.5} & $-$1.1 & \sm{$+$0.2} & \sm{0.0} & \sm{$+$0.4} & \sm{$+$0.1} & \sm{$-$0.1} & $-$1.5 & \sm{$+$0.4} & \sm{$+$0.3} & \sm{$-$0.1} & \sm{$-$0.1} & $+$1.1 & $+$0.9 & \sm{$+$0.1} & \sm{$+$0.1} \\
\quad every tensor that TTA changed & $+$1.4 & \sm{0.0} & \sm{$+$0.1} & \sm{0.0} & \sm{$+$0.1} & \sm{0.0} & $+$3.4 & \sm{0.0} & \sm{$-$0.4} & \sm{$-$0.1} & \sm{0.0} & \sm{0.0} & \sm{$-$0.3} & \sm{$-$0.1} & \sm{$-$0.2} & \sm{$-$0.1} \\
\multicolumn{17}{@{}l}{\emph{Bank and bank check}} \\
\quad no bank & $-$5.9 & \sm{0.0} & $-$19.6 & \sm{0.0} & \sm{0.0} & \sm{0.0} & $-$2.5 & \sm{$-$0.1} & \sm{$-$0.1} & \sm{0.0} & \sm{$+$0.1} & \sm{0.0} & \sm{$-$0.1} & \sm{0.0} & \sm{0.0} & \sm{$-$0.1} \\
\quad bank always, without bank check & $+$1.0 & \sm{0.0} & \sm{0.0} & \sm{$+$0.1} & $-$4.9 & \sm{0.0} & $-$2.3 & \sm{$-$0.1} & \sm{0.0} & \sm{0.0} & \sm{$+$0.1} & \sm{0.0} & \sm{$-$0.3} & \sm{$+$0.1} & \sm{0.0} & \sm{$-$0.1} \\
\multicolumn{17}{@{}l}{\emph{Prototype head}} \\
\quad no prototype head & \sm{$-$0.4} & \sm{$-$0.1} & \sm{$-$0.2} & \sm{$-$0.1} & \sm{$-$0.5} & \sm{$-$0.1} & \sm{$-$0.1} & \sm{$-$0.2} & $-$0.6 & $-$0.7 & \sm{$-$0.1} & \sm{$-$0.2} & $-$1.1 & $-$0.8 & $-$0.6 & $-$0.7 \\
\multicolumn{17}{@{}l}{\emph{Next task}} \\
\quad next task from the adapted network (as \cltta) & \sm{$-$0.3} & \sm{0.0} & \sm{$+$0.1} & \sm{0.0} & \sm{$-$0.4} & \sm{0.0} & \sm{$-$0.1} & \sm{$-$0.1} & \sm{0.0} & \sm{$-$0.1} & \sm{$+$0.1} & \sm{0.0} & \sm{$-$0.3} & \sm{$-$0.1} & \sm{0.0} & \sm{$-$0.1} \\
\bottomrule
\end{tabular}
}
\vspace{-2mm}
\end{table}

\begin{table}[H]
\centering
\caption{\textbf{\cltta with statistics recorded during TTA} instead of those of the \evl{evaluation} batch (ResNet, mixed): applying the statistics of the \prv{preview} to every image collapses the old domains.}
\label{tab:s_bn}
\vspace{-2mm}
\small
\setlength{\tabcolsep}{4pt}
\fitwidth{\begin{tabular}{@{}ll cccccc@{}}
\toprule
 & & +TTA (test batch) & +TTA, stored stats & recorded, EMA & recorded, exact & recorded, all previews & \textbf{Ours} \\
\midrule
FT & C100-TF & 63.1 & 49.9 & 47.6 & 48.3 & 57.0 & 68.4 \\
 & C20 & 64.4 & 63.9 & -- & 64.3 & 62.3 & 64.6 \\
 & Inet-TF & 70.4 & 42.5 & 41.0 & 41.3 & 53.2 & 77.7 \\
 & DomainNet & 46.2 & 38.6 & 37.9 & 37.9 & 43.3 & 47.0 \\
\midrule
ER & C100-TF & 65.9 & 62.8 & 51.2 & 51.6 & 62.2 & 68.8 \\
 & C20 & 67.9 & 68.2 & -- & 68.3 & 65.5 & 68.7 \\
 & Inet-TF & 75.1 & 70.7 & 42.2 & 42.6 & 57.5 & 80.1 \\
 & DomainNet & 49.6 & 49.4 & 40.2 & 40.2 & 45.5 & 51.5 \\
\bottomrule
\end{tabular}}
\label{sec:s_bn}
\vspace{-3mm}
\end{table}

\subsection{Held-out images, other CL methods and other benchmarks}
\Cref{tab:heldout} scores the preview domain on held-out images that were never in the stream. \Cref{tab:othercl} replaces ER and FT by DER++ and CODA-Prompt, and \cref{tab:inra} gives ImageNet-R and ImageNet-A on ViTs.
With DER++~\cite{buzzega2020derpp}, which adds to replay a distillation of the logits stored with the buffer images, \ours scores highest among the configurations in \cref{tab:othercl}.
The reference comes from the same run of \ours, where the next task starts from $\theta_\mathrm{tr}$, so that TTA never changes what DER++ keeps training. DER++ alone is an independent DER++ run that never uses TTA during the sequence, and it agrees with the reference within $0.3$ points. With CODA-Prompt~\cite{smith2023coda}, which learns prompts on a frozen ViT without any buffer, \ours is above CODA-Prompt with TTA on all four benchmarks ($+0.3$ to $+0.9$), and above CODA-Prompt alone on three ($-0.4$ on DomainNet). Notably, Continual Normalization~\cite{pham2022cn}, which adapts BatchNorm to online CL by adding a per-image GroupNorm before every BatchNorm layer, remains below \ours ($53.5$ against $58.7$, \cref{sec:s_cn}).

\begin{table}[H]
\centering
\caption{\textbf{Held-out \prv{preview} domain}: score with the \prv{preview} domain scored on held-out images never in the stream, and the accuracy of \ours on the \prv{preview} domain on the stream and on held-out images. Bold: best of the systems without test-batch statistics; \ours is best or tied in \nHeldWins{} of \nHeldCells{} settings.}
\label{tab:heldout}
\vspace{-2mm}
\small
\setlength{\tabcolsep}{3pt}
\fitwidth{\begin{tabular}{@{}ll cccc cc@{}}
\toprule
 & & \multicolumn{4}{c}{Score, preview domain held out} & \multicolumn{2}{c}{Preview domain, \ours} \\
\cmidrule(lr){3-6}\cmidrule(lr){7-8}
 & & no TTA & +TTA & {\scriptsize +TTA$^\mathrm{batch}$} & \textbf{Ours} & stream & held-out \\
\midrule
ResNet, FT & C100-C & 49.4 & 43.5 & \textcolor{black!50}{57.8} & \textbf{62.1} & 67.9 & 68.8 \\
 & C100-TF & 49.1 & 51.6 & \textcolor{black!50}{63.8} & \textbf{68.5} & 68.7 & 66.8 \\
 & C20 & 63.7 & 64.5 & \textcolor{black!50}{64.6} & 64.1 & 44.9 & 42.9 \\
 & DomainNet & 36.2 & 34.7 & \textcolor{black!50}{44.7} & \textbf{42.0} & 37.9 & 33.3 \\
 & Inet-TF & 43.6 & 43.5 & \textcolor{black!50}{70.5} & \textbf{76.6} & 74.9 & 73.9 \\
 & IN-C & 17.6 & 15.1 & \textcolor{black!50}{22.8} & \textbf{23.1} & 40.3 & 40.4 \\
\midrule
ResNet, ER & C100-C & 65.0 & 47.3 & \textcolor{black!50}{62.6} & \textbf{66.1} & 67.9 & 68.6 \\
 & C100-TF & 62.9 & 55.1 & \textcolor{black!50}{66.6} & \textbf{69.1} & 68.1 & 66.3 \\
 & C20 & 67.8 & 67.7 & \textcolor{black!50}{67.6} & \textbf{68.2} & 46.9 & 44.6 \\
 & DomainNet & 49.3 & 36.9 & \textcolor{black!50}{48.4} & \textbf{49.5} & 41.5 & 38.0 \\
 & Inet-TF & 67.4 & 45.7 & \textcolor{black!50}{73.4} & \textbf{78.6} & 76.0 & 73.6 \\
 & IN-C & 32.2 & 15.6 & \textcolor{black!50}{28.2} & \textbf{34.8} & 38.0 & 38.1 \\
\midrule
ViT-B/16, FT & C100-C & 65.6 & 66.2 &  & \textbf{67.1} & 70.4 & 70.8 \\
 & C100-TF & 72.7 & 74.4 &  & \textbf{74.9} & 72.5 & 70.7 \\
 & C20 & 75.5 & 76.0 &  & \textbf{77.2} & 57.1 & 55.7 \\
 & DomainNet & 61.5 & 61.8 &  & \textbf{62.3} & 56.4 & 52.8 \\
 & Inet-TF & 95.7 & 95.7 &  & \textbf{95.9} & 94.2 & 93.9 \\
 & IN-C & 51.2 & 52.3 &  & \textbf{52.5} & 56.1 & 55.4 \\
\midrule
ViT-B/16, ER & C100-C & 68.8 & 69.4 &  & \textbf{69.7} & 71.7 & 72.2 \\
 & C100-TF & 74.6 & 76.1 &  & \textbf{76.6} & 73.3 & 71.8 \\
 & C20 & 77.6 & 77.8 &  & \textbf{78.9} & 58.5 & 56.6 \\
 & DomainNet & 62.2 & 62.4 &  & \textbf{63.0} & 56.7 & 52.8 \\
 & Inet-TF & 95.8 & 95.7 &  & \textbf{96.0} & 94.1 & 93.8 \\
 & IN-C & 56.8 & 57.0 &  & \textbf{57.7} & 57.8 & 57.4 \\
\bottomrule
\end{tabular}
}
\vspace{-2mm}
\end{table}

\begin{table}[H]
\centering
\caption{\textbf{Other CL methods} (score, mixed batches, Tent/ROID, \nSeeds{} seeds). Reference: for DER++, the network immediately before each \prv{preview} within a run of \ours, which may retain effects of TTA from earlier tasks; for CODA-Prompt, the standalone run without TTA. DER++ alone: an independent run of DER++ that never uses TTA during the whole sequence, within $0.3$ points of the reference.
+TTA: \cltta with its own statistics (batch-independent); +TTA$^\mathrm{batch}$ (gray): with the \evl{evaluation} batch's statistics (test-batch protocol; on ViTs both coincide).}
\label{tab:othercl}
\vspace{-2mm}
\small
\setlength{\tabcolsep}{4pt}
\fitwidth{\begin{tabular}{@{}ll ccccc@{}}
\toprule
 CL method & Benchmark & Reference & {\scriptsize DER++ alone} & +TTA & {\scriptsize +TTA$^\mathrm{batch}$} & \textbf{Ours} \\
\midrule
DER++ (ResNet) & C100-TF & 62.2 & 62.2 & 52.1 & \textcolor{black!50}{66.7} & \textbf{69.1} \\
 & Inet-TF & 68.0 & 67.7 & 42.7 & \textcolor{black!50}{74.6} & \textbf{77.8} \\
 & DomainNet & 50.1 & 50.1 & 41.6 & \textcolor{black!50}{50.1} & \textbf{50.7} \\
\midrule
DER++ (ViT-B/16) & C100-TF & 76.5 & 76.4 & 77.5 & \textcolor{black!50}{77.5} & \textbf{78.0} \\
 & C20 & 80.2 & 80.2 & 80.4 & \textcolor{black!50}{80.4} & \textbf{81.1} \\
 & Inet-TF & 96.4 & 96.4 & 96.6 & \textcolor{black!50}{96.6} & \textbf{96.9} \\
 & DomainNet & 66.2 & 66.3 & 66.3 & \textcolor{black!50}{66.3} & \textbf{66.9} \\
\midrule
CODA-Prompt (ViT-B/16) & C100-TF & 75.3 & -- & 75.3 & \textcolor{black!50}{75.3} & \textbf{76.2} \\
 & C20 & 76.2 & -- & 76.4 & \textcolor{black!50}{76.4} & \textbf{77.0} \\
 & Inet-TF & 95.6 & -- & 96.1 & \textcolor{black!50}{96.1} & \textbf{96.4} \\
 & DomainNet & \textbf{62.9} & -- & 61.9 & \textcolor{black!50}{61.9} & 62.6 \\
\bottomrule
\end{tabular}
}
\vspace{-3mm}
\end{table}

\begin{table}[H]
\centering
\caption{\textbf{ImageNet-R / ImageNet-A} (ViT, Tent/ROID, \nSeeds{} seeds). With 1 to 3 \prv{preview} images per class, the prototypes are weak, and TTA has little to give.}
\label{tab:inra}
\vspace{-2mm}
\small
\setlength{\tabcolsep}{4pt}
\fitwidth{\begin{tabular}{@{}ll cccc@{}}
\toprule
 & & no TTA & +TTA & \ours & \ours without TTA \\
\midrule
FT & IN-A & 28.0 & 27.9 & 27.9 & 27.9 \\
FT & IN-R & 66.3 & 66.3 & 66.5 & 66.2 \\
ER & IN-A & 28.7 & 28.7 & 28.5 & 28.8 \\
ER & IN-R & 66.2 & 66.1 & 66.3 & 66.2 \\
\bottomrule
\end{tabular}
}
\vspace{-3mm}
\end{table}

\subsection{Preview length and cost}
\label{sec:s_len}
\begin{figure}[H]
\centering
\input{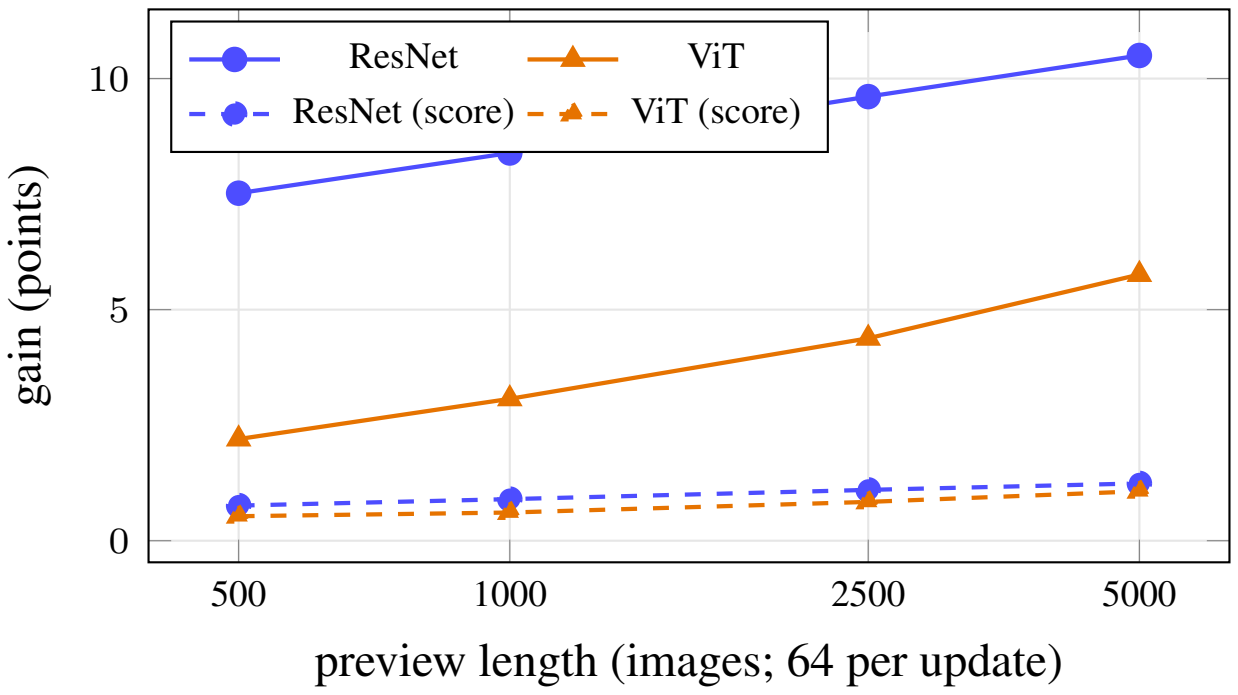}
\begin{tikzpicture}
\begin{axis}[width=0.5\textwidth, height=0.3\textwidth, xlabel={\small preview length (images; 64 per update)}, ylabel={\small gain (points)},
  xmode=log, log basis x=10, xtick={500,1000,2500,5000}, xticklabels={500,1000,2500,5000}, tick label style={font=\scriptsize}, grid=both, grid style={black!10},
  legend style={font=\scriptsize, at={(0.02,0.98)}, anchor=north west}, legend columns=2]
\addplot[mark=*, blue!70, thick] coordinates {\lenresnetgainafter};
\addplot[mark=triangle*, orange!90!black, thick] coordinates {\lenvitgainafter};
\addplot[mark=*, blue!70, thick, dashed] coordinates {\lenresnetgainscore};
\addplot[mark=triangle*, orange!90!black, thick, dashed] coordinates {\lenvitgainscore};
\legend{ResNet, ViT, ResNet (score), ViT (score)}
\end{axis}
\end{tikzpicture}
\vspace{-2mm}
\caption{\textbf{\prv{Preview} length} (C100-C, ER, \ours with Tent/ROID, \nSeeds{} seeds). Solid: accuracy of the adapted network on the \prv{preview} domain after the pass minus before it. Dashed: score of \ours minus the score of the same network before the \prv{preview}.}
\label{fig:previewlen}
\vspace{-3mm}
\end{figure}

From 500 to 5000 preview images on C100-C, the TTA gain on the preview grows from $7.5$ to $10.5$ points on ResNets and from $2.2$ to $5.8$ points on ViTs (\cref{fig:previewlen}).
The official SAR code~\cite{niu2023sar} shows the same effect on ImageNet-C with a ViT-B/16 and batches of 64. Averaged over four blur corruptions, its online accuracy is $38.2\%$ after 5{,}000 images and $51.1\%$ after the 50{,}000 images of the usual protocol (Tent: $38.1$ and $51.0$). After 5{,}000 images, our implementations give the same accuracy ($38.3$ and $38.0$). \Cref{tab:latency} gives the test-time cost of \ours.

\begin{table}[H]
\centering
\caption{\textbf{Test-time cost}: forward time of the deployed system per image (ms), measured on one RTX A4500 with nothing else running, after the \prv{preview} of the last task without a buffer, the costliest point of a run (every trained domain has its recorded statistics and a bank entry; domains: number of trained domains). Each row uses the architecture, input size, \evl{evaluation} batch size (mixed) and number of classes of its benchmark. No TTA: $\theta_\mathrm{tr}$ with its stored statistics. +TTA: \cltta with the statistics estimated during the \prv{preview}. \ours includes the frozen first block, the posterior, the dosed BatchNorm layers and the prototype head. Per-domain BatchNorm with the inferred domain costs the same as \ours ($1.46$ and $1.60$ times \cltta on ResNet-50 and ResNeXt-29). The last two columns give the throughput in images per second (1000 divided by the time per image). Times are measured with the released implementation, separately from the runs.}
\label{tab:latency}
\vspace{-2mm}
\small
\setlength{\tabcolsep}{4pt}
\fitwidth{\begin{tabular}{@{}ll cc ccc c cc@{}}
\toprule
 & & input & domains & no TTA & +TTA & \textbf{Ours} & Ours / +TTA & +TTA (img/s) & \textbf{Ours} (img/s) \\
\midrule
ResNet & C100-C & 32\,px & 14 & 0.24 & 0.24 & 0.40 & 1.67$\times$ & 4\,167 & 2\,500 \\
 & C100-TF & 32\,px & 5 & 0.24 & 0.24 & 0.39 & 1.62$\times$ & 4\,167 & 2\,564 \\
 & C20 & 112\,px & 4 & 0.23 & 0.23 & 0.34 & 1.46$\times$ & 4\,348 & 2\,941 \\
 & DomainNet & 224\,px & 5 & 0.80 & 0.80 & 1.17 & 1.46$\times$ & 1\,250 & 855 \\
 & Inet-TF & 112\,px & 5 & 0.23 & 0.23 & 0.34 & 1.46$\times$ & 4\,348 & 2\,941 \\
 & IN-C & 224\,px & 14 & 0.80 & 0.80 & 1.19 & 1.48$\times$ & 1\,250 & 840 \\
\midrule
ViT-B/16 & C100-C & 224\,px & 14 & 3.62 & 3.63 & 3.72 & 1.02$\times$ & 275 & 269 \\
 & C100-TF & 224\,px & 5 & 3.67 & 3.68 & 3.74 & 1.02$\times$ & 272 & 267 \\
 & C20 & 224\,px & 4 & 3.70 & 3.71 & 3.77 & 1.01$\times$ & 270 & 265 \\
 & DomainNet & 224\,px & 5 & 3.73 & 3.74 & 3.80 & 1.01$\times$ & 267 & 263 \\
 & Inet-TF & 224\,px & 5 & 3.76 & 3.76 & 3.81 & 1.01$\times$ & 266 & 262 \\
 & IN-C & 224\,px & 14 & 3.77 & 3.77 & 3.85 & 1.02$\times$ & 265 & 260 \\
\bottomrule
\end{tabular}
}
\vspace{-3mm}
\end{table}

\subsection{Per-task results}
\label{sec:s_pertask}
As is usual in continual TTA, \cref{tab:pertask_c100c,tab:pertask_inc} give the accuracy at every task of C100-C and IN-C, and \cref{tab:pertask_seg} the mIoU at every task of Cityscapes $\to$ ACDC. Task $t$ trains on domain $t$ and previews domain $t{+}1$, whose name heads the column.

On IN-C, the CL method alone often falls to chance on the preview corruption, for instance on defocus blur after impulse noise or on contrast after brightness: the stored statistics of one corruption do not fit the next. TTA recovers $17$ to $25$ points on the preview, and \ours keeps this gain ($38.0$ against $37.5$ with ER, $40.3$ against $41.0$ with FT). On the trained domains, \cltta collapses as soon as a noise is followed by a blur, while \ours stays within $0.1$ points of the CL method alone with ER and $2.3$ points above it with FT.
IN-C is also hard for the CL method alone itself. Each task gives 25 labeled images per class, trained for one epoch, and without a buffer the stored statistics follow the latest corruption, so the older ones fall to about $11\%$.
On C100-C, TTA gains $10$ to $23$ points on the preview, and \ours keeps all but $1$ to $3$ of them while matching or beating the CL method alone on the trained domains.
In segmentation, TTA gains little on the preview conditions except at night, and BN Stats Adapt and Tent lose $5$ to $7$ points on fog. With ER, \ours keeps the trained conditions at the level of the CL method alone. Without a buffer, it loses on them in the tasks where its bank check keeps an outdated bank.

\begin{table}[H]
\centering
\caption{\textbf{C100-C per task} (ResNeXt-29, accuracy, mixed batches, Tent and ROID averaged). +TTA: \cltta with the statistics estimated during the \prv{preview}. +TTA$^\mathrm{batch}$ (gray): with the statistics of each test batch, the usual protocol of continual TTA.}
\label{tab:pertask_c100c}
\small
\setlength{\tabcolsep}{2.5pt}
\fitwidth{\begin{tabular}{@{}ll cccccccccccccc c@{}}
\toprule
 & task $t$ (previewed domain) & 0 & 1 & 2 & 3 & 4 & 5 & 6 & 7 & 8 & 9 & 10 & 11 & 12 & 13 & mean \\
 & & shot & impul & defoc & glass & motn & zoom & snow & frost & fog & brit & contr & elast & pixel & jpeg & \\
\midrule
\multicolumn{17}{@{}l}{\emph{FT, previewed domain $n(t)$}} \\
 & CL alone & 65.9 & 62.3 & 55.2 & 8.9 & 55.6 & 71.6 & 23.8 & 63.1 & 42.2 & 51.7 & 35.1 & 33.4 & 27.5 & 44.8 & 45.8 \\
 & +TTA & 67.9 & 67.5 & 74.0 & 63.5 & 71.7 & 74.0 & 69.0 & 69.7 & 67.9 & 74.8 & 72.7 & 66.2 & 71.4 & 63.3 & 69.6 \\
 & {\scriptsize +TTA$^\mathrm{batch}$} & \textcolor{black!50}{67.2} & \textcolor{black!50}{67.2} & \textcolor{black!50}{67.7} & \textcolor{black!50}{60.1} & \textcolor{black!50}{66.0} & \textcolor{black!50}{68.7} & \textcolor{black!50}{63.1} & \textcolor{black!50}{65.1} & \textcolor{black!50}{59.1} & \textcolor{black!50}{71.4} & \textcolor{black!50}{49.2} & \textcolor{black!50}{62.0} & \textcolor{black!50}{62.7} & \textcolor{black!50}{61.4} & \textcolor{black!50}{63.6} \\
 & \textbf{\ours} & 66.4 & 67.8 & 74.3 & 62.8 & 71.7 & 74.2 & 68.9 & 68.4 & 67.9 & 67.6 & 73.5 & 58.8 & 70.2 & 58.0 & 67.9 \\
\midrule
\multicolumn{17}{@{}l}{\emph{FT, trained domains, mean of $a_j(t)$}} \\
 & CL alone & 65.8 & 67.4 & 64.1 & 28.1 & 50.7 & 46.2 & 40.0 & 51.3 & 52.8 & 47.6 & 50.7 & 35.6 & 48.8 & 38.2 & 49.1 \\
 & +TTA & 65.7 & 62.8 & 6.2 & 43.9 & 31.8 & 31.6 & 37.0 & 38.8 & 39.6 & 40.4 & 29.5 & 43.4 & 39.1 & 45.2 & 39.6 \\
 & {\scriptsize +TTA$^\mathrm{batch}$} & \textcolor{black!50}{65.4} & \textcolor{black!50}{64.2} & \textcolor{black!50}{59.4} & \textcolor{black!50}{58.6} & \textcolor{black!50}{55.3} & \textcolor{black!50}{55.1} & \textcolor{black!50}{55.3} & \textcolor{black!50}{56.4} & \textcolor{black!50}{55.0} & \textcolor{black!50}{53.4} & \textcolor{black!50}{54.8} & \textcolor{black!50}{53.0} & \textcolor{black!50}{51.9} & \textcolor{black!50}{52.3} & \textcolor{black!50}{56.4} \\
 & \textbf{\ours} & 65.7 & 66.1 & 61.1 & 63.4 & 62.0 & 63.8 & 62.8 & 63.2 & 62.7 & 61.0 & 51.7 & 60.3 & 50.0 & 46.7 & 60.0 \\
\midrule
\multicolumn{17}{@{}l}{\emph{ER, previewed domain $n(t)$}} \\
 & CL alone & 66.5 & 62.3 & 57.4 & 52.5 & 65.9 & 71.7 & 50.7 & 63.8 & 49.1 & 63.6 & 51.1 & 56.4 & 37.8 & 55.1 & 57.4 \\
 & +TTA & 67.8 & 67.1 & 73.0 & 61.7 & 70.7 & 72.7 & 67.6 & 68.7 & 66.5 & 73.6 & 71.2 & 63.7 & 69.8 & 62.3 & 68.3 \\
 & {\scriptsize +TTA$^\mathrm{batch}$} & \textcolor{black!50}{66.6} & \textcolor{black!50}{65.2} & \textcolor{black!50}{68.5} & \textcolor{black!50}{58.1} & \textcolor{black!50}{68.5} & \textcolor{black!50}{71.4} & \textcolor{black!50}{61.3} & \textcolor{black!50}{66.1} & \textcolor{black!50}{57.9} & \textcolor{black!50}{70.0} & \textcolor{black!50}{55.6} & \textcolor{black!50}{60.8} & \textcolor{black!50}{59.2} & \textcolor{black!50}{61.1} & \textcolor{black!50}{63.6} \\
 & \textbf{\ours} & 66.7 & 67.5 & 73.8 & 62.3 & 71.0 & 73.3 & 68.1 & 68.5 & 66.5 & 69.0 & 72.7 & 60.6 & 69.8 & 61.3 & 67.9 \\
\midrule
\multicolumn{17}{@{}l}{\emph{ER, trained domains, mean of $a_j(t)$}} \\
 & CL alone & 65.8 & 67.0 & 66.5 & 67.4 & 66.0 & 66.6 & 66.8 & 65.8 & 65.8 & 64.2 & 65.5 & 64.6 & 64.9 & 63.4 & 65.8 \\
 & +TTA & 65.4 & 62.0 & 6.1 & 51.4 & 37.8 & 31.9 & 43.4 & 47.2 & 43.8 & 45.6 & 35.4 & 50.5 & 47.0 & 51.4 & 44.2 \\
 & {\scriptsize +TTA$^\mathrm{batch}$} & \textcolor{black!50}{64.9} & \textcolor{black!50}{64.1} & \textcolor{black!50}{62.0} & \textcolor{black!50}{64.8} & \textcolor{black!50}{62.3} & \textcolor{black!50}{62.9} & \textcolor{black!50}{63.2} & \textcolor{black!50}{62.7} & \textcolor{black!50}{61.1} & \textcolor{black!50}{60.6} & \textcolor{black!50}{60.9} & \textcolor{black!50}{61.0} & \textcolor{black!50}{59.9} & \textcolor{black!50}{59.4} & \textcolor{black!50}{62.1} \\
 & \textbf{\ours} & 66.0 & 67.1 & 66.4 & 67.2 & 65.8 & 66.6 & 66.5 & 65.5 & 65.3 & 64.2 & 65.2 & 64.5 & 64.8 & 63.2 & 65.6 \\
\bottomrule
\end{tabular}
}
\end{table}

\begin{table}[H]
\centering
\caption{\textbf{IN-C per task} (ResNet-50, severity 5, accuracy, mixed batches, Tent and ROID averaged). Rows as in \cref{tab:pertask_c100c}. With 1{,}000 classes, $0.1\%$ is chance. It is a collapse of the predictions, not of the evaluation: at task 2 of the run of seed 0 without memory, one class receives $19\%$ of all predictions of the CL method alone, against $1\%$ at task 0, because the stored statistics of impulse noise do not fit defocused images. Re-estimating the statistics on the same images, as TTA does, brings them back to $28.5\%$.}
\label{tab:pertask_inc}
\small
\setlength{\tabcolsep}{2.5pt}
\fitwidth{\begin{tabular}{@{}ll cccccccccccccc c@{}}
\toprule
 & task $t$ (previewed domain) & 0 & 1 & 2 & 3 & 4 & 5 & 6 & 7 & 8 & 9 & 10 & 11 & 12 & 13 & mean \\
 & & shot & impul & defoc & glass & motn & zoom & snow & frost & fog & brit & contr & elast & pixel & jpeg & \\
\midrule
\multicolumn{17}{@{}l}{\emph{FT, previewed domain $n(t)$}} \\
 & CL alone & 35.7 & 35.4 & 0.1 & 3.7 & 14.1 & 31.8 & 2.1 & 25.3 & 8.4 & 7.4 & 0.6 & 0.4 & 28.6 & 33.0 & 16.2 \\
 & +TTA & 44.7 & 43.2 & 28.5 & 35.0 & 38.3 & 44.1 & 37.3 & 36.4 & 44.9 & 53.9 & 29.8 & 44.0 & 48.0 & 45.4 & 41.0 \\
 & {\scriptsize +TTA$^\mathrm{batch}$} & \textcolor{black!50}{43.7} & \textcolor{black!50}{42.3} & \textcolor{black!50}{20.3} & \textcolor{black!50}{26.8} & \textcolor{black!50}{27.1} & \textcolor{black!50}{33.6} & \textcolor{black!50}{23.8} & \textcolor{black!50}{28.2} & \textcolor{black!50}{27.0} & \textcolor{black!50}{46.6} & \textcolor{black!50}{10.0} & \textcolor{black!50}{31.1} & \textcolor{black!50}{40.4} & \textcolor{black!50}{38.5} & \textcolor{black!50}{31.4} \\
 & \textbf{\ours} & 42.8 & 39.4 & 28.6 & 32.7 & 35.7 & 42.5 & 39.0 & 37.3 & 46.1 & 51.0 & 32.8 & 41.9 & 48.4 & 46.2 & 40.3 \\
\midrule
\multicolumn{17}{@{}l}{\emph{FT, trained domains, mean of $a_j(t)$}} \\
 & CL alone & 43.5 & 44.0 & 42.5 & 11.0 & 11.2 & 15.8 & 18.7 & 8.5 & 11.8 & 9.7 & 11.4 & 5.0 & 11.0 & 11.1 & 18.2 \\
 & +TTA & 40.4 & 43.0 & 0.1 & 3.2 & 9.4 & 12.9 & 1.9 & 5.7 & 4.9 & 6.8 & 1.9 & 6.0 & 7.3 & 11.7 & 11.1 \\
 & {\scriptsize +TTA$^\mathrm{batch}$} & \textcolor{black!50}{40.5} & \textcolor{black!50}{43.7} & \textcolor{black!50}{23.5} & \textcolor{black!50}{18.2} & \textcolor{black!50}{17.3} & \textcolor{black!50}{17.0} & \textcolor{black!50}{15.2} & \textcolor{black!50}{16.7} & \textcolor{black!50}{15.1} & \textcolor{black!50}{13.3} & \textcolor{black!50}{16.3} & \textcolor{black!50}{14.4} & \textcolor{black!50}{16.2} & \textcolor{black!50}{17.0} & \textcolor{black!50}{20.3} \\
 & \textbf{\ours} & 43.3 & 43.3 & 42.6 & 16.9 & 15.2 & 15.3 & 18.8 & 14.6 & 13.8 & 15.3 & 11.4 & 13.9 & 11.0 & 11.3 & 20.5 \\
\midrule
\multicolumn{17}{@{}l}{\emph{ER, previewed domain $n(t)$}} \\
 & CL alone & 34.5 & 37.6 & 0.1 & 5.6 & 15.9 & 28.4 & 3.5 & 22.9 & 17.2 & 31.1 & 3.7 & 14.9 & 34.5 & 35.2 & 20.4 \\
 & +TTA & 44.1 & 42.3 & 27.6 & 32.4 & 34.9 & 39.0 & 33.3 & 31.6 & 40.3 & 49.7 & 24.8 & 39.9 & 44.7 & 40.9 & 37.5 \\
 & {\scriptsize +TTA$^\mathrm{batch}$} & \textcolor{black!50}{42.9} & \textcolor{black!50}{41.2} & \textcolor{black!50}{19.8} & \textcolor{black!50}{26.5} & \textcolor{black!50}{29.4} & \textcolor{black!50}{35.8} & \textcolor{black!50}{21.5} & \textcolor{black!50}{28.3} & \textcolor{black!50}{27.9} & \textcolor{black!50}{40.4} & \textcolor{black!50}{7.9} & \textcolor{black!50}{29.5} & \textcolor{black!50}{39.8} & \textcolor{black!50}{37.7} & \textcolor{black!50}{30.6} \\
 & \textbf{\ours} & 42.3 & 40.2 & 27.9 & 31.1 & 34.4 & 38.5 & 35.9 & 33.9 & 42.8 & 50.0 & 27.2 & 40.0 & 46.0 & 42.5 & 38.0 \\
\midrule
\multicolumn{17}{@{}l}{\emph{ER, trained domains, mean of $a_j(t)$}} \\
 & CL alone & 43.3 & 44.3 & 44.0 & 40.5 & 36.2 & 34.2 & 33.5 & 32.3 & 32.0 & 29.1 & 28.0 & 29.5 & 27.5 & 27.6 & 34.4 \\
 & +TTA & 39.1 & 42.8 & 0.1 & 5.2 & 9.7 & 13.8 & 3.2 & 9.7 & 5.1 & 8.9 & 1.7 & 9.3 & 9.0 & 16.0 & 12.4 \\
 & {\scriptsize +TTA$^\mathrm{batch}$} & \textcolor{black!50}{39.1} & \textcolor{black!50}{42.7} & \textcolor{black!50}{23.1} & \textcolor{black!50}{30.5} & \textcolor{black!50}{26.2} & \textcolor{black!50}{26.8} & \textcolor{black!50}{24.5} & \textcolor{black!50}{26.3} & \textcolor{black!50}{19.8} & \textcolor{black!50}{21.3} & \textcolor{black!50}{20.3} & \textcolor{black!50}{21.7} & \textcolor{black!50}{21.8} & \textcolor{black!50}{21.4} & \textcolor{black!50}{26.1} \\
 & \textbf{\ours} & 43.3 & 44.3 & 44.0 & 40.4 & 36.1 & 34.0 & 33.6 & 32.3 & 31.9 & 28.9 & 27.8 & 29.4 & 27.4 & 27.2 & 34.3 \\
\bottomrule
\end{tabular}
}
\end{table}

\begin{table}[H]
\centering
\caption{\textbf{Cityscapes $\to$ ACDC per task} (DeepLabV3+ R50, mIoU, 5 seeds). Left: the \prv{preview} condition. Right: the mean over the trained domains (Cityscapes and the conditions trained so far). +TTA: \cltta with the statistics estimated during the \prv{preview}.}
\label{tab:pertask_seg}
\small
\setlength{\tabcolsep}{3.5pt}
\fitwidth{\begin{tabular}{@{}ll cccc c cccc c@{}}
\toprule
 & & \multicolumn{5}{c}{previewed condition, task $t$} & \multicolumn{5}{c}{trained domains (mean), task $t$} \\
\cmidrule(lr){3-7}\cmidrule(lr){8-12}
 & & 0 fog & 1 night & 2 rain & 3 snow & mean & 0 & 1 & 2 & 3 & mean \\
\midrule
FT & CL alone & 61.0 & 17.2 & 39.2 & 50.0 & 41.8 & 74.5 & 57.4 & 49.9 & 52.9 & 58.7 \\
 & BN adapt, +TTA & 54.2 & 25.9 & 47.7 & 53.1 & 45.2 & 62.8 & 32.5 & 49.4 & 49.2 & 48.5 \\
 & BN adapt, \ours & 53.9 & 26.3 & 46.6 & 51.4 & 44.5 & 74.5 & 58.5 & 40.1 & 52.6 & 56.4 \\
 & Tent, +TTA & 56.2 & 27.0 & 48.2 & 51.9 & 45.8 & 62.1 & 30.8 & 48.5 & 47.1 & 47.1 \\
 & Tent, \ours & 55.8 & 27.8 & 47.1 & 52.5 & 45.8 & 74.5 & 53.5 & 47.0 & 48.8 & 56.0 \\
 & CoTTA, +TTA & 64.0 & 19.1 & 41.9 & 52.5 & 44.4 & 74.6 & 59.4 & 51.0 & 55.7 & 60.2 \\
 & CoTTA, \ours & 64.0 & 17.9 & 41.9 & 51.3 & 43.8 & 74.6 & 58.0 & 44.9 & 54.7 & 58.1 \\
\midrule
ER & CL alone & 62.4 & 20.8 & 47.6 & 54.4 & 46.3 & 75.7 & 71.4 & 59.5 & 61.2 & 66.9 \\
 & BN adapt, +TTA & 55.7 & 31.7 & 54.0 & 55.9 & 49.3 & 63.5 & 57.5 & 55.8 & 56.5 & 58.3 \\
 & BN adapt, \ours & 55.5 & 31.7 & 52.7 & 56.1 & 49.0 & 75.7 & 71.0 & 59.8 & 61.0 & 66.9 \\
 & Tent, +TTA & 56.5 & 31.0 & 54.3 & 54.7 & 49.1 & 62.2 & 54.3 & 53.4 & 54.5 & 56.1 \\
 & Tent, \ours & 56.3 & 31.4 & 53.5 & 55.0 & 49.0 & 75.7 & 71.6 & 59.4 & 61.2 & 67.0 \\
 & CoTTA, +TTA & 64.6 & 23.5 & 50.9 & 55.5 & 48.6 & 75.7 & 71.8 & 61.2 & 61.6 & 67.6 \\
 & CoTTA, \ours & 64.6 & 23.3 & 50.4 & 55.4 & 48.4 & 75.7 & 71.8 & 59.6 & 61.7 & 67.2 \\
\bottomrule
\end{tabular}
}
\end{table}

\section{Additional analyses}
\label{sec:s_more}

\subsection{Same statistics for every method}
\label{sec:s_protocols}
Since TTA methods are usually evaluated with the statistics of the test batch, which our main protocol forbids, we check that the gain of \ours does not come from the protocol. In \cref{tab:protocols}, every method gets the same statistics. With the statistics of each evaluation batch, which mixes all evaluated domains (\cref{sec:eval_protocols}), \ours is above \cltta by $4.4$ points without memory and $3.6$ with a buffer. On single-domain batches (Oracle), \cltta scores $7.0$ points more without memory and $2.7$ with a buffer than on mixed ones, with the same model. In that case, \ours reduces to \cltta with the prototype head and is at most $0.6$ points below it in all 12 settings, and up to $1.4$ above. Without memory, \ours with its own statistics stays below the Oracle, by up to $13.7$ points on IN-C, because single-domain batches hand \cltta fresh statistics of every old domain, which memory-free CL cannot keep. With a buffer, \ours is at most $0.5$ points below it. UnMix-TNS~\cite{tomar2024unmixtns}, built for correlated labels within one domain, is below \cltta with batch statistics in all 12 settings, and below \ours in all 12.

\begin{table}[H]
\centering
\caption{\textbf{Protocols} (ResNet, score averaged over the six benchmarks, Tent and ROID). Each row says where the statistics come from and how batches are formed, and both methods get the same statistics. UnMix-TNS runs around ROID. On the same ROID runs, \cltta scores $42.9$/$45.4$ (batch-independent) and $54.4$/$58.0$ (test-batch), and \ours $55.8$/$61.6$ (FT/ER).}
\label{tab:protocols}
\vspace{-2mm}
\small
\setlength{\tabcolsep}{3pt}
\fitwidth{\begin{tabular}{llcccc}
\toprule
 & & \multicolumn{2}{c}{FT} & \multicolumn{2}{c}{ER} \\
\cmidrule(lr){3-4}\cmidrule(lr){5-6}
statistics & batches & \cltta & \ours & \cltta & \ours \\
\midrule
evaluation batch & mixed & 53.9 & \textbf{58.3} & 58.0 & \textbf{61.6} \\
evaluation batch & single-domain & 60.9 & \textbf{61.2} & 60.7 & \textbf{61.3} \\
preview (ours) & either & 41.9 & \textbf{55.8} & 44.4 & \textbf{61.6} \\
\midrule
\multicolumn{2}{l}{UnMix-TNS around ROID, components frozen} & 41.7 & & 43.4 & \\
\multicolumn{2}{l}{UnMix-TNS around ROID, components adapting (test batch)} & 47.1 & & 49.2 & \\
\midrule
\multicolumn{2}{l}{no TTA (CL method alone)} & \multicolumn{2}{c}{43.4} & \multicolumn{2}{c}{58.3} \\
\bottomrule
\end{tabular}
}
\vspace{-3mm}
\end{table}

\subsection{Why the next task starts from \texorpdfstring{$\theta_\mathrm{tr}$}{theta\_tr}}
\label{sec:s_c100c}
With the next task from the adapted network, as in \cltta, \ours on C100-C without memory reaches $60$ to $61$ points with Norm, CoTTA and POEM, but $41$ to $50$ with Tent, RoTTA, SAR, CMF, ROID and EATA (\cref{tab:s_thetatr}). RoTTA changes the structure of the network, so that without a buffer its changes are shared and \ours equals \cltta. With the five other methods, the older domains collapse right after the preview, from the second task on. With ROID, domains 0 and 1 fall to $14.9\%$ after the preview of task 2, against $56.1\%$ with Tent at the rate of the PALM benchmark ($2{\cdot}10^{-5}$), while their posterior weight on the preview is about $3\%$ with every method and the latest domain is unaffected. The damage therefore comes from the statistics that the bank gives the older domains. These five methods adapt with a learning rate of $10^{-3}$ on CIFAR, and C100-C has the longest preview of the CIFAR benchmarks (5,000 images, against 1,000 for C100-TF). The next task then carries their larger change, and the statistics stored in the bank for the older domains no longer match their activations. The bank check cannot see it, since it scores images of the latest domain only, which the bank barely affects. Therefore, \ours starts the next task from $\theta_\mathrm{tr}$ (\cref{sec:dosing}).
\Cref{tab:c100c} changes one choice at a time, at the official rates. Starting the next task from $\theta_\mathrm{tr}$ removes the collapse, and the three methods then score $60.3$ to $60.9$, above \ours without a TTA method ($58.2$). A preview of 1,000 images removes more than half of it ($54.8$ to $56.5$). Without the bank, the older domains no longer collapse at task 2, but memory-free CL then forgets them over the following tasks, as without \ours.

\begin{table}[H]
\centering
\caption{\textbf{C100-C without memory} (ResNeXt-29, \nSeeds{} seeds, official rates: $10^{-3}$ for Tent, ROID and EATA). $A$: score. Old 0-1: accuracy on domains 0 and 1 right after the preview of task 2. Adapted: the next task starts from the adapted network, as in \cltta. From $\theta_\mathrm{tr}$: the setting of \cref{tab:main}. Preview 1,000: 1,000 random images of each preview, next task from the adapted network. No bank: \texttt{--dose-bank off}, next task from the adapted network.}
\label{tab:c100c}
\small
\setlength{\tabcolsep}{4pt}
\begin{tabular}{@{}l cc cc cc@{}}
\toprule
 & \multicolumn{2}{c}{ROID} & \multicolumn{2}{c}{EATA} & \multicolumn{2}{c}{Tent} \\
\cmidrule(lr){2-3}\cmidrule(lr){4-5}\cmidrule(lr){6-7}
 & $A$ & old 0-1 & $A$ & old 0-1 & $A$ & old 0-1 \\
\midrule
adapted & 47.5 & 14.9 & 48.3 & 21.7 & 43.4 & 10.2 \\
from $\theta_\mathrm{tr}$ & \textbf{60.9} & 56.9 & \textbf{60.4} & 56.0 & \textbf{60.3} & 57.3 \\
preview 1,000 & 55.1 & 41.7 & 56.5 & 48.6 & 54.8 & 45.3 \\
no bank & 48.0 & 55.0 & 48.2 & 54.4 & 46.6 & 54.2 \\
\bottomrule
\end{tabular}
\end{table}

\paragraph{Both starts on C100-C and IN-C.} \Cref{tab:s_thetatr} compares the two starts with every TTA method of \cref{tab:main}, with and without memory, with Tent at $10^{-3}$ on both sides, on the same runs and seeds. Starting from $\theta_\mathrm{tr}$ needs no additional storage, since \ours already keeps what TTA changed as a difference to $\theta_\mathrm{tr}$. On ResNets without memory, it brings the median on C100-C from $48.3$ to $60.7$, above \ours without a TTA method ($58.2$), and on IN-C from $21.4$ to $22.3$. The other cells change by $0.0$ to $+1.4$ points. Method by method, it is at least as good in 45 of the 56 pairs of method and setting, while CoTTA on ViTs loses $8.6$ points on IN-C without memory. \cltta keeps the adapted start, as TTA methods are usually deployed.

\begin{table}[H]
\centering
\caption{\textbf{Next task from the adapted network or from $\theta_\mathrm{tr}$} (score of \ours, \%, medians over the TTA methods of \cref{tab:main}, mixed batches). Adapted: the next task starts from the adapted network, as in \cltta. $\theta_\mathrm{tr}$: as in \cref{tab:main}. Tent at $10^{-3}$ on both sides, $5$ seeds per method ($3$ for Norm). Last row: methods for which $\theta_\mathrm{tr}$ is at least as good, over both benchmarks.}
\label{tab:s_thetatr}
\vspace{-2mm}
\small
\setlength{\tabcolsep}{4pt}
\fitwidth{\begin{tabular}{@{}l cc cc cc cc@{}}
\toprule
 & \multicolumn{2}{c}{ResNet, FT} & \multicolumn{2}{c}{ResNet, ER} & \multicolumn{2}{c}{ViT-B/16, FT} & \multicolumn{2}{c}{ViT-B/16, ER} \\
\cmidrule(lr){2-3}\cmidrule(lr){4-5}\cmidrule(lr){6-7}\cmidrule(lr){8-9}
 & adapted & $\theta_\mathrm{tr}$ & adapted & $\theta_\mathrm{tr}$ & adapted & $\theta_\mathrm{tr}$ & adapted & $\theta_\mathrm{tr}$ \\
\midrule
C100-C & 48.3 & \textbf{60.7} & 64.5 & \textbf{65.9} & 66.7 & \textbf{67.0} & \textbf{69.5} & \textbf{69.5} \\
IN-C & 21.4 & \textbf{22.3} & 33.5 & \textbf{34.3} & 52.5 & \textbf{52.8} & 56.7 & \textbf{57.3} \\
\midrule
{\scriptsize methods, $\theta_\mathrm{tr}\geq$ adapted} & \multicolumn{2}{c}{13 / 16} & \multicolumn{2}{c}{13 / 16} & \multicolumn{2}{c}{9 / 12} & \multicolumn{2}{c}{10 / 12} \\
\bottomrule
\end{tabular}
}
\vspace{-2mm}
\end{table}

\subsection{TTA learning rate and cost}
\label{sec:s_ttalr}
Every TTA method keeps its official learning rate, which its authors tuned on their own benchmarks and backbones (for Tent on CIFAR, the value of the CoTTA code, \cref{tab:s_tta}). In our setup, the preview domain has no labels before its task, so that this rate cannot be tuned, and it may be too large for a new backbone, benchmark or CL method. Notably, a too large learning rate can make the TTA method collapse on the preview, and \cltta then predicts every image with the collapsed network. Therefore, we test how each method copes with a wrong learning rate by multiplying the official one by ten (\cref{tab:x10}). In that case, several methods collapse under \cltta, such as CoTTA on ViT C100-TF ($1.6$), while \ours stays above \cltta in \nXTenWin{} of \nXTen{} pairs and above the CL method alone in \nXTenAlone{}.
A smaller rate does not protect \cltta either, since its damage on ResNets comes from the statistics of the preview and not from the size of the update: on C100-C, Tent at $2{\cdot}10^{-5}$ instead of $10^{-3}$ gives $43.8$ against $42.0$ without memory and $43.0$ against $45.2$ with a buffer. It only helps the adapted start of \ours, which scores $60.3$ against $45.0$ without memory, whereas the start from $\theta_\mathrm{tr}$ scores $60.7$ at $10^{-3}$ (\cref{sec:s_c100c}).
Regarding cost, \ours adds one pass through the frozen first block per image. At the last task, an image costs $1.46$ to $1.67$ times the forward time of \cltta on ResNets and $1.01$ to $1.02$ times on ViTs (\cref{tab:latency}).

\subsection{Stability against plasticity}
\label{sec:s_balanced}
The score $A$ averages the accuracy over all domains seen so far. Wang \etal~\cite{wang2024ccl} note that the average accuracy of continual learning mostly rewards stability, and that plasticity needs its own measure. The same holds for $A$, where the preview domain weighs only $1/(t{+}2)$ at task $t$. \Cref{tab:balanced} therefore reports a balanced score $B=\frac{1}{2}(\text{old}+\text{new})$, which gives the preview domain the same weight as all trained domains together.
Under $B$, the gains of \ours grow. On average over the six benchmarks, \ours is above the CL method alone by $15.8$ and $5.7$ points on ResNets (FT and ER) and by $1.7$ points on ViTs with both, against $10.0$, $2.8$, $0.8$ and $0.5$ points under $A$. \ours is at least as good as the CL method alone in all 24 settings and as \cltta in 22 of them. \cltta with test-batch statistics is above \ours only on three memory-free ResNet benchmarks (C100-C, DomainNet and C20).

\begin{table}[H]
\centering
\caption{\textbf{Balanced score} $B=\frac{1}{2}(\text{old}+\text{new})$ (\%, mixed batches, median over the TTA methods of \cref{tab:main}). Old is the mean accuracy on the trained domains and new the accuracy on the \prv{preview} domain, both averaged over tasks. Columns and systems are those of \cref{tab:main}, and bold marks the best system that uses no statistic of the \evl{evaluation} batch.}
\label{tab:balanced}
\vspace{-2mm}
\small
\setlength{\tabcolsep}{3pt}
\fitwidth{\begin{tabular}{@{}ll ccccccc@{}}
\toprule
 & & C100-C & C100-TF & C20 & DomainNet & Inet-TF & IN-C & Mean \\
\midrule
ResNet, FT & no TTA & 47.4 & 46.2 & 59.8 & 35.9 & 36.5 & 17.2 & 40.5 \\
 & +TTA & 54.4 & 55.3 & \textbf{60.3} & 39.5 & 51.0 & 26.1 & 47.8 \\
 & {\scriptsize +TTA$^\mathrm{batch}$} & \textcolor{black!50}{60.3} & \textcolor{black!50}{63.0} & \textcolor{black!50}{60.4} & \textcolor{black!50}{44.9} & \textcolor{black!50}{69.0} & \textcolor{black!50}{26.6} & \textcolor{black!50}{54.0} \\
 & \textbf{Ours} & \textbf{63.7} & \textbf{67.8} & 60.1 & \textbf{43.3} & \textbf{77.1} & \textbf{30.6} & \textbf{57.1} \\
\midrule
ResNet, ER & no TTA & 61.6 & 60.1 & 63.6 & 48.8 & 65.8 & 27.4 & 54.5 \\
 & +TTA & 55.0 & 57.2 & 63.4 & 41.2 & 52.4 & 24.8 & 49.0 \\
 & {\scriptsize +TTA$^\mathrm{batch}$} & \textcolor{black!50}{62.5} & \textcolor{black!50}{64.6} & \textcolor{black!50}{63.1} & \textcolor{black!50}{48.0} & \textcolor{black!50}{72.9} & \textcolor{black!50}{29.4} & \textcolor{black!50}{56.8} \\
 & \textbf{Ours} & \textbf{65.4} & \textbf{68.9} & \textbf{64.2} & \textbf{49.6} & \textbf{79.5} & \textbf{34.8} & \textbf{60.4} \\
\midrule
ViT-B/16, FT & no TTA & 65.1 & 72.7 & 71.8 & 61.7 & 95.2 & 50.1 & 69.4 \\
 & +TTA & 67.3 & 73.2 & 72.1 & 61.6 & 95.3 & \textbf{55.3} & 70.8 \\
 & \textbf{Ours} & \textbf{67.5} & \textbf{74.2} & \textbf{73.3} & \textbf{62.8} & \textbf{95.8} & 53.9 & \textbf{71.2} \\
\midrule
ViT-B/16, ER & no TTA & 67.8 & 74.2 & 73.9 & 62.5 & 95.6 & 54.0 & 71.3 \\
 & +TTA & 69.1 & 74.8 & 74.0 & 62.7 & 95.4 & \textbf{57.7} & 72.3 \\
 & \textbf{Ours} & \textbf{69.6} & \textbf{76.0} & \textbf{76.1} & \textbf{64.0} & \textbf{96.0} & 57.3 & \textbf{73.2} \\
\bottomrule
\end{tabular}
}
\vspace{-3mm}
\end{table}

\subsection{Semantic segmentation: Cityscapes \texorpdfstring{$\to$}{to} ACDC}
\label{sec:s_seg}
From Cityscapes to the ACDC conditions (\cref{tab:seg}), DeepLabV3+ shows the BatchNorm problem of ResNets. With ER, \ours keeps the score of the CL method alone with every TTA method, while BN Stats Adapt and Tent cost \cltta $5$ to $7$ points. With FT, \ours gains $3$ to $5$ points over \cltta with these two but stays below the CL method alone, and loses $1.8$ points with CoTTA, as its bank restores outdated statistics (\cref{sec:s_pertask}). On SegFormer-B5, which uses LayerNorm, TTA barely harms the trained conditions, as on ViTs, and \ours keeps the gain of SVDP ($+0.6$ points over the CL method alone).

\begin{table}[H]
\centering
\caption{\textbf{Cityscapes $\to$ ACDC} (score, mIoU \%, DeepLabV3+ with a ResNet-50 and SegFormer-B5). +TTA uses the statistics estimated during the \prv{preview}. Tent is the original method on DeepLabV3+, and on SegFormer, whose only BatchNorm layers are in its head, the version released with the official CoTTA code~\cite{wang2022cotta} for this benchmark, which adapts the affine parameters of every normalization layer. \ours without prototypes (see text). --: not run.}
\label{tab:seg}
\vspace{-2mm}
\small
\setlength{\tabcolsep}{3pt}
\fitwidth{\begin{tabular}{@{}l cc cc cc cc@{}}
\toprule
 & \multicolumn{4}{c}{DeepLabV3+ (BatchNorm)} & \multicolumn{4}{c}{SegFormer-B5 (LayerNorm)} \\
\cmidrule(lr){2-5}\cmidrule(lr){6-9}
 & \multicolumn{2}{c}{FT} & \multicolumn{2}{c}{ER} & \multicolumn{2}{c}{FT} & \multicolumn{2}{c}{ER} \\
\cmidrule(lr){2-3}\cmidrule(lr){4-5}\cmidrule(lr){6-7}\cmidrule(lr){8-9}
no TTA & \multicolumn{2}{c}{52.8} & \multicolumn{2}{c}{60.0} & \multicolumn{2}{c}{65.8} & \multicolumn{2}{c}{66.1} \\
\midrule
TTA & +TTA & \textbf{\ours} & +TTA & \textbf{\ours} & +TTA & \textbf{\ours} & +TTA & \textbf{\ours} \\
\midrule
BN adapt & 46.9 & 50.2 & 55.1 & 60.3 & 65.7 & 65.9 & 65.7 & 66.0 \\
Tent & 46.3 & 51.7 & 53.5 & 60.5 & 65.7 & 65.8 & 66.0 & 66.0 \\
CoTTA & 54.8 & 53.0 & 61.2 & 60.9 & -- & -- & -- & -- \\
SVDP & -- & -- & -- & -- & 66.7 & 66.4 & 67.0 & 66.7 \\
\bottomrule
\end{tabular}
}
\vspace{-3mm}
\end{table}

We apply the same protocol to dense prediction. The network is DeepLabV3+ with a ResNet-50 backbone, which uses BatchNorm, initialized from the official mmsegmentation Cityscapes weights. It is trained task after task on Cityscapes and then on the ACDC conditions fog, night and rain, for one epoch each with SGD on crops of $512\times512$ pixels, and ER keeps 200 images. Each task previews the next condition on its test split, whose 500 images per condition have no public labels. The score is the mIoU over 19 classes on the validation split of every trained domain and of the preview one, averaged over domains and tasks. Images are used at half resolution. BN Stats Adapt, Tent and CoTTA, the three baselines of the Cityscapes-to-ACDC experiment of CoTTA, follow its official code, with one image per step. Its Tent keeps the network in evaluation mode, adapts the affine parameters of every normalization layer and minimizes the cross-entropy of its own predictions. CoTTA averages its teacher over 7 scales and their flips. This Tent was written for SegFormer, whose BatchNorm layers are only in its head. On DeepLabV3+ we therefore also run the original Tent, which normalizes each test image with its own statistics and minimizes the entropy, as in our classification experiments. CoTTA keeps the stored statistics, so its two protocols coincide. We also run SegFormer-B5, the network of the Cityscapes-to-ACDC benchmark of CoTTA, with the official mmsegmentation Cityscapes weights. Its encoder uses LayerNorm, and its only BatchNorm layers are the five of its head. It is trained with AdamW at a tenth of its official rate, as DeepLabV3+ with SGD. On SegFormer we run BN Stats Adapt, Tent and SVDP~\cite{yang2024svdp}, a recent method with official code for this benchmark. SVDP adds sparse learnable prompts to the most uncertain pixels of the input. We use the settings of its official run (prompts on $0.1\%$ of the pixels, learning rate $10^{-4}$), and each image is predicted with prompts placed on its own uncertainty, as in the official code. Its prompt map is defined for the size of the ACDC images, so we resize it for Cityscapes images. \ours is unchanged, except that it has no prototype head and that its two checks compare images by pixel accuracy. The prototype head would keep one prototype per class, the mean feature of the preview pixels weighted by their predicted probability of that class, and score every pixel against these prototypes. We tested it, and it lowers the score in 13 of the 14 settings of \cref{tab:seg}, by $1.0$ point on average and mostly on the preview condition. We therefore leave it out. We use 5 seeds.
With the Tent of CoTTA's code on DeepLabV3+, \cltta scores $51.2$ with FT and $58.5$ with ER, and \ours $50.8$ and $59.9$. This Tent leaves the statistics untouched, so there is little for \ours to dose. On SegFormer, with BN Stats Adapt and Tent, \cltta and \ours stay within $0.4$ points of the CL method alone in our protocol (\cref{tab:seg}). With the statistics of each test image instead, BN Stats Adapt costs $0.7$ points without a buffer and $1.4$ with one. The five BatchNorm layers of the head are all that \ours can dose. SVDP changes every weight and adds prompts. It raises the preview condition by $2.8$ points over the CL method alone without hurting the trained ones, with and without a buffer, and \ours stays within $0.3$ points of \cltta ($66.4$ and $66.7$ against $66.7$ and $67.0$).

\subsection{Along a long stream}
\begin{figure}[H]
\centering
\includegraphics[width=0.8\textwidth]{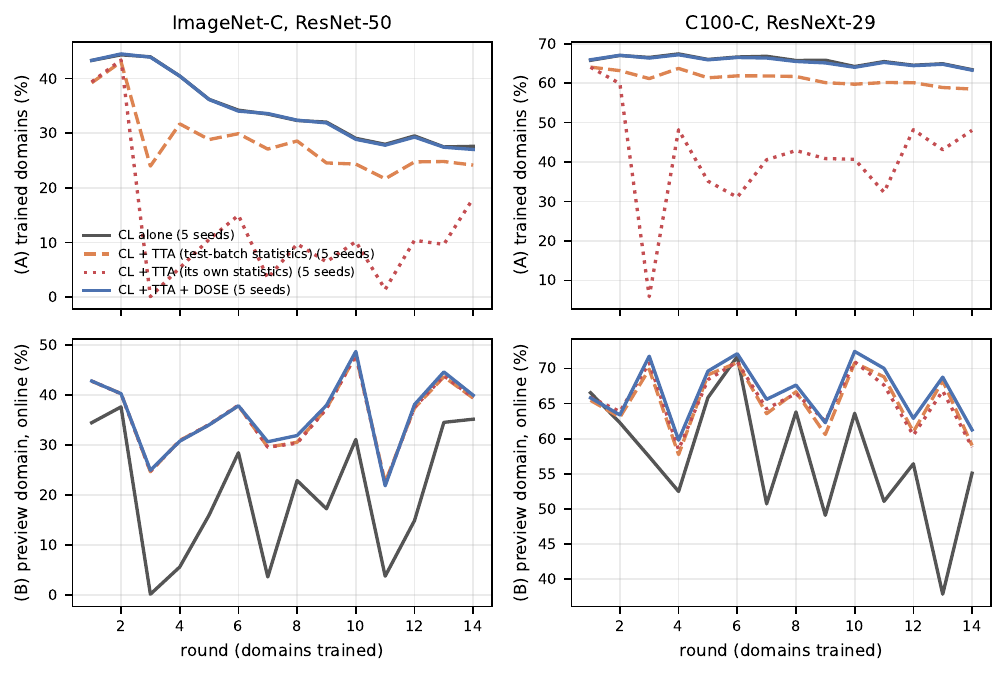}
\vspace{-2mm}
\caption{\textbf{Along the 14-domain streams} (ER, Tent, mixed batches). (A) Accuracy on the trained domains after each \prv{preview}. (B) Online accuracy on the \prv{preview} domain during its stream. Each \prv{preview} batch is predicted before the adaptation on that batch: Tent predicts it, normalized with its own statistics, then updates on it. \ours and \cltta share these online predictions, and the CL method alone predicts with its unchanged network.}
\label{fig:stream}
\vspace{-3mm}
\end{figure}

\Cref{fig:stream} follows the two sequences of 14 domains task by task, with ER and Tent. On the trained domains (A), \ours follows the CL method alone, which itself declines on ImageNet-C as CL forgets. \cltta falls about $20$ points below it at the third domain with the statistics of the test batch, and it collapses with its own statistics. On the preview domain (B), \ours keeps the online accuracy of TTA during the preview, far above the CL method alone. It thus uses the adaptation without paying for it on the trained domains.
With NOTE on C100-C (\nSeeds{} seeds), \ours ties \cltta ($54.8$ against $54.8$), and both are $10$ points below the CL method alone ($64.7$). NOTE replaces every BatchNorm layer by its own instance-aware normalization, and \ours changes little around it, as for NOTE in \cref{tab:methods}. On IN-C, \ours is $1.1$ points above \cltta ($24.1$ against $23.0$), and both stay far below the CL method alone ($31.8$).

\subsection{What the recorded statistics see}
\begin{figure}[H]
\centering
\includegraphics[width=\textwidth]{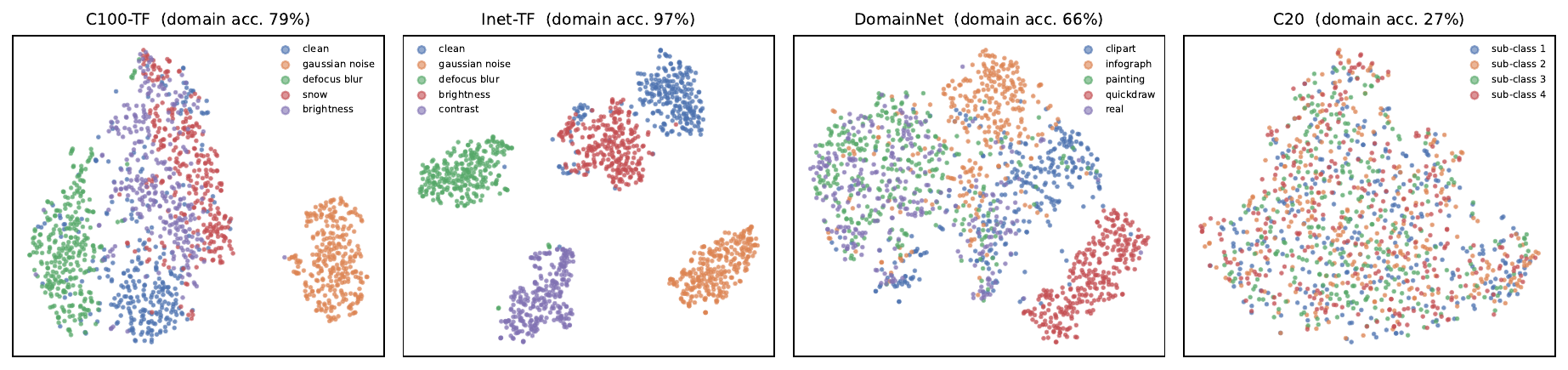}
\vspace{-5mm}
\caption{\textbf{The domains as seen by the recorded statistics.} t-SNE of the vectors $s(x)$ (channel means and log-variances of the input and of the frozen first block) for 300 \trn{training} images per domain. In parentheses is the domain accuracy of the posterior on held-out images.}
\label{fig:tsne}
\vspace{-3mm}
\end{figure}

\Cref{fig:tsne} projects the vectors $s(x)$ of the training images of each domain, computed on the network input and the frozen first block. Corruptions and styles form separate clusters. The sub-class domains of C20 overlap, which is why the bank check drops the bank and \ours performs similarly to the CL method alone there (\cref{sec:method}). We fit the posterior of \cref{sec:method} on half of the images of each domain. It finds the domain of the other half for $97\%$ of the images on Inet-TF, $79\%$ on C100-TF and $66\%$ on DomainNet, but only $27\%$ on C20, where chance is $25\%$.

\subsection{Stress test: a preview mixed with trained domains}
\label{sec:s_mix}
Our protocol spends the preview on the next domain, as in a targeted rollout where the data of the next task is collected. Adapting to an uncurated stream in which recurring and new domains mix is the problem of general continual TTA, such as NOTE~\cite{gong2022note}, RoTTA~\cite{yuan2023rotta} or ReservoirTTA~\cite{vray2025reservoirtta}, which \ours can wrap (\cref{tab:methods}). We nevertheless evaluate this harder case. We add to each preview as many images of the trained domains as it holds, in equal shares per domain, so that half of the stream is old. They come from the halves of the held-out sets that the task does not score. The preview domain is still scored on its own images.
Such a stream changes what TTA learns. \cltta gains up to $24$ points, but not from the new domain: its statistics now also describe the trained domains, which it sees again. The recorded statistics and the BatchNorm statistics of the preview, in turn, describe a mixture. \ours can however use its recorded statistics during the preview. With the \emph{filter}, an image that the recorded statistics of a trained domain explain better than those of the preview so far is left out of the TTA update, the preview's record and its statistics. This is the posterior of \cref{sec:method} with the images seen so far, so it needs no threshold. The filter starts after 100 images.
\Cref{tab:mixpreview} gives the scores (Tent and ROID, ER and FT, \nSeeds{} seeds). %
With half old images, \ours stays the best system on TF/Inet and on C100-C ($62.6$ against $54.4$ for \cltta). On DomainNet and C20, it falls $1.4$ and $0.6$ points below \cltta, which profits from seeing old images again, and the filter does not help there. The filter drops $83\%$ of the old images but also $33\%$ of the new ones (\nSeeds{} seeds). %
On the pure preview, it drops $26\%$ of the images ($6\%$ on Inet-TF, $59\%$ on C20) and costs at most $0.5$ points. A ViT keeps no domain statistics that old images could pull toward a mixture. With half old images, \cltta and \ours move by at most $0.6$ points on ViTs, and \ours stays above \cltta. Mixed previews on ResNets thus call for a sample-selection or stream-clustering step in front of \ours, which we leave to future work.

\begin{table}[H]
\centering
\caption{\textbf{A \prv{preview} that is half old images} (score $A$, Tent and ROID, ER and FT, \nSeeds{} seeds; a ViT has no BatchNorm statistics, so its two \cltta protocols coincide). Pure: the protocol of the paper. Half old: each \prv{preview} also holds as many images of the trained domains. The filter leaves out of the TTA update the images that the recorded statistics of a trained domain explain better than those of the \prv{preview}.}
\label{tab:mixpreview}
\vspace{-2mm}
\small
\fitwidth{\begin{tabular}{@{}ll cccc cc@{}}
\toprule
 & & \multicolumn{4}{c}{ResNet} & \multicolumn{2}{c}{ViT-B/16} \\
\cmidrule(lr){3-6}\cmidrule(lr){7-8}
preview & system & TF/Inet & C100-C & DN & C20 & TF/Inet & C20 \\
\midrule
pure & no TTA & 56.4 & 56.7 & 45.1 & 65.9 & 85.1 & 78.0 \\
 & +TTA & 46.3 & 45.8 & 39.1 & 66.2 & 85.8 & 78.3 \\
 & {\scriptsize +TTA$^\mathrm{batch}$} & 68.8 & 59.9 & 47.9 & 66.1 & -- & -- \\
 & \ours & 73.1 & 63.5 & 47.6 & 66.3 & 86.3 & 79.2 \\
 & \ours + filter & 73.1 & 63.3 & 47.2 & 65.8 & 86.2 & 79.0 \\
\midrule
half old & +TTA & 69.3 & 54.4 & 48.9 & 67.0 & 86.0 & 78.9 \\
 & {\scriptsize +TTA$^\mathrm{batch}$} & 69.9 & 54.7 & 48.1 & 66.4 & -- & -- \\
 & \ours & 71.5 & 62.6 & 47.5 & 66.4 & 86.4 & 79.5 \\
 & \ours + filter & 72.6 & 63.1 & 46.9 & 65.9 & 86.2 & 79.4 \\
\bottomrule
\end{tabular}
}
\vspace{-3mm}
\end{table}

\subsection{ViT gains and preview length}
\label{sec:vitdata}
On ViTs, the prompt and adapter methods built for them (DPCore, DePT and ViDA) move the preview by less than $1$ point in 6 to 78 updates. The methods that adapt LayerNorm with stronger losses (ROID, CMF, DAS and EATA) gain $6$ to $14$ points on corrupted previews (\cref{tab:vitgain}) and lose at most $1.8$ points on the trained domains. Yet the score rises by at most $3.7$ points, because the preview weighs only $0.17$ (IN-C) to $0.29$ (C100-TF) of it. The gain of TTA follows the amount of preview data. Over every TTA method and without memory, its median on the preview is $7.9$ points on IN-C (20{,}000 preview images), $2.8$ on C100-C (5{,}000), $0.9$ on C100-TF (1{,}000) and $0.3$ on Inet-TF (about 400). On C100-C, the gain grows from $2.2$ to $5.8$ points between 500 and 5{,}000 images (\cref{fig:previewlen}), and the official SAR code on ImageNet-C reaches $51\%$ after the 50{,}000 images of the usual protocol, against $38\%$ after 5{,}000 (\cref{sec:s_len}). Longer previews provide both more images and more adaptation updates; these experiments do not isolate their respective effects. Labeled data is scarce too. IN-C and C100-C give 25 and 33 labeled images per class and task (\cref{tab:data}), and the CL method alone scores lowest on them.

\subsection{Dosing what TTA changes on a ViT}
A ViT has no domain statistics, and TTA methods change its LayerNorm parameters, prompts or adapters. \ours shares these changes unless the shared-parameter check doses them (\cref{sec:method}). On ViTs, dosing every changed tensor by $\lambda$ changes the score by $-0.1$ and $-0.2$ points (\cref{tab:ablation}). The reason is that these changes barely hurt the trained domains of a ViT, by at most $2.4$ points on the domain just learned (\cref{tab:oldnew}), so there is little to protect. On C100-C and IN-C, where TTA gains most on ViTs, dosing every changed tensor costs $0.9$ and $0.7$ points ($1.3$ without a buffer), so sharing remains the better default.

The prototypes, in contrast, need their weight $\lambda$ on ViTs. We gave them a weight of one on every image (Tent/ROID, \nSeeds{} seeds). They then gain up to $1.4$ points on C20 and C100-C, where the trained and preview domains look alike. On ImageNet-R and ImageNet-A they lose $0.3$ to $1.6$ points in each setting ($37$ of $40$ runs), mostly on the trained domains,
 and on ImageNet-A they fall below the CL method alone. Averaged over the seven benchmarks, the gain is only $0.2$ points. The weight $\lambda$ keeps the prototypes off the images of the trained domains, and this is what makes them safe.

\subsection{Three training epochs per task}
\label{sec:s_epochs}
\Cref{tab:epochs} trains every task for three epochs instead of one, with everything else unchanged (ER, Tent/ROID, 5 seeds). On average, the CL method alone gains at most $0.7$ points. \ours stays above the CL method alone by $4.2$ points on ResNets and $0.9$ on ViTs, against $4.0$ and $1.1$ with one epoch. It stays above \cltta by $17.1$ and $0.7$ points, against $16.5$ and $0.6$. The single epoch of our protocol therefore does not create the gain of \ours.

\begin{table}[H]
\centering
\caption{\textbf{One against three CL epochs per task} (score, ER, mixed batches, mean of Tent and ROID, 5 seeds). +TTA: \cltta with its own statistics.}
\label{tab:epochs}
\vspace{-2mm}
\small
\setlength{\tabcolsep}{4pt}
\fitwidth{\begin{tabular}{@{}ll ccc ccc@{}}
\toprule
 & & \multicolumn{3}{c}{1 epoch} & \multicolumn{3}{c}{3 epochs} \\
\cmidrule(lr){3-5}\cmidrule(lr){6-8}
 & & no TTA & +TTA & \ours & no TTA & +TTA & \ours \\
\midrule
ResNet & C100-TF & 63.1 & 52.1 & 69.1 & 61.7 & 50.3 & 68.6 \\
ResNet & C20 & 68.1 & 68.1 & 68.6 & 67.0 & 66.9 & 67.1 \\
ResNet & DomainNet & 51.1 & 40.2 & 51.3 & 52.9 & 42.1 & 53.3 \\
ResNet & Inet-TF & 70.8 & 42.5 & 80.1 & 71.4 & 42.3 & 80.8 \\
ResNet & Mean & 63.3 & 50.7 & 67.3 & 63.3 & 50.4 & 67.4 \\
\midrule
ViT-B/16 & C100-TF & 75.2 & 76.6 & 77.2 & 75.7 & 76.6 & 77.4 \\
ViT-B/16 & C20 & 79.1 & 79.4 & 80.2 & 78.7 & 78.8 & 79.5 \\
ViT-B/16 & DomainNet & 63.9 & 64.2 & 64.8 & 66.0 & 66.1 & 66.9 \\
ViT-B/16 & Inet-TF & 96.0 & 95.8 & 96.2 & 96.4 & 95.8 & 96.4 \\
ViT-B/16 & Mean & 78.5 & 79.0 & 79.6 & 79.2 & 79.3 & 80.1 \\
\bottomrule
\end{tabular}
}
\vspace{-3mm}
\end{table}

\subsection{BatchNorm-aware CL}
\label{sec:s_cn}
On ResNets, the damage of TTA comes from the BatchNorm statistics that all domains share (\cref{sec:analysis}). Therefore, we also run two CL variants that change how BatchNorm is used, with everything else unchanged (ER and FT, Tent/ROID, 5 seeds). Continual Normalization (CN)~\cite{pham2022cn} applies a GroupNorm without affine parameters (32 groups) before every BatchNorm layer of the pretrained network. Cumulative statistics (Cum.) replace the running average of BatchNorm by the average over every training batch so far. Since Cum. only changes how the CL training estimates the statistics, we run it as the CL method alone.

Without memory, CN raises the CL method alone from $43.4$ to $50.2$ points, while with a buffer it is $1.5$ points below BatchNorm ($56.8$ against $58.3$, see \cref{tab:cn}). Cum. is below CN in both cases. Notably, CN makes \cltta less harmful, but it does not remove the damage. With a buffer, TTA still lowers the accuracy on the trained domains from $57.9$ to $47.9$ points on DomainNet and from $31.8$ to $21.2$ on IN-C, and \ours brings it back to $56.9$ and $31.7$. On CN, \ours is at least as good as CN alone in 21 of 24 pairs of TTA method and setting, and as CN with TTA in 20. Overall, \ours on BatchNorm remains the best system. It is at least as good as the better of CN and Cum. alone in 24 of 24 pairs, and scores $58.7$ against $53.5$ for CN alone on average.

\begin{table}[H]
\centering
\caption{\textbf{BatchNorm-aware CL} (ResNet, score, mixed batches, mean of Tent and ROID, 5 seeds). BatchNorm: the runs of \cref{tab:main} with these two TTA methods, which therefore differ from its medians. Cum.: cumulative BatchNorm statistics. CN: Continual Normalization. +TTA: \cltta with its own statistics. Bold marks the best system.}
\label{tab:cn}
\vspace{-2mm}
\small
\setlength{\tabcolsep}{3pt}
\fitwidth{\begin{tabular}{@{}l ccc c ccc ccc c ccc@{}}
\toprule
 & \multicolumn{7}{c}{FT} & \multicolumn{7}{c}{ER} \\
\cmidrule(lr){2-8}\cmidrule(lr){9-15}
 & \multicolumn{3}{c}{BatchNorm} & Cum. & \multicolumn{3}{c}{CN} & \multicolumn{3}{c}{BatchNorm} & Cum. & \multicolumn{3}{c}{CN} \\
\cmidrule(lr){2-4}\cmidrule(lr){5-5}\cmidrule(lr){6-8}\cmidrule(lr){9-11}\cmidrule(lr){12-12}\cmidrule(lr){13-15}
 & no TTA & +TTA & \textbf{Ours} & no TTA & no TTA & +TTA & \textbf{Ours} & no TTA & +TTA & \textbf{Ours} & no TTA & no TTA & +TTA & \textbf{Ours} \\
\midrule
C100-C & 48.6 & 43.8 & \textbf{61.0} & 55.0 & 54.8 & 53.1 & 54.8 & 64.7 & 47.7 & \textbf{65.9} & 60.6 & 61.0 & 56.0 & 61.7 \\
C100-TF & 49.7 & 49.3 & \textbf{66.6} & 54.5 & 58.3 & 60.8 & 62.7 & 63.1 & 52.1 & \textbf{69.1} & 61.3 & 62.9 & 63.1 & 66.0 \\
C20 & 63.7 & \textbf{64.3} & 64.1 & 58.9 & 61.2 & 61.7 & 60.6 & 68.1 & 68.1 & \textbf{68.6} & 61.3 & 64.8 & 65.1 & 65.2 \\
DomainNet & 39.1 & 38.0 & \textbf{43.8} & 41.8 & 39.9 & 40.2 & 42.1 & 51.1 & 40.2 & \textbf{51.3} & 46.9 & 49.6 & 43.5 & 49.9 \\
Inet-TF & 42.1 & 41.2 & \textbf{76.8} & 55.0 & 65.5 & 68.2 & 74.4 & 70.8 & 42.5 & \textbf{80.1} & 66.9 & 72.3 & 70.8 & 75.8 \\
IN-C & 17.1 & 14.5 & 22.3 & 10.4 & 21.6 & 19.9 & \textbf{23.0} & 31.8 & 15.4 & \textbf{34.3} & 14.2 & 30.5 & 22.9 & 31.8 \\
\midrule
Mean & 43.4 & 41.8 & \textbf{55.8} & 45.9 & 50.2 & 50.6 & 52.9 & 58.3 & 44.3 & \textbf{61.6} & 51.9 & 56.8 & 53.6 & 58.4 \\
\bottomrule
\end{tabular}
}
\vspace{-3mm}
\end{table}

\subsection{Online against offline adaptation}
One could also use the preview offline, with several passes of self-supervised learning over the stored images. Our setup excludes this because it keeps the images, and it also costs far more. For each preview domain, \cref{tab:compute} gives the FLOPs, time and energy of one online pass of TTA over the preview, of what \ours adds to it, and of SimCLR and DINO over the stored preview.

\begin{table}[H]
\centering
\caption{\textbf{Compute per \prv{preview} domain.} FLOPs of the forward and backward passes, counted with the FLOP counter of PyTorch on convolutions and matrix products. Time and energy (NVML counter) on one RTX A4500 with nothing else running, measured over 10 steps and scaled to the whole \prv{preview}. SimCLR uses two views, batches of 256 (C100-C) and 64 (IN-C, to fit in memory) and a projection head 2048-2048-128. DINO uses 2 global and 6 local crops, a teacher forward pass and a head to 65\,536 outputs. \ours adds $6.7$ to $7.5\%$ of FLOPs and $5$ to $12\%$ of time and energy to TTA during the \prv{preview}, while the offline methods cost $18$ to $416$ times as much and keep the \prv{preview}. Peak memory: $1.8$ and $5.8$\,GB for \cltta and \ours, $14.0$ and $11.5$\,GB for SimCLR, $6.9$ and $9.7$\,GB for DINO (C100-C and IN-C).}
\label{tab:compute}
\vspace{-2mm}
\small
\fitwidth{\begin{tabular}{@{}l rrr rrr c@{}}
\toprule
 & \multicolumn{3}{c}{C100-C (ResNeXt-29, 5{,}000 images)} & \multicolumn{3}{c}{IN-C (ResNet-50, 20{,}000 images)} & stores the \\
\cmidrule(lr){2-4}\cmidrule(lr){5-7}
 & TFLOPs & time (s) & energy (kJ) & TFLOPs & time (s) & energy (kJ) & preview \\
\midrule
\cltta: one online pass over the preview (Tent) & 42.3 & 5.8 & 1.17 & 486.0 & 51.4 & 9.89 & no \\
\textbf{\ours}: the same pass + frozen first block and moments, no gradient & 45.5 & 6.3 & 1.23 & 518.4 & 57.6 & 11.0 & no \\
\midrule
SimCLR, 10 epochs & 847.7 & 115.3 & 22.6 & 9{,}725.5 & 1{,}031.6 & 203.1 & yes \\
SimCLR, 100 epochs & 8{,}477.3 & 1{,}153.1 & 225.7 & 97{,}254.5 & 10{,}316.0 & 2{,}031.1 & yes \\
DINO, 10 epochs & 1{,}756.6 & 256.6 & 48.4 & 18{,}604.6 & 2{,}316.5 & 458.9 & yes \\
DINO, 100 epochs & 17{,}565.9 & 2{,}566.2 & 483.7 & 186{,}046.1 & 23{,}165.2 & 4{,}589.0 & yes \\
\bottomrule
\end{tabular}
}
\vspace{-3mm}
\end{table}

\subsection{Why CODA-Prompt is not above DER++}
CODA-Prompt learns each new domain as well as FT. However, it has no buffer and a single shared classifier head, so it forgets across domains like FT. Its published gains are for class-incremental learning, with logit masking and many more epochs per task, which our single-epoch domain-incremental protocol does not use.

\end{document}